\documentclass[journal=jacsat,manuscript=article]{achemso}
\usepackage[version=3]{mhchem} 
\usepackage{amsfonts}

\author{Jianyuan Deng}
\affiliation[Stony Brook University]
{Department of Biomedical Informatics, Stony Brook University, Stony Brook, NY, USA}
\email{jianyuan.deng@stonybrook.edu}

\author{Zhibo Yang}
\affiliation{Department of Computer Science, Stony Brook University, Stony Brook, NY, USA}

\author{Iwao Ojima}
\affiliation{Department of Chemistry, Stony Brook University, Stony Brook, NY, USA}

\author{Dimitris Samaras}
\affiliation{Department of Computer Science, Stony Brook University, Stony Brook, NY, USA}

\author{Fusheng Wang}
\affiliation[Stony Brook University]
{Department of Biomedical Informatics, Stony Brook University, Stony Brook, NY, USA}
\alsoaffiliation{Department of Computer Science, Stony Brook University, Stony Brook, NY, USA}

\title[An \textsf{achemso} demo]
  {Artificial Intelligence in Drug Discovery: Applications and Techniques}

\abbreviations{AI}

\begin{document}
\begin{abstract}

Artificial intelligence (AI) has been transforming the practice of drug discovery in the past decade. Various AI techniques have been used in many drug discovery applications, such as virtual screening and drug design. In this survey, we first give an overview on drug discovery and discuss related applications, which can be reduced to two major tasks, i.e., molecular property prediction and molecule generation. 
We then present common data resources, molecule representations and benchmark platforms.
As a major part of the survey, AI techniques are dissected into model architectures and learning paradigms. To reflect the technical development of AI in drug discovery over the years, the surveyed works are organized chronologically.
We expect that this survey provides a comprehensive review on AI in drug discovery.
We also provide a GitHub repository with a collection of papers (and codes, if applicable) as a learning resource, which is regularly updated.
  
\end{abstract}

\section{Introduction}\label{sec1}
Drug discovery is well known as an expensive, time-consuming process, with low success rates. On average, developing a new drug costs 2.6 billion US dollars \cite{mullard2014new} and can take more than 10 years. Moreover, the success rate of launching a drug to market from Phase I clinical trial is daunting, less than 10\% \cite{dowden2019trends}.
In the past decade, the practice of drug discovery has been undergoing radical transformations driven by the rapid development in artificial intelligence (AI) \cite{schneider2018automating, chen2018rise, mater2019deep, vamathevan2019applications, paul2020artificial}. 
Popular applications of AI in drug discovery include virtual screening \cite{stumpfe2020current}, \textit{de novo} drug design \cite{schneider2020rethinking}, retrosynthesis and reaction prediction \cite{bostrom2018expanding}, and \textit{de novo} protein design \cite{strokach2020fast}, among others, which can be reduced to two categories, i.e., predictive and generative tasks.
To power these AI applications, a wide range of AI techniques are involved, with model architectures evolving from traditional machine learning models to deep neural networks, such as convolutional neural networks, recurrent neural networks, graph neural networks and transformers, etc. 
Learning paradigms also shift from supervised learning to self-supervised learning and reinforcement learning.

In this survey, we focus on the applications and techniques of AI-driven discovery on small-molecule drugs. Biologics (e.g. antibodies, vaccines) are not covered. 
We first provide an overview of key applications in drug discovery and point out a collection of previously published perspectives, reviews, and surveys. 
We then introduce common data resources and representations of small molecules. We also discuss existing benchmark platforms for both molecular property prediction and molecule generation.
With knowledge on data and representations, the related techniques, including model architectures and learning paradigms, will be elaborated. 
Finally, we discuss existing challenges and highlight some future directions.
By assembling a Github repository \footnote{\url{https://github.com/dengjianyuan/Survey_AI_Drug_Discovery}} with the surveyed papers (and codes, if applicable), we expect this survey not only provides a comprehensive overview of AI in drug discovery but also serves as a learning resource for researchers interested in the interdisciplinary field. 

\section{Drug Discovery Overview}\label{sec1_1}
In this section, we first go over the definitions of key concepts in drug discovery, mainly on the screening and design of small molecules. 
Note that AI-powered drug repositioning, which repurposes existing drugs or drug combinations for new indications, is not included. 
Besides, target identification, exploiting -omics data for its druggability, is also out of the scope and thus not discussed. 
Rather, we refer the readers to previous publications on drug repositioning \cite{pushpakom2019drug, tsigelny2019artificial} and -omics data for target identification \cite{paananen2020omics}.

\textit{Drug discovery} \cite{hughes2011principles} is a project motivated by the situation when there are no drugs for a disease or when existing drugs have limited efficacy and/or severe toxicity.
At the earliest stage, an underlying hypothesis needs to be developed that activation or inhibition of a target (e.g., an enzyme, a receptor, an ion channel, etc) results in therapeutic effects for the disease, which involves target identification and target validation. 
For the selected target, intensive assays will be performed to find the hits and subsequently the leads (i.e., drug candidates), which involves \textit{hit discovery}, \textit{hit-to-lead phase} and \textit{lead optimization}.
The drug candidates then enter preclinical studies and clinical trials. If successful, the drug candidate can be launched into market as a medical product to treat the disease.

To accelerate the small-molecule drug discovery, \textit{high-throughput screening (HTS)} \cite{pereira2007origin, bender2008aspects} has been proposed to increase the discovery efficiency since the 1980s, which is a hit-finding approach underpinned by development in automation and the availability of large chemical libraries. 
A prominent outcome of HTS is the large-scale \textit{structure-activity relationship (SAR)} datasets, which contribute to the chemical databases such as PubChem \cite{wang2017pubchem} and ZINC \cite{sterling2015zinc}.
Various computational techniques have been developed to search the chemical libraries for potentially active molecules to be tested in subsequent \textit{in vitro} and \textit{in vivo} assays \cite{kim2016getting}, which is also known as \textit{virtual screening (VS)}. In other words, VS is to identify active molecules using computational approaches, based on knowledge about the target (\textit{structure-based VS}) or known active ligands (\textit{ligand-based VS}) to increase the odds of identifying active molecules \cite{scior2012recognizing}. 
For the concept of active molecules, as mentioned above, activation or inhibition of a target is the underlying hypothesis to treat a disease, which corresponds to two major classes of drugs with regard to the \textit{mechanism of action (MoA)}, i.e., \textit{agonists} and \textit{antagonists} \cite{salahudeen2017overview}.
An agonist is a molecule which activates the target to exert a biologic response as its endogenous ligand. On the contrary, an antagonist is a molecule which binds to the target to block the response. Based on more specific effects and mechanisms, agonists can further be classified as partial agonists, inverse agonists, biased agonists. Antagonists include competitive and non-competitive antagonists.
To quantify the activity, various assays have been developed to measure \textit{affinity} (or \textit{potency}) and \textit{efficacy}. Affinity is the fraction or extent to which a molecule binds to a target at a given concentration whereas potency is the necessary amount of a molecule to produce an effect of a given magnitude, inversely proportional to the affinity. 
Efficacy, on the other hand, describes the effect size, such as inhibition of an enzyme to 60\%.
Common activity measures are summarized in Table~\ref{tbl:prpt_measures}.

\begin{table}[ht]
  \caption{Common Measures of Molecule Activity}
  \label{tbl:prpt_measures}
  \begin{tabular}{ll}
    \hline
    Measures  & Definition  \\
    \hline
    Kd & Equilibrium dissociation constant \\
    Km & Michaelis constant \\
    Ki & Inhibition constant \\
    \hline
    IC50 & Half maximal inhibitory concentration \\
    EC50 & Half maximal effective concentration  \\
    \hline
  \end{tabular}
\end{table}

Nevertheless, sufficient activity is not the only criterion for an ideal drug candidate, which only makes it a \textit{ligand}.
Binding specificity is another concern \cite{hu2013compound}. Most often, a molecule can bind to multiple targets and unexpected side effects may arise due to binding promiscuity. Thus, high selectivity is another desired feature. 
In fact, drug candidates should also satisfy a combination of criteria \cite{yusof2014finding}, with optimal physicochemical (\textit{water solubility}, \textit{acid-base dissociation constant}, \textit{lipophilicity}, \textit{permeability}), pharmacokinetic (\textit{absorption}, \textit{distribution}, \textit{metabolism}, \textit{excretion}), and pharmacodynamic (\textit{activity}, \textit{selectivity}) properties. 
Other properties considered during compound synthesis include the Synthetic Accessibility Score (SAS) and the Quantitative Estimation of Drug-likeness (QED). SAS is a heuristic score of how hard (10) or easy (1) it is to synthesize a given molecule based on a combination of the molecular fragments' contributions. QED is an estimate (0-1) on how likely a molecule is a viable drug candidate.
At its core, drug discovery is a multi-objective optimization problem \cite{nicolaou2013multi}.
Usually, for each property of interest, a predictive model is built to map the molecular structure to the property value with either classification or regression, which is broadly referred to as \textit{quantitative structure-activity relationship (QSAR)} modeling \cite{muratov2020qsar}. 
A more intriguing prospect from QSAR is that these models can be exploited inversely to reveal the structural features underlying the optimal properties to guide \textit{drug design} from scratch, also known as \textit{de novo} drug design. 

Rather than merely screening existing chemical libraries \cite{schneider2005computer}, drug design takes a step further to explore the vast \textit{chemical space}, i.e., the space encompassing all possible small molecules \cite{dobson2004chemical} which has an estimated size around $10^{30}-10^{60}$ \cite{sliwoski2014computational}. 
In drug design, there is a \textit{design-make-test-analysis (DMTA)} cycle, which consists of iterative organic synthesis and property assays \cite{schneider2018automating}. 
To efficiently navigate the chemical space, quantitative drug design has been proposed since late 1970s \cite{van2007computer}. 
Essentially, drug design epitomizes in two questions \cite{jimenez2020drug}: 1) ``Can molecular properties be deduced from molecular structures?" and 2) ``Which structural features are relevant for certain molecular properties?" The former one also underlies the core assumption of VS and the latter is what QSAR tries to answer. In this sense, drug design can be viewed as an extension to VS, and involves both \textit{molecular property prediction} and \textit{molecule generation}, which are the major tasks in current AI-driven drug discovery.

\subsection{\textbf{Summary of Existing Reviews}}\label{sec1_2}
Next, we briefly discuss existing reviews on AI-driven discovery by dividing them into three categories: 1) ``General drug discovery review", 2) ``Drug discovery in the AI-era", and 3) ``Rethinking AI-driven drug discovery".

\subsubsection{\textbf{General Drug Discovery Reviews}}\label{sec1_2_1}
Many existing papers have covered the general aspects of drug discovery \cite{hughes2011principles, schneider2018automating} and related concepts, such as chemical space \cite{dobson2004chemical}, VS and HTS \cite{bajorath2002integration, schneider2010virtual, scior2012recognizing}, optimal properties for drug candidates \cite{yusof2014finding}, QSAR \cite{polishchuk2017interpretation, muratov2020qsar}, target prediction \cite{sydow2019advances}, and computer-aided drug design \cite{schneider2005computer}.
Besides, one prominent challenge in drug discovery is that molecular properties can be highly sensitive to minor structural changes. This is also known as the \textit{activity cliffs (ACs)}, where pairs of structurally similar molecules exhibit significantly different activities \cite{maggiora2006outliers, stumpfe2014recent, bajorath2019duality}.
We strongly recommend the readers (especially those new to drug discovery) refer to these reviews for a better understanding on drug discovery and recognition of potential pitfalls.

\subsubsection{\textbf{Drug Discovery in the AI Era}}\label{sec1_2_2}
AI has been widely applied in drug discovery. Since the early 2000s, machine learning models, such as random forest (RF), have been exploited for VS and QSAR \cite{ma2015deep, lavecchia2015machine}.
In 2012, AlexNet \cite{krizhevsky2012imagenet} marked the advent of the deep learning era \cite{alom2018history}. 
Shortly after in the 2012 Merck Kaggle competition, deep neural networks (DNN) outperformed the standard RF model in predicting molecular activities \cite{ma2015deep}.
More recently, the success of AI techniques in computer vision and natural language processing has shed more light into drug discovery \cite{chen2018rise, vamathevan2019applications, ozturk2020exploring, jimenez2021artificial} and led to the burgeoning field of deep learning in chemistry \cite{mater2019deep}. 
In 2019, potent inhibitors of discoidin domain receptor 1 (DDR1) were discovered in 21 days by researchers from Insilico Medicine \cite{zhavoronkov2019deep}. In 2020, a novel antibiotic candidate against antibiotic-resistant bacteria, halicin, was identified by researchers from MIT \cite{stokes2020deep}. 
Note that AI can be applied at different stages in drug discovery, from target identification and validation to drug response determination \cite{vamathevan2019applications}.
Lead identification, the focus of this survey, involves two fundamental tasks, i.e., molecular property prediction and molecule generation.
Molecular property prediction, at the core of VS, is to predict the property value of a molecule given its structure or learned representation \cite{chuang2020learning}, which can be served for various purposes, such as drug-target interaction (DTI) prediction \cite{sydow2019advances}, toxicity prediction \cite{mayr2016deeptox} and drug-induced liver injury (DILI) prediction \cite{andrade2019drug}, among others.
Molecule generation, underlying drug design, involves two levels of tasks: 
1) realistic molecule generation, i.e., generating molecules within constraints imposed by the chemical rules, and 
2) goal-directed molecule generation, i.e., generating chemically valid molecules with desired properties \cite{elton2019deep, mercado2020practical}.

\subsubsection{\textbf{Rethinking AI-driven Drug Discovery}}\label{sec1_2_3}
Despite the promise of AI in drug discovery, pitfalls still exist, which have been widely discussed \cite{schneider2020rethinking, schaduangrat2020towards, stumpfe2020current, jimenez2020drug, bender2020artificial, bender2021artificial, walters2021critical}. 
As opined by Bender et al \cite{bender2020artificial, bender2021artificial}, ``a method cannot save an unsuitable representation which cannot remedy irrelevant data for an ill thought-through question".
To circumvent potential hypes and unrealistic expectations thereof, there is indeed a necessity to take into consideration the hypotheses, the data, the representations, the models, the learning paradigms and moreover, these components as a whole, for any AI-driven application in drug discovery.

\begin{figure*}[h]
\centering
  \includegraphics[width=1\linewidth]{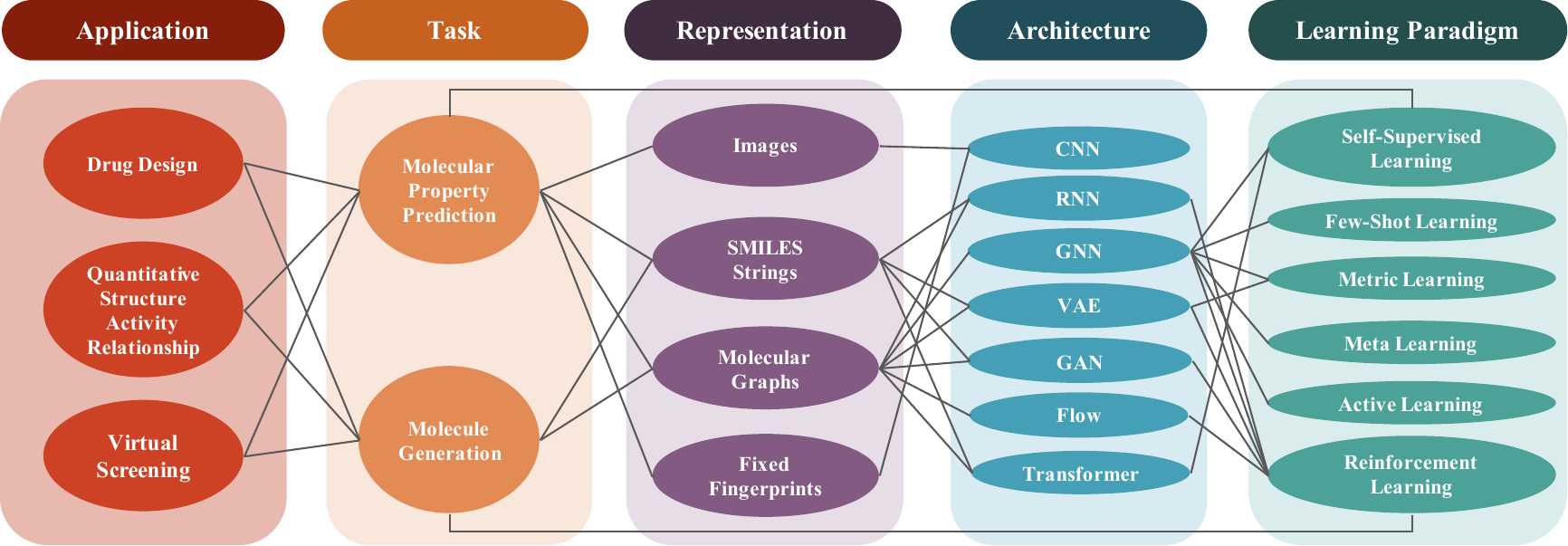}
  \caption{Applications and Techniques of AI in Drug Discovery. The applications of AI in small-molecule drug discovery include virtual screening, quantitative structure-activity relationship and drug design, which can be reduced to two major tasks: molecular property prediction and molecule generation. Small molecules can be represented by fixed fingerprints, molecular graphs, simplified molecular input entry system (SMILES) strings, and images. Various model architectures have been applied on each representation format, including convolutional neural networks (CNN), recurrent neural networks (RNN), graph neural networks (GNN), variational autoencoders (VAE), generative adversarial networks (GAN), normalizing flow models and transformers. Still, challenges exist for the low-data molecular property prediction and goal-directed molecule generation. To tackle these challenges, different learning paradigms have been proposed, such as self-supervised learning for the pretraining-finetuning practice and reinforcement learning for navigating the chemical space search. Other paradigms surveyed here also include few-shot learning, metric learning, meta learning and active learning. }
  \label{fgr:roadmap}
\end{figure*}

\subsection{\textbf{Structure of the Survey}}\label{sec1_3}
Given the necessity of a clear understanding on both drug discovery applications and AI techniques, this survey starts from general aspects in drug discovery and then moves on to AI-driven drug discovery, covering data resources, molecule representations, model architectures, and learning paradigms. The organization of this survey is depicted in Fig~\ref{fgr:roadmap}.

Notably, current literature on the AI techniques is often fragmented.
More often than not, the rationale behind the choice of a technique is simply because it has not been previously investigated \cite{mercado2020practical}.
To gain more insights into the strengths and weaknesses of these AI techniques, we focus on the model architectures and learning paradigms. 
We also try to present the surveyed works chronologically so as to reflect the technical development over the years.
Finally, we highlight existing challenges and future directions (see Section~\ref{sec5} ``Discussion").

\section{Data, Representation and Benchmark Platforms}\label{sec2}
In this section, we first discuss the publicly available data resources.
Then, we discuss how small molecules can be represented in machine-readable formats. 
Lastly, we summarize current benchmark platforms for both molecular property prediction and molecule generation.

\subsection{\textbf{Public Data Resources}}\label{sec2_1}
With the improvements in HTS and related assays, data on molecular activity and related properties are ever increasing, which contribute to various public data resources.
These resources typically provide information on molecular structures, molecular properties and target information \cite{rifaioglu2019recent}, as discussed below. 

\textit{PubChem} \cite{kim2021pubchem} was launched by the National Institutes of Health in 2004. With a collection of chemical information from 750 data sources, PubChem is the largest chemical database. As of August 2020, PubChem contains 111 million unique chemical structures with 271 million activity data points from 1.2 million biological assays experiments. 
PubChem provides direct download as well as web interfaces for online queries.
Notably, PubChem is non-curated \cite{rifaioglu2019recent} and the bioactivity datasets from PubChem can be highly imbalanced \cite{korkmaz2020deep}. Researchers may curate the data on their own. For example, Chithrananda et al \cite{chithrananda2020chemberta} recently released a curated dataset of 77 million SMILES strings from PubChem.
\textit{ChEMBL} \cite{gaulton2017chembl}, maintained by the European Molecular Biology Laboratory, is another large-scale chemical database. For example, in ChEMBL22 (version 22), there are more than 1.6 million distinct chemical structures with over 14 million activity values. Moreover, ChEMBL is manually curated in a comprehensive manner \cite{rifaioglu2019recent}.  ChEMBL provides downloads in a variety of formats (e.g., Oracle, MySQL or PostgreSQL database) and also allows web application program interface (API) for data retrieval in XML or JSON format \cite{davies2015chembl}.
Notably, based on ChEMBL, Mayr et al \cite{mayr2018large} extracted a large-scale benchmark dataset for target prediction.
ZINC \cite{sterling2015zinc}, developed by the Irwin and Shoichet Laboratories in UCSF, contains a suite of molecules, annotated ligands and targets as well as the purchasability for over 120 million ``drug-like" compounds. ZINC supports direct download from the website and also provides an API for retrieving data.
Notably, some subsets of the ZINC database are more commonly used, such as the ZINC-250k \cite{rong2020grover} and the ZINC Clean Leads collections \cite{polykovskiy2020molecular}.

In addition to the aforementioned large-scale databases, there are also other data repositories \cite{lagarde2015benchmarking}, such as PDBbind, BindingDB, DUD, DUD-E, MUV, STITCH, GLL\&GDD, NRLiST BDB, KEGG, among others. 
Besides databases mainly derived from preclinical studies, there are also public data resources for the marketed drugs and their effects in human subjects, such as the datasets for adverse drug reactions (ADR) (e.g., DrugBank, SIDER, OFFSIDES and TWO-SIDES) and the datasets for DILI ( e.g., DILIrank) \cite{chen2016dilirank}.

\subsection{\textbf{Small Molecule Representations}}\label{sec2_2}
\begin{figure}[ht]
\centering
  \includegraphics[width=1\linewidth]{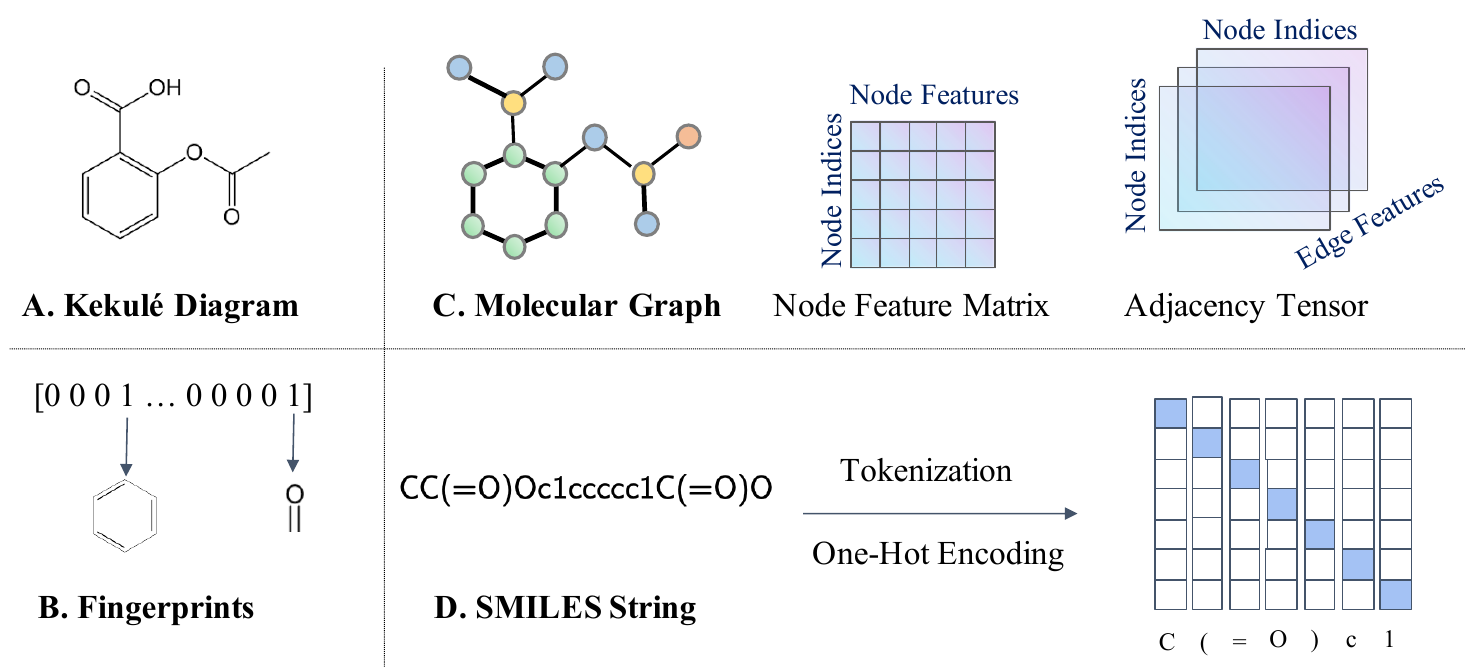}
  \caption{Illustration of Small Molecule Representations. }
  \label{fgr:molrep}
\end{figure}

Molecules are often depicted as Kekulé diagrams with bonds and atoms (Fig~\ref{fgr:molrep}A). 
Over the years, machine-readable representations have been developed to enable rapid computation, querying and storage of the molecules \cite{david2020molecular}. 
Molecules can be represented by the fixed molecular descriptors, which are further categorized by its dimensionality \cite{rifaioglu2019recent}. 
Specifically, there are 0D descriptors for molecules, such as molecular weight (MW), atom number, and atom-type count. 
0D descriptors can be directly derived from the empirical formula and barely provide information on how atoms are connected.
For example, the empirical formula of alanine, $C_3H_7NO_2$, can also represent lactamide \cite{david2020molecular}.
To highlight different functional groups, descriptors incorporating more structural information have been proposed, such as fingerprints (Fig~\ref{fgr:molrep}B). Fingerprints are binary vectors with each dimension in the vector indicating the presence or absence of a particular substructure. 
Among them, there are 1D descriptors to represent the substituent atoms, chemical bonds, structural fragments, and functional groups. 
There are also 2D descriptors to represent the atom connectivity and molecular topology, such as 1) Keyed fingerprints - molecular access system (MACCS) keys, 2) Path-based fingerprints - DayLight fingerprints and 3) Circular fingerprints - extended connectivity fingerprints (ECFPs) based on the Morgan algorithm \cite{morgan1965generation}. 
Furthermore, 3D descriptors have also been developed to encode 3D-structural information like the steric properties, surface area, volume and binding site properties, among others. 

Molecular descriptors have greatly boosted the application of computational methods, including machine learning models, in drug discovery \cite{subramanian2016computational, zang2017silico}. Nevertheless, these descriptors are fixed and not learnable towards improving model performance. With the advent of the AI era, various deep learning models have paved the way for end-to-end (E2E) predictions, where molecules can be embedded into a continuous latent space without hand-crafted rules. Among them, two major representation formats are molecular graphs and the simplified molecular input entry system (SMILES) strings \cite{david2020molecular}.

\subsubsection{\textbf{Molecular Graphs}}\label{sec2_2_1}
The idea of graph representation is intuitive, where atoms are typically mapped to nodes and bonds to edges. 
Formally, a graph is defined as $G = (V, E)$, a set of of nodes (atoms) $V$ and a set of edges (bonds) $E$, where $(v_i, v_j) \in E$ indicates a bond between atoms $v_i$ and $v_j$ \cite{david2020molecular}. 
The attributes of atoms are represented by a node feature matrix $X$ and each node $v$ can be represented by an initial vector $x_v$ and a hidden vector $h_v\in R^D$. 
Similarly, the attributes of bonds can also be represented by an edge feature matrix. 
Note that both node and edge feature matrices do not directly encode connections. Instead, an adjacency matrix $A$ keeps track of the pairwise connection status. The element of $A$, $a_{ij}$, if equals to 1, means that there is a bond connecting node $v_i$ and $v_j$; otherwise, if $a_{ij}$ equals to 0, these two nodes are not connected by a bond. Usually, the edge feature matrix and the adjacency matrix are combined to form an adjacency tensor (Fig~\ref{fgr:molrep}C).
Common node and edge features \cite{yang2019analyzing, mercado2020graph} are summarized in Table~\ref{tbl:molgraph_features}.

\begin{table}[ht]
  \caption{Common Node and Edge Features in Molecular Graphs}
  \label{tbl:molgraph_features}
  \begin{tabular}{lll}
    \hline
    Type  & Feature & Notes  \\
    \hline
    Node & Atom type  & Element type  \\
    Node & Formal charge & Assigned charges  \\
    Node & Implicit Hs & Number of bonded hydrogens\\
    Node & Chirality & R or S configuration \\
    Node & Hybridization & Orbital hybridization: $sp^{x}$, $sp^{x}d^{y}$ \\
    Node & Aromaticity & Aromatic atom or not\\
    \hline
    Edge & Bond type & Single, double, triple or aromatic\\
    Edge & Conjugated  & Conjugated or not \\
    Edge & Stereoisomers  & cis (Z) or trans (E)\\
    \hline
  \end{tabular}
\end{table}

One advantage of the graph representations is that they carry more structural information. Besides, molecular graphs as well as the subgraphs can be directly mapped to a chemical (sub-)structure and thus are highly interpretable \cite{jin2020multi}. 
One drawback of the graph representation, however, is that these matrices require a large amount of disk space for storage and significant memory during computation, which may slow down the efficiency during molecule generation \cite{david2020molecular}.

\subsubsection{\textbf{SMILES Strings}}\label{sec2_2_2}
To accommodate the storage and computation efficiency, molecules are also commonly represented by the SMILES strings \cite{weininger1988smiles}. 
In SMILES, an atom is represented by the atomic symbols; for two-character symbols, the second letter will be represented in lower case. Elements in the organic subset, namely B, C, N, O, P, S, F, Cl, Br and I, can be written without brackets whereas for those not included, brackets should be applied with the attached hydrogens and formal charges written inside, such as [Fe2+]. The lower-case letters represent the atoms in aromatic rings; for instance, C is used for the normal carbon and c is used for the aromatic carbon.
For bonds, there are single, double, triple and aromatic bonds, represented by the symbols -, =, \# and :, respectively, where single bonds and aromatic bonds are usually omitted.
For the branches in a molecule, they are denoted by enclosures in parentheses. 
To represent the cyclic structure, a single or aromatic bond is first broken down in the ring and then the bonds are numbered in any order with the ring-opening bonds by a digit following the atomic symbol at each ring.
Notably one molecule may correspond to multiple SMILES strings \cite{david2020molecular}. To avoid conflicts, canonicalization methods \cite{weininger1989smiles} have been introduced to ensure only one unique SMILES string is designated for the same molecule.

Usually the SMILES string are converted into one-hot vectors before fed into the machine learning models (Fig~\ref{fgr:molrep}D) \cite{bian2021generative}. 
Comparing to the graph representation, SMILES string is less computationally expensive.
However, since SMILES strings do not directly encode the atomic connection, there can be a loss of the structural information \cite{xiong2019pushing}.
Besides, due to the internal syntax of the SMILES (e.g., ring opening and closure, atom valency), using this linear notations for molecule generation is prone to generate invalid molecules \cite{gomez2018automatic, popova2018deep}. 

\subsubsection{\textbf{Other Representations}}\label{sec2_2_3}
Molecules can also be represented by more sophisticated 3D-atomic coordinates, commonly seen in structure-based VS or QSAR studies \cite{ragoza2017protein, jimenez2018k, lim2019predicting, hernandez2019quantum}. Molecular topology, such as bond lengths, bond angles and torsional angles, can also be incorporated \cite{wu2018quantitative}. 
Some works have already exploited the 3D-representation to generate molecules \cite{skalic2019shape, simm2020reinforcement}.
In addition to the raw 3D coordinates, well-established 3D properties, which capture the molecular conformation, can also be readily utilized for prediction tasks \cite{hemmerich2020cover}.
Furthermore, with the advances of computer vision, images of molecular structures (Fig~\ref{fgr:molrep}A) emerge as another modality to represent molecules \cite{fernandez2018toxic, meyer2019learning, cortes2019kekulescope, rifaioglu2020deepscreen, rajan2021decimer}.

\subsection{\textbf{Benchmark Platforms}}\label{sec2_3}
To evaluate the performance of molecular property prediction and molecule generation, there are several benchmark platforms, which are discussed below.

As a major benchmark dataset platform for molecular property prediction, MoleculeNet was released by Wu et al in 2018 \cite{wu2018moleculenet}, which includes a set of datasets along with an open-source DeepChem package \cite{Ramsundar-et-al-2019}. 
The benchmark datasets cover four categories: 
1) Quantum mechanics (QM7, QM7b, QM8, QM9), 
2) Physical chemistry (ESOL, FreeSolv, Lipophilicity), 
3) Biophysics (PCBA, MUV, HIV, PDDBind, BACE), and 4) Physiology (BBBP, Tox21, ToxCast, SIDER, ClinTox), involving single task or multi tasks. 
Notably, for molecular property prediction, datasets can be highly imbalanced. Thus, when choosing evaluation metrics (Table~\ref{tbl:metrics}), positive rates should be considered. For instance, AUPRC is favored over AUROC in case of a low positive rate (e.g. less than 2\%) \cite{wu2018moleculenet}.
With regard to the datset splitting, in addition to the common random split, MoleculeNet also provides other splitting ways, namely, scaffold split, stratified split and time split for different datasets. In other words, for each dataset, the recommended split way varies. For example, for the BACE dataset, since it is for a single target, the scaffold splitting is more suitable, whereas for the PDBind dataset, since the data collection is over a long period, time splitting is recommended to better reflect the actual drug discovery effort over the years.

One limitation of MoleculeNet, however, is that it does not provide explicit training, validation and test folds for the datasets \cite{fabian2020molecular}. To improve reproducibility, the ChemBench package from MolMapNet \cite{shen2021out} was released recently.
MolMapNet also expands the MoleculeNet by adding pharmacokinetics-related datasets, such as PubChem CYP inhibition and liver microsomal clearance data.
In addition to the benchmark datasets, Chemprop \cite{yang2019analyzing}, for benchmarking the learned molecular representations, was proposed in 2019, which systematically compared the fixed molecular descriptors (e.g. ECFPs) and the learned molecular representations for molecular property prediction.
In Chemprop, models were benchmarked extensively on 19 public and 16 proprietary industrial datasets. 
As a side note, Chemprop is related to the discovery of halicin \cite{stokes2020deep}.

As for benchmarking the molecule generation models, Olivecrona et al \cite{olivecrona2017molecular} developed REINVENT in 2017, which is a sequence-based generative model utilizing SMILES strings. 
REINVENT can be used to execute a range of tasks, such as generating analogues to a query structure and generating ligands for a given target. 
In 2020, Blaschke et al \cite{blaschke2020reinvent} proposed the updated version, REINVENT 2.0, as a production-ready tool for drug design.
For benchmarking molecule generation utilizing molecular graphs, Mercado et al \cite{mercado2020graph} proposed GraphINVENT in 2020. 
To standardize the assessment for molecule generation, an evaluation framework GuacaMol \cite{brown2019guacamol} was proposed in 2019, which set a suite of tasks for distribution learning and goal-directed design. More specifically, the generative model is examined on whether it can reproduce the property distribution of training sets (usually for VS purpose \cite{segler2018generating}), and find the optimal molecule (for multi-objective optimization).
A more recent evaluation platform MOSES was released by Polykovskiy et al in 2020 \cite{polykovskiy2020molecular}, which compiles a list of metrics (Table~\ref{tbl:metrics} \footnote{QSAR: quantitative structure-activity relationship; Recall@k: recall among top k molecules; Precision@k: precision among top k molecules; AP@k: average precision among top k molecules; AUROC: area under the receiver-operating characteristic curve; AUPRC: area under the precision-recall curve; MAE: mean absolute error; RMSE: rooted mean square error; Unique@k: uniqueness of the first k valid (generated) molecules; FCD: Fréchet ChemNet Distance; KL divergence: Kullback-Leibler divergence.
}) for detecting common issues in generative models, such as overfitting and mode collapse.

\begin{table}[ht]
  \caption{Commonly Used Evaluation Metrics }
  \label{tbl:metrics}
  \begin{tabular}{llll}
    \hline
    Application & Task & Metric & Purpose  \\
    \hline
    Virtual screening & Molecular property prediction & Recall@k & Retrieval \\
    Virtual screening & Molecular property prediction & Precision@k & Retrieval \\
    Virtual screening & Molecular property prediction & AP@k & Retrieval \\
    \hline
    QSAR & Molecular property prediction & Accuracy & Classification \\ 
    QSAR & Molecular property prediction & Recall  & Classification  \\
    QSAR & Molecular property prediction & Precision  & Classification  \\
    QSAR & Molecular property prediction & AUROC & Classification \\
    QSAR & Molecular property prediction & AUPRC & Classification \\
    QSAR & Molecular property prediction & MAE & Regression \\
    QSAR & Molecular property prediction & RMSE & Regression \\
    \hline
    Drug design & Molecule generation & Validity & Distribution learning\\
    Drug design & Molecule generation & Unique@k & Distribution learning \\
    Drug design & Molecule generation & Novelty & Distribution learning \\
    Drug design & Molecule generation & Diversity & Distribution learning \\
    Drug design & Molecule generation & FCD & Distribution learning\\
    Drug design & Molecule generation & KL divergence & Distribution learning\\
    Drug design & Molecule generation & Scaffold similarity & Goal-directed design\\
    Drug design & Molecule generation & Rediscovery & Goal-directed design\\
    \hline
  \end{tabular}
\end{table}

For instance, validity measures how well a model explicitly captures the chemical rules, such as valency; uniqueness and diversity examine whether the generative model collapses to producing only a limited set of molecules; novelty indicates whether the model is overfitted to just memorize the training examples.
Furthermore, Fréchet ChemNet Distance (FCD) is a measure of how close the distributions of the generated set are to the distribution of molecules in the training set. A low FCD value corresponds to similar molecule distributions.
Kullback-Leibler (KL) divergence measures the difference between two probability distributions. When the KL divergence value is small, the generated molecules can well approximate the targeted property in the training set. 
For goal-directed design, it relies on a formalism where molecules are scored individually based on each pre-defined criterion, such as containing a specific sub-structure, having certain physicochemical properties or exhibiting similarity or dissimilarity to certain molecules.
Consequently, similarity and rediscovery are usually used for evaluation purpose. Specifically, rediscovery assesses if the generative model is able to rediscover a given molecule and similarity evaluates whether the model can generate molecules similar or dissimilar to a given molecule. 

\section{Model Architectures}\label{sec3}
Prior to the ``deep learning" era, traditional machine learning models were widely used in VS \cite{lavecchia2015machine}. Pertinent tasks include predictions of drug likeliness \cite{walters2002prediction, schneider2008gradual}, physicochemical properties \cite{palmer2007random, schroeter2007machine}, pharmacokinetic parameters \cite{hou2007adme, tian2011adme, sakiyama2008predicting, vasanthanathan2009classification} and pharmacodynamic properties \cite{riddick2011predicting, zhao2006application}.
There are score-based classification models, support vector machines (SVM) \cite{heikamp2014support} and K nearest neighbors (KNN) and probability-based classification models, random forest (RF) \cite{svetnik2003random}, naive bayes (NB), and logistic regression (LR).
Despite the success of traditional machine learning models, deep neural networks (DNNs) have outperformed them in a variety of tasks \cite{dahl2012deep, mayr2018large}.

\subsection{\textbf{Convolutional Neural Networks}}\label{sec3_1}
\begin{figure}[ht]
\centering
  \includegraphics[width=1\linewidth]{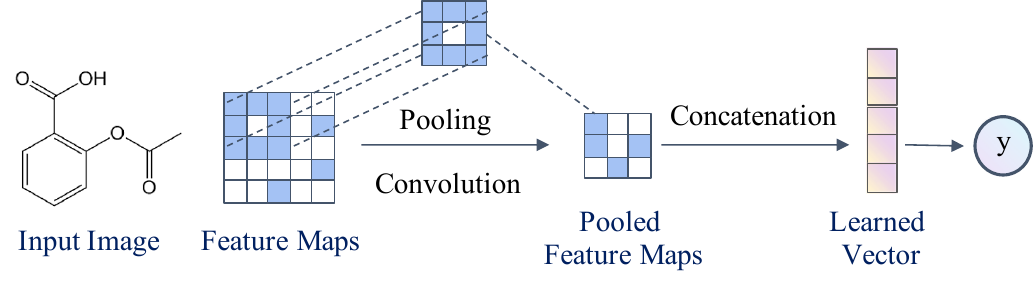}
  \caption{Illustration of Convolutional Neural Networks.}
  \label{fgr:archs_cnn}
\end{figure}

Convolutional neural networks (CNNs) are mainly used in computer vision to process pixels of data in images \cite{lecun2015deep}. 
In CNNs, there are convolution layers and pooling (i.e. subsampling) layers (Fig~\ref{fgr:archs_cnn}). 
On top of these convolution layers and pooling layers, a vector representation is learned by concatenating the feature maps for a final prediction.
CNNs share parameters across the filters, which largely reduces the number of parameters to be learned, thereby decreasing memory consumption and increasing computation speed. 

In drug discovery, CNNs can be applied to elucidate the bioactivity profiles based on microscopy images \cite{simm2018repurposing, hofmarcher2019accurate}.
Moreover, CNNs are also used for molecular property prediction \cite{ramsundar2015massively, duvenaud2015convolutional}. 
In 2015, Duvenaud et al \cite{duvenaud2015convolutional} applied CNNs on circular fingerprints, a refinement of the ECFPs \cite{glen2006circular}, to create a differentiable fingerprint, which is among the first efforts using data-driven representation learning for molecular property prediction, instead of fixed chemical descriptors. This work has greatly motivated learning molecular representations. 

In addition to fingerprints, CNNs can also effectively extract features directly from the images of molecular structure. For instance, Chemception \cite{goh2017chemception} is trained on the 2D-structural images to predict free energy of solvation and inhibition of HIV replication.
Later, Fernández et al \cite{fernandez2018toxic} developed Toxic Colors, a framework for toxicity classification with the images as input. 
Cortes-Ciriano et al \cite{cortes2019kekulescope} further extended existing CNNs architectures (e.g. AlexNet \cite{krizhevsky2012imagenet}, DenseNet-201 \cite{huang2017densely}, ResNet152 \cite{he2016deep} and VGG-19 \cite{simonyan2014very}) to Kekulé structure images for molecular property prediction, also known as KekuleScope. The experimental results of KekuleScope showed that CNNs on images as input can achieve comparable performance to RF and DNNs on ECFPs. 
Meyer et al \cite{meyer2019learning} also predicted the MeSH-therapeutic-use classes based on compound images, which outperformed previous predictions based on transcriptomic data. 
More recently, Rifaioglu et al \cite{rifaioglu2020deepscreen} proposed a large-scale DTI prediction system, DEEPScreen.
Indeed, molecular property prediction with images as input are closely related to the progress in computer vision, which also prompts automatic extraction of chemical structures from literature and patents \cite{staker2019molecular, rajan2021decimer}.
The chemical structure recognition model can be further integrated with models in natural language processing, such as DECIMER \cite{rajan2020decimer} and DECIMER 1.0 \cite{rajan2021decimer}, which are able to translate the bitmap images of a molecule into a SMILES string, as an image captioning task \cite{hossain2019comprehensive}.

\subsection{\textbf{Recurrent Neural Networks}}\label{sec3_2}
\begin{figure*}
\centering
  \includegraphics[width=1\linewidth]{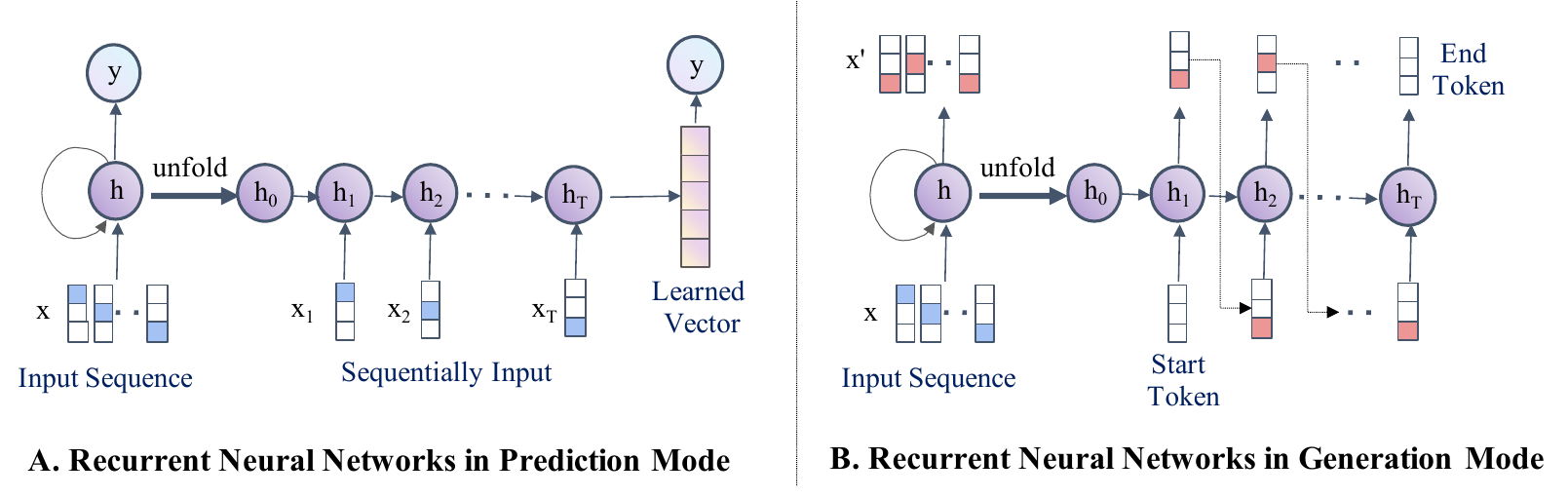}
  \caption{Illustration of Recurrent Neural Networks.}
  \label{fgr:archs_rnn}
\end{figure*}

Recurrent neural networks (RNNs) are mainly used for processing sequential data \cite{lecun2015deep}. RNNs allow the connection among neurons in the same hidden layer to form a directed cycle (Fig~\ref{fgr:archs_rnn}A), thereby enabling the use of sequential input, such as language modeling \cite{mikolov2011extensions} and music generation \cite{boulanger2012modeling}. 
If unfolded in time steps, RNNs can be seen as a very deep feed-forward networks where all layers sharing the same weights. However, the long-term dependency of RNNs makes it difficult to learn the parameters due to the gradient explosion or vanishing problem \cite{lecun2015deep}.
As a result, long short-term memory (LSTM) \cite{hochreiter1997long} and gated recurrent unit (GRU) \cite{chung2014empirical}, two variants of the vanilla RNN, have been developed to augment the network with a memory module.
Different from CNNs' operation on images, RNNs mainly take the SMILES strings as input for molecular property prediction and molecule generation. As discussed in Section~\ref{sec2_2} ``Small Molecule Representation", the characters in a SMILES string are firstly converted into one-hot vectors (Fig~\ref{fgr:molrep}) and then sequentially fed into RNNs, with a hidden vector to be updated at each step. For molecular property prediction, RNNs generate a final output after all steps are taken. For example, SMILES2Vec \cite{goh2017smiles2vec} uses RNNs to learn features from SMILES and predicts a wide range of chemical properties. Mayr et al \cite{mayr2018large} proposed SmilesLSTM to perform DTI prediction, which outperformed traditional machine learning models.

RNNs can also be applied for molecule generation, similar to language models for text generation \cite{olivecrona2017molecular, segler2018generating, neil2018exploring}. 
More specifically, RNNs generate output at each step in an auto-regressive manner (Fig~\ref{fgr:archs_rnn}B), where the output is dependent on the input from previous steps. Based on the input from the current step and prior steps, RNNs output a probability distribution over all possible tokens, from which a token is sampled as the output of the current step and will be used to predict the next token. 
However, due to the syntax of the ``SMILES language" such as the ring opening-closure and the matched brackets, regular RNNs, including LSTM and GRU, cannot capture the algorithmic patterns of the sequence well owing to their inability to count \cite{popova2018deep}. As a result, the generated SMILES strings very often violate the chemical rules and become invalid. Thus, a memory-augmented version, Stack-RNN \cite{joulin2015inferring, popova2018deep}, was developed to alleviate the validity issue for SMILES-based molecule generation.
Another solution for this problem is to adopt bidirectional RNNs, such as the bidirectional LSTM \cite{staahl2019deep, zheng2019identifying}.

In addition to the SMILES strings, RNNs can also be applied on molecular graphs for generation purpose \cite{you2018graphrnn, li2018learning,li2018multi, popova2019molecularrnn}. 
For example, You et al \cite{you2018graphrnn} proposed GraphRNN to generate molecular graphs in an autoregressive manner, decomposing it as a process into generating a sequence of node and edge formations conditioned on the graph structure generated so far. Nonetheless, generating molecular graphs with RNNs requires a full trajectory of the graph generation, which tends to forget the states of initial generation steps quickly. Later You et al proposed GCPN and \cite{you2018graph} designed the graph generation procedure as a Markov Decision Process (MDP), which only needs the intermediate state to generate the graph.
Notably, RNNs can also be components of more complicated generative models, such as variational autoencoders \cite{gomez2018automatic, sattarov2019novo} and generative adversarial networks \cite{guimaraes2017objective, sanchez2017optimizing}. 

\subsection{\textbf{Graph Neural Networks}}\label{sec3_3}
CNNs and RNNs are usually applied on data represented in the Euclidean space. In recent years, graph neural networks (GNNs) are gaining popularity to model data represented in graphs with a set of nodes and edges \cite{wu2020comprehensive}. GNNs can handle node-level (e.g., node classification), edge-level (e.g., link prediction) and graph-level (e.g., graph regression) tasks, with neighborhood aggregation, pooling and readout operations. Small molecules, when represented as molecular graphs (Fig~\ref{fgr:molrep}C), are naturally appealing to the application of GNNs for both molecular property prediction and molecule generation tasks (see Section~\ref{sec2_2_1} ``Molecular Graphs)". 

Two major types of GNNs are convolutional GNNs (ConvGNNs) and recurrent GNNs \cite{wu2020comprehensive}. In the recurrent GNNs, node representation is learned via some recurrent neural architectures, such as the graph gated neural network (GGNN) \cite{li2015gated}.
On the other hand, ConvGNNs generalize the convolution operation from grid data to graph data and can stack multiple graph convolutional layers to extract high-level node representations. ConvGNNs play a central role in building up many other complex GNNs, which can be further categorized into two subtypes: 1) Spectral-based: ChebNet \cite{defferrard2016convolutional}, graph convolutional network (GraphConv) \cite{kipf2016semi} and 2) Spatial-based: message passing neural networks (MPNN) \cite{gilmer2017neural}, GraphSAGE \cite{hamilton2017inductive}, graph attention network (GAT) \cite{velivckovic2017graph}, graph isomorphism network (GIN) \cite{xu2018powerful}. 

\begin{figure*}[ht]
\centering
  \includegraphics[width=0.9\linewidth]{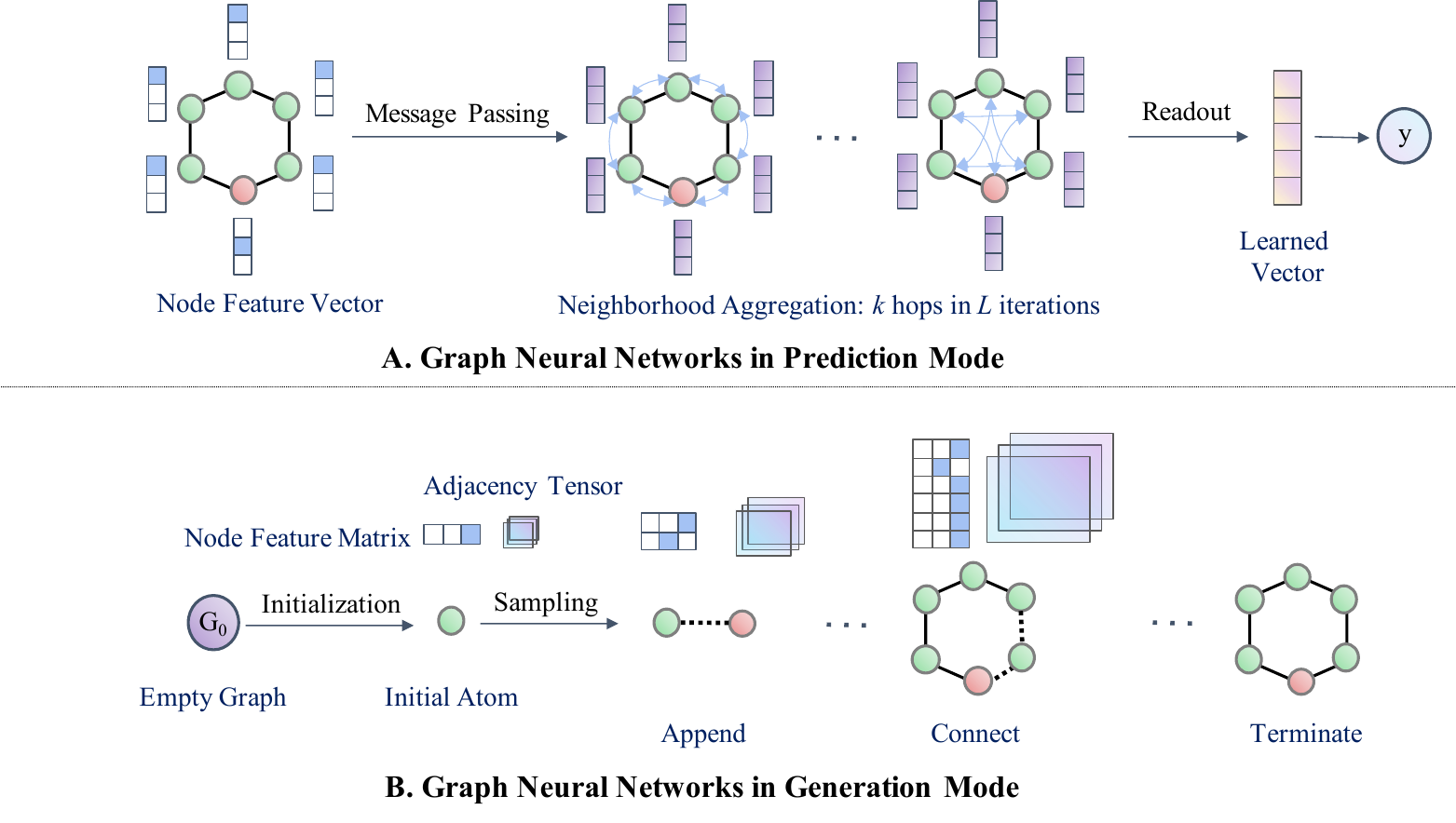}
  \caption{Illustrations of Graph Neural Networks.}
  \label{fgr:archs_gnn}
\end{figure*}

In drug discovery, GNNs are often exploited for molecular property prediction (Fig~\ref{fgr:archs_gnn}A).
For instance, Kearnes et al \cite{kearnes2016molecular} developed Weave to perform graph convolutions on molecular graphs for representation learning, where the graph convolutions, nonetheless, did not outperform the fingerprint-based models back in 2016.
Later, Gilmer et al \cite{gilmer2017neural} proposed MPNN as a unified framework for quantum chemical properties prediction. MPNN has two phases in the forward pass, namely, message passing and readout. During message passing, for each atom, feature vectors from its neighbors are propagated into a message vector wherein the hidden vector for the atom is updated by the message vector. A readout function is used to aggregate the feature vectors into a graph feature vector, which is then passed to a fully connected layer for downstream predictions.
Yang et al \cite{yang2019analyzing} then expanded MPNN into directed MPNN (D-MPNN), which uses messages associated with directed edges (bonds) instead of nodes (atoms) used in MPNN, thereby preventing repeated message passing from the same node. 
Notably, Yang et al \cite{yang2019analyzing} also introduced a practice to concatenate the 200 global molecular features calculated by RDKit \cite{Landrum2016RDKit2016_09_4} with the learned features by D-MPNN for downstream predictions, also adopted in later works \cite{rong2020grover}. 
Xiong et al \cite{xiong2019pushing} integrated graph attention mechanism into GNNs and developed Attentive FP, which is able capture topologically adjacent atoms' interactions for improved molecular property prediction. 
More recently, Withnall et al \cite{withnall2020building} also made augmentations to the MPNN and proposed attention MPNN (AMPNN) and edge memory neural network (EMNN) for physicochemical property prediction. 
So far, GNNs have been widely applied for molecular property prediction. More examples include as SchNet \cite{schutt2017schnet}, PotentialNet \cite{feinberg2018potentialnet}, and DimeNet \cite{klicpera2020directional}, among others \cite{altae2017low, mayr2018large, liu2018n, lu2019molecular, cai2019deep, wang2019molecule, hu2019strategies,  hao2020asgn, nguyen2020meta}. 
Moreover, subgraphs can directly map to molecular substructures, which also improves interpretability \cite{li2019deepchemstable, xiong2019pushing, tang2020self, pathak2020chemically}. 

Partly encouraged by the superior performance of GNNs for molecular property prediction, GNNs are also exploited for molecule generation (Fig~\ref{fgr:archs_gnn}B). 
As mentioned above, RNNs can be used to generate molecular graphs, which, nevertheless, needs to store a full trajectory for the graph generation process and tends to forget initial states \cite{you2018graph}.
In 2018, Li et al \cite{li2018multi} developed a conditional graph generator, MolMP, which does not involve atom-level recurrent units. MolMP models the graph generation as a MDP problem, where the action to grow graph only depends on its current state. There are three actions in total - append, connect and terminate, the sampling process of which is parameterized by a neural network.
Experimental results show that MolMP outperforms SMILES-based molecule generation in a variety of evaluation metrics, especially the validity. GNNs-based molecule generation can be used in common drug design applications such as designing molecules with certain scaffolds, presumably due to more straightforward mapping to chemical substructure with the graph representation.
Furthermore, since molecule generation is usually driven by certain desired properties, reinforcement learning (see Section~\ref{sec4} ``Learning Paradigms"), therefore, is often integrated with GNNs for goal-directed drug design. Examples include GCPN \cite{you2018graph}, MolDQN \cite{zhou2019optimization}, DeepGraphMolGen \cite{khemchandani2020deepgraphmolgen}, and MNCE-RL \cite{xu2020reinforced}.
For more practical issues on GNNs for molecule generation, such as generation schemes (single-shot vs iterative) and computation, we refer the readers to the guide by Mercado et al \cite{mercado2020practical}.

\subsection{\textbf{Variational Autoencoders}}\label{sec3_4}
\begin{figure}[ht]
\centering
  \includegraphics[width=1\linewidth]{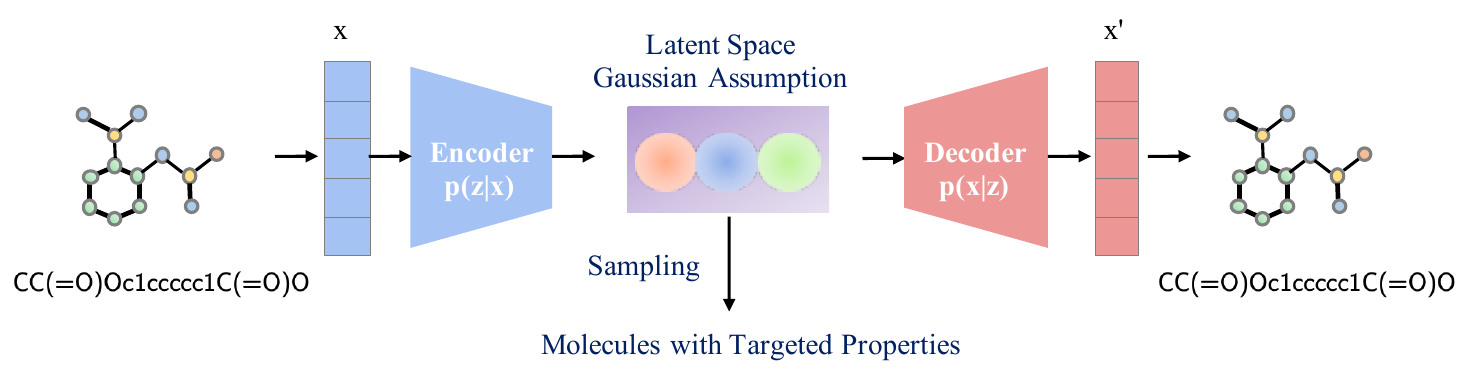}
  \caption{Illustration of Variational Autoencoders.}
  \label{fgr:archs_vae}
\end{figure}

Variational autoencoders (VAEs), a class of powerful probablistic generative models, were first introduced by Kingma et al \cite{kingma2013auto} in 2013. VAEs, consisting of an encoder $E$ and a decoder $D$. The encoder maps high-dimensional data into a low-dimensional, continuous latent space (Fig~\ref{fgr:archs_vae}). Compared to common autoencoders, the latent space is regularized to be organized, ideally, through the KL divergence. In addition to reconstruction, VAEs approximate a probability distribution, which can be sampled for generation purpose. 
Thus, given input $x$, the parameters of VAEs are optimized by minimizing the reconstruction loss and the KL divergence \cite{kingma2019introduction}:
\begin{equation}\label{eq:elbo}
    ||x - D(E(x))||^2 + KL\big(N(\mu_x, \sigma_x), N(0,1)\big),
\end{equation}
which is equivalent to maximize the evidence lower bound (ELBO). In Equation~\ref{eq:elbo}, $N(0,1)$ denotes the unit normal distribution; $\mu_x$ and $\sigma_x$ are learnable parameters, representing mean and variance of a Gaussian distribution. 

VAEs can be applied on SMILES strings for molecule generation. 
For example, Gómez-Bombarelli et al \cite{gomez2018automatic} developed a VAE model for automatic molecule design, where a pair of deep neural networks (i.e. an encoder and a decoder) is trained as an autoencoder to convert the input SMILES strings into a continuous vector representation.  
To train the autoencoder, a reconstruction loss is adopted in attempt to reproduce the original SMILES string. However, the ultimate goal is not to merely reconstruct the input. Rather, the autoencoder aims to learn a compact representation for the molecules. 
Thus, a constraint is applied in the autoencoder by jointly training a physical property regression model to organize the VAE's latent space subjected to the property value, which can be used to sample molecules towards the desired property value. 
Partly due to the syntax of SMILES, the latent space learned by the autoencoder can be sparse and contain large ``dead areas", which correspond to invalid molecules.
Therefore, VAEs with a focus on the syntax for valid molecule generation are proposed later, such as GrammarVAE \cite{kusner2017grammar} and syntax-directed VAE \cite{dai2018syntax}.
Other related works also include semi-supervised VAE (SSVAE) for continuous output \cite{kang2018conditional}, conditional VAE (CVAE) \cite{lim2018molecular}, 
constrained graph VAE (CGVAE) \cite{liu2018constrained}, NeVAE \cite{samanta2020nevae}, GTM VAE \cite{sattarov2019novo}
and CogMol \cite{chenthamarakshan2020target}. 
A variant of the VAE is the adversarial autoencoder (AAE) \cite{makhzani2015adversarial}, which replaces the KL divergence with an adversarial objective. More specifically, the Gaussian distribution assumption as a prior on the latent representations for KL-divergence computation is replaced by other priors, i.e., an additional discriminator is added to force the encoder generates latent representations in a specific distribution (e.g. a uniform distribution). AAEs can also be used for molecule generation \cite{kadurin2017drugan, blaschke2018application, polykovskiy2018entangled}, which can improve reconstruction and the validity of generated molecules.

VAEs can also be applied on the graph representations for generation purpose. In 2018, Simonovsky et al \cite{simonovsky2018graphvae} proposed a VAE framework (GraphVAE) to generate molecular graphs.Their main idea is to output a probabilistic fully-connected graph and use a standard graph matching algorithm to align it to the ground truth.
Jin et al \cite{jin2018junction} developed the junction tree VAE (JT-VAE). In JT-VAE, a molecular graph is first mapped into a junction tree via a tree decomposition algorithm and the junction tree then undergoes the VAE's encoding-decoding process. The learned latent space of the junction tree can be used to search for substructures, which then assemble into molecules with specific properties. A prominent merit of JT-VAE is that the validity of all generated molecules can be guaranteed.
Ma et al \cite{ma2018constrained} also proposed a regularization framework for VAEs (Regularized VAE) that regularize the output distribution of the decoder, thereby improving the validity. 
Later, Kajino et al \cite{kajino2019molecular} developed molecular hypergraph grammar VAE (MHG-VAE), where a molecular graph is described as a hypergraph and the grammar VAE \cite{kusner2017grammar} is trained by inputting the grammar for sequence production of the hypergraph. 
In 2019, Kwon et al \cite{kwon2019efficient} developed a non-autoregressive graph VAE and incorporated three additional learning objectives into the model, namely, approximate graph matching, reinforcement learning, and auxiliary property prediction, which is able to generate valid and diverse molecular graphs with various constraints. 
Moreover, graph-based VAEs can also embrace a strategy for drug design with the ability to retain a particular scaffold (i.e., substructure), such as the ScaffoldVAE \cite{lim2019scaffold}. 

One drawback of VAEs, nonetheless, is that the set of substructures by partitioning molecules can be quite large. Consequently, the iterative prediction of which substructure to add can be inaccurate, especially for infrequent substructures. To address this challenge, Fu et al \cite{fu2020core} proposed a novel strategy, CORE, by combining scaffolding tree generation and adversarial training.
Besides, the computational cost increasing with the number of nodes in a graph is another major challenge here, limiting the application to larger molecules.
In 2020, Kwon et al \cite{kwon2020compressed} proposed a compressed graph representation to alleviate computational complexity while maintaining the validity and diversity of generated molecules.
More recently, Jin et al \cite{jin2020hierarchical} developed hierarchical graph VAE (HierVAE) which can employ larger and more flexible graph motifs as building blocks for molecules. More specifically, the encoder produces a multi-resolution representation for each molecule in a fine-to-coarse fashion, from atoms to connected motifs while the autoregressive coarse-to-fine decoder adds one motif at a time. Notably, HierVAE can even be used to generate polymers.

\subsection{\textbf{Generative Adversarial Networks}}\label{sec3_5}
\begin{figure}[ht]
\centering
  \includegraphics[width=1\linewidth]{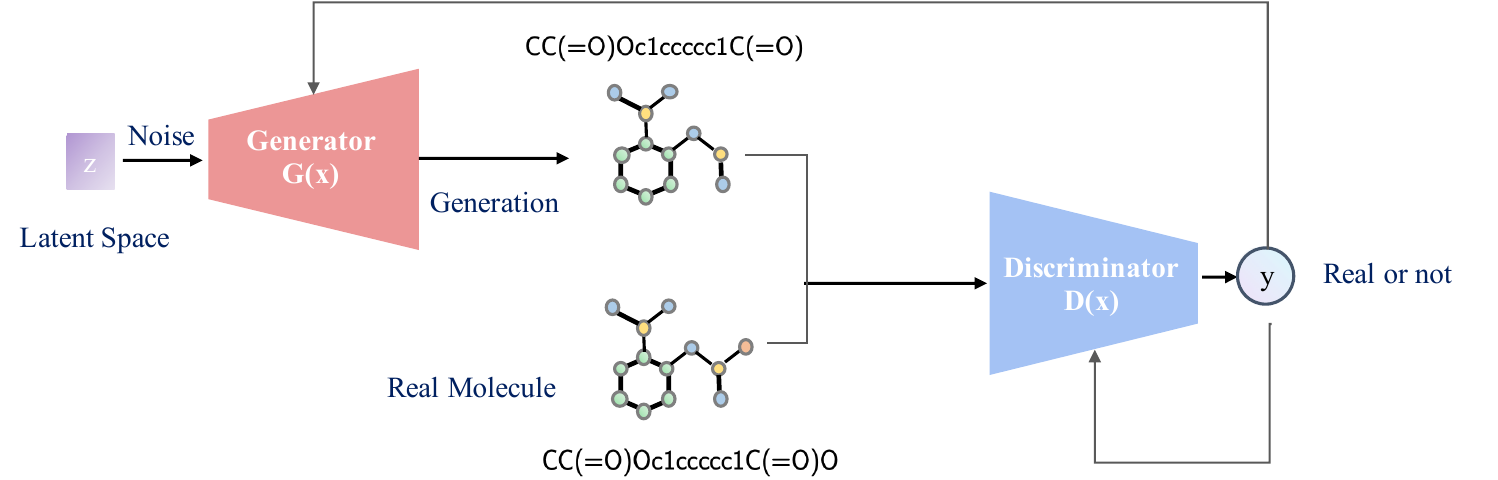}
  \caption{Illustration of Generative Adversarial Networks.}
  \label{fgr:archs_gan}
\end{figure}

Generative Adversarial Networks (GANs), developed by Goodfellow et al \cite{goodfellow2014generative} in 2014, have made remarkable achievements in generating realistic synthetic samples. GANs consist of a generative model $G$, and a discriminative model $D$ (Fig~\ref{fgr:archs_gan}). The generator aims to generate new data points from a random distribution whereas the discriminator aims to classify whether the generated samples are from the training data distribution or from the generator. 
GANs can be trained by the min-max loss, which alternatively optimizes the generator and the discriminator using a min-max objective: 

\begin{equation}\label{eq:minimax}
    \min_G\max_D \mathcal{L}(G, D) = \mathbb{E}_{x\sim p_x} [\log(D(x))] + \mathbb{E}_{z\sim p_z} [\log(1-D(G(z)))],
\end{equation}
where $p_x$ and $p_z$ denote the distribution of the real data $x$ and the noise prior $z$.

GANs can be applied to SMILES strings for molecule generation.
In 2017, Guimaraes et al \cite{guimaraes2017objective} developed objective-reinforced GANs (ORGAN), built upon SeqGAN \cite{yu2017seqgan}, to generate molecules in SMILES strings while also optimizing several domain-specific metrics. The generator is based on LSTM, which is modeled as a stochastic policy in a reinforcement learning setting (more details in Section~\ref{sec4} ``Learning Paradigms"), whereas the Wasserstein loss is used train the discriminator (a CNN model). 
Experimental results showed that the generated molecules exhibit drug-like structures as well as improvement in the evaluation metrics. 
Shortly after, an objective-reinforced GANs for inverse-design chemistry (ORGANIC) \cite{sanchez2017optimizing} was developed based upon ORGAN. ORGANIC can generate molecules with biased distribution towards certain attributes for both drug discovery and material design.
In 2018, Putin et al \cite{putin2018reinforced} presented a reinforced adversarial neural computer (RANC) framework, which also combines GANs and RL. The generator of the RANC framework is a differentiable neural computer (DNC) with an explicit memory bank, instead of the LSTM model in ORGANIC and ORGAN. This is because the generation of discrete data using RNNs, particularly, LSTM with maximum likelihood estimation, can suffer from the so-called ``exposure bias", i.e., missing salient features of the data. RANC outperforms ORGANIC, as measured by several metrics: number of unique structures, passing medicinal chemistry filters (MCFs), Muegge criteria and high QED scores. RANC is able to generate molecules that match the distributions of the key chemical features/descriptors (e.g., MW, logP) and lengths of the SMILES strings from the training set. 

GANs can also be applied on molecular graphs for molecule generation.
In 2018, De Cao et al \cite{de2018molgan} developed MolGAN, an implicit, likelihood-free generative model for small molecular graph generation, which circumvents the expensive graph matching procedures \cite{simonovsky2018graphvae}. Moreover, they adapted the GANs to enable direct operation on molecular graphs. RL is also integrated to encourage the generation of molecules towards desired properties.
Experimental results on the QM9 dataset showed that MolGAN is able to generate nearly 100\% valid molecules, which outperforms ORGAN in validity.
One drawback of MolGAN is its susceptibility to mode collapse, i.e., repeated samples being generated multiple times, leading to low uniqueness \cite{salimans2016improved}.

\subsection{\textbf{Normalizing Flow Models}}\label{sec3_6}
\begin{figure}[ht]
\centering
  \includegraphics[width=1\linewidth]{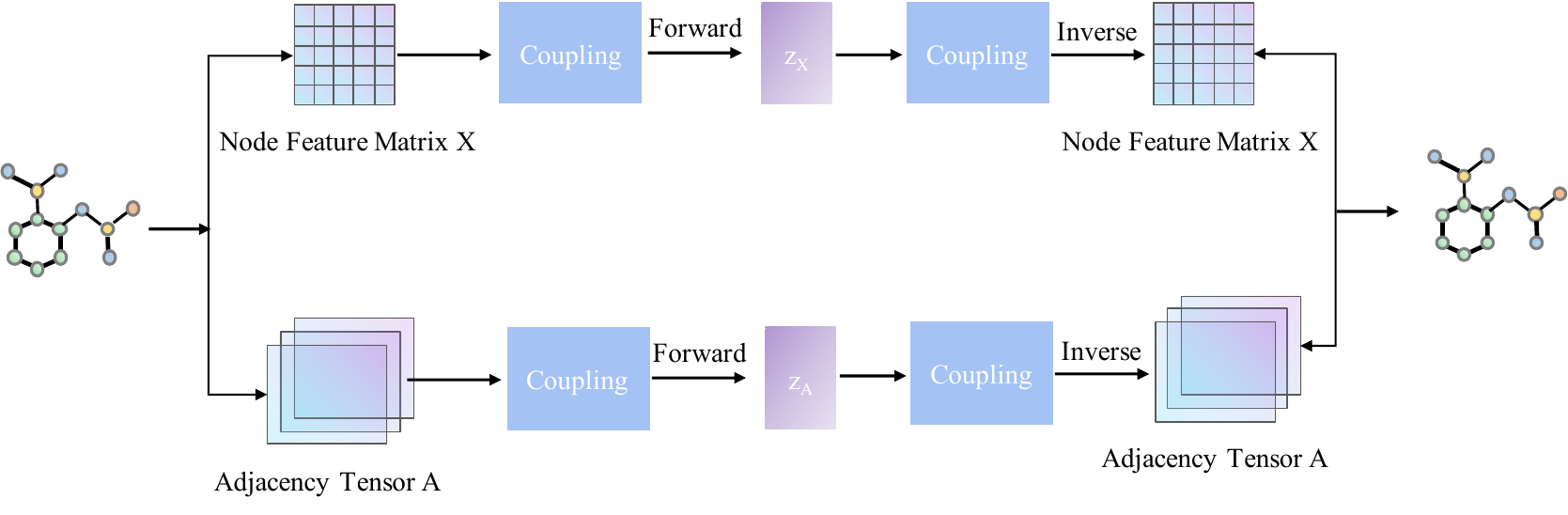}
  \caption{Illustrations of Flow Models.}
  \label{fgr:archs_flow}
\end{figure}

In addition to the RNNs, VAEs and GANs, another major class of generative models is the normalizing flow models \cite{kobyzev2020normalizing}. 
Representative works include the Non-linear Independent Component Estimation model (NICE) \cite{dinh2014nice}, the Real-valued Non-Volume Preserving model (RealNVP) \cite{dinh2016density}, and the Glow model, among others.
In NICE, Dinh et al \cite{dinh2014nice} introduced tractable calculation for reversible transformations, which are also known as the affine coupling layers underlying the flow models.
The basic idea of flow models is to learn an invertible mapping between complex distributions and simple prior distributions (Fig~\ref{fgr:archs_flow}). 
By exploiting exact and tractable likelihood estimation for training, flow models enable efficient one-shot inference and 100\% reconstruction of the training data \cite{mercado2020practical}. 

In drug discovery, flow models have been applied to generate molecules, mainly on molecular graphs.
In 2019, Madhawa et al \cite{madhawa2019graphnvp} developed GraphNVP, the first flow model for molecular graph generation. In GraphNVP, the graph generation is decomposed into two steps, i.e., generation of an adjacency tensor and generation of node attributes, which yields the exact likelihood maximization on the graph with two reversible flows. GraphNVP is able to generate valid molecules with minimal duplicates. The learned latent space can be further exploited to generate molecules with desired properties.
Honda et al \cite{honda2019graph} also developed an invertible flow model for molecular graph generation based on residual flows, also known as graph residual flow (GRF), which enables more flexible and complex non-linear mappings than the traditional coupling flows. Experimental results showed that GRF can achieve comparable performance with GraphNVP, while having much less parameters to learn.
Notably, GraphNVP \cite{madhawa2019graphnvp} and GRF \cite{honda2019graph} generate molecular graphs in a single-shot manner \cite{mercado2020practical}, which may lead to low validity, nevertheless.
Consequently, a sequential iterative graph generation manner is proposed for flow models. For example, GraphAF \cite{shi2020graphaf}, an autoregressive flow-based model, adopts an iterative sampling process to leverage chemical domain knowledge, such as valency checking in each step. With the integration of chemical rules, GraphAF is able to generate molecules of 100\% validity. Moreover, its training process is significantly faster than GCPN \cite{you2018graph}. GraphAF can be further finetuned with RL, which achieves better performance on molecular property optimization compared to JT-VAE \cite{jin2018junction} and GCPN \cite{you2018graph}.
MoFlow, later developed by Zang et al \cite{zang2020moflow}, applies a validity correction to the generated graph, which not only enables efficient molecular graph generation in a single-shot manner, but also guarantees the chemical validity. The continuous latent space learned via encoding the molecular graphs can be further used to generate novel and optimized molecules during the decoding process towards desired properties. 
More recently, Luo et al \cite{luo2021graphdf} developed GraphDF, which, on the contrary, aims to learn a discrete latent representation with the flow models and capture the original discrete distribution of the discrete graph structures without adding real-valued noise. For molecule generation, GraphDF sequentially samples the discrete latent variables and maps them to new nodes and edges via invertible transforms. The discrete transforms can circumvent the cost of computation while also achieving state-of-the-art performance in random molecule generation, property optimization and constrained optimization tasks. 

For the flow-based generative models, the most prominent feature is that they are able to exactly reconstruct all the input data without duplicates due to the precise likelihood maximization, which can be an important complement for molecule generation. In particular, when the molecular property is highly sensitive to minor structural changes, i.e., activity cliffs \cite{maggiora2006outliers, stumpfe2014recent}, a replacement of a specific atom (node) might be needed. In other words, flow models can offer more precise modifications on existing molecular structures.

\subsection{\textbf{Transformers}}\label{sec3_7}
RNNs have been widely applied to handle sequential input. However, RNNs can suffer from the gradients explosion or the vanishing problem \cite{lecun2015deep}. 
In 2017, a seminal work ``Attention is all you need" proposed a novel transformer architecture, built with the self-attention mechanism \cite{vaswani2017attention}. Transformers have now become the \textit{de facto} standard in powerful language models, such as GPT \cite{radford2018improving}, 
BERT \cite{devlin2018bert}, 
GPT-2 \cite{radford2019language}, 
RoBERTa \cite{liu2019roberta}, 
and GPT-3 \cite{brown2020language}, and even in advanced computer vision models, such as DETR \cite{carion2020end} and Vision Transformer \cite{dosovitskiy2020image}.
Unlike RNNs, transformers forfeit recurrent connections. By adopting positional embedding, transformers are even better at dealing with long sequences \cite{vaswani2017attention}.

\begin{figure}[ht]
\centering
  \includegraphics[width=1.0\linewidth]{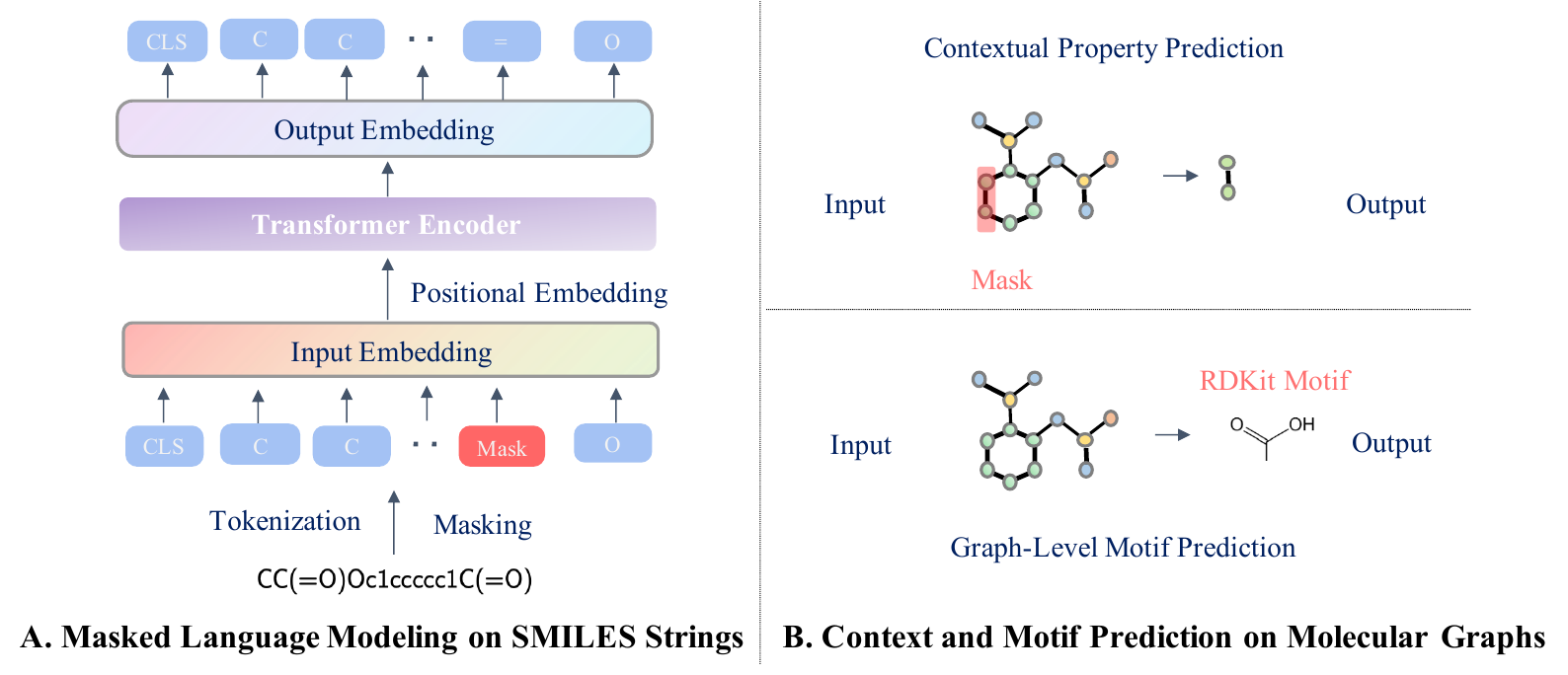}
  \caption{Illustrations of Self-Supervised Learning with Transformers.}
  \label{fgr:archs_transformers}
\end{figure}

Not unexpectedly, transformers are being actively applied in drug discovery. Notably, transformers enable effective self-supervised pretraining, such as masked language modeling (Fig~\ref{fgr:archs_transformers}A). In 2019, Wang et al \cite{wang2019smiles} developed SMILES-BERT, which consists of several transformer encoder layers, to improve molecular property prediction. SMILES-BERT is first pretrained on a large-scale corpus of SMILES strings via a SMILES recovery task and then fine-tuned on the downstream prediction tasks.
Later, Honda et al \cite{honda2019smiles} proposed to learn molecular representations through pretraining a sequence-to-sequence language model, which is termed as the SMILES Transformer. Chithrananda et al \cite{chithrananda2020chemberta} also developed ChemBERTa, built upon the RoBERTa model \cite{liu2019roberta} for molecular property prediction.
More recently, Fabian et al \cite{fabian2020molecular} applied the architecture of BERT \cite{devlin2018bert} to learn molecular representations, also referred to as MolBERT. When pretrained with masked language modeling and other tasks, MolBERT achieves improved performance for molecular property prediction compared to the fixed fingerprints.
Moreover, transformers can also be applied on molecular graphs, especially considering that the transformer encoder can be viewed as a GAT variant \cite{velivckovic2017graph}. For example, Rong et al \cite{rong2020grover} developed a novel framework, GROVER, to learn graph representations with the message passing transformer. 
By designing self-supervised contextual property prediction and graph-level motif prediction tasks (Fig~\ref{fgr:archs_transformers}B), GROVER is pretrained on 10 million unlabeled molecules and achieves state-of-the-art performance on 11 benchmark datasets. 

In addition to molecular property prediction, transformers can also be exploited for molecule generation, such as MoleculeChef \cite{bradshaw2019model}, which can generate the reactants for a given product, similar to machine translation.
More recently, transformers are also exploited for protein-specific molecule generation \cite{grechishnikova2021transformer}, where the input is the amino acid sequence of the target protein and the output are ligands in the SMILES representation.

\section{Learning Paradigms}\label{sec4}
Drug discovery, despite the light shed by AI, still faces major challenges.
For molecular property prediction, labeled data points are at the core of machine learning models. Nonetheless, in real-world settings, generating labeled data points in wet lab can be very expensive. Consequently, the datasets for model training are usually limited in size, exhibit high sparsity, and can be heavily biased and noisy, which is also termed as the low-data drug discovery problem \cite{altae2017low, nguyen2020meta}.
For molecule generation, although existing generative models, such as VAEs, can be used to generate molecules towards desired properties, the mechanism by mapping from the points in the latent space to real molecules which are most proximal can limit the exploration of the chemical space, leading to low novelty and diversity \cite{gomez2018automatic}. 
To address these challenges, various learning paradigms have been proposed. 
In this survey, we mainly focus on self-supervised learning and reinforcement learning to address molecular property prediction and molecule generation, respectively.
Other learning paradigms are also discussed. 

\subsection{\textbf{Self-Supervised Learning}}\label{sec4_1}
The performance of deep neural networks, especially supervised learning, hinges on a large labeled dataset. Nevertheless, supervised learning is meeting its bottleneck due to its heavy reliance on expensive manually-labeled data \cite{liu2020self}.
In real-world problems such as molecular property prediction, the labeled data is often limited, sparse and biased, which leads to low generalizalibility of models. 
Self-supervised learning is promising paradigm and has achieved state-of-the-art performance in learning with limited labels,
as adopted in the aforementioned language models, for instance, BERT \cite{devlin2018bert}.
Notably, self-supervised learning should be distinguished from unsupervised learning. Unsupervised learning focuses on detecting patterns in data without labels, such as clustering, whereas self-supervised learning aims to recover the data. More specifically, it can be classified into two main types, i.e., generative and contrastive self-supervised learning. 

For the generative self-supervised learning, a canonical task is the masked language modeling, as proposed in BERT \cite{devlin2018bert}, where the model is trained to predict the masked tokens, thereby recovering the original input. Model parameterization is usually implemented by optimizing the cross-entropy loss between the output and the masked tokens in the input. 
Representative works of self-supervised learning in drug discovery are discussed in Section~\ref{sec3_7} ``Transformers" (e.g., MolBERT \cite{fabian2020molecular} and GROVER \cite{rong2020grover}). Notably, self-supervised pretraining can avoid the negative transfer caused by supervised pretraining - transfer of knowledge from pretraining harms model generalization, as shown by Hu et al \cite{hu2019strategies}. 
Besides, contrastive learning is another type of self-supervised learning. More specifically, contrastive learning aims to learn latent representations through contrasting data pairs (positive vs negative), where the positive and negative examples are constructed by augmenting the unlabeled samples in a self-supervised manner. 
Recently, contrastive learning has been employed to address the low-data drug discovery problem. For instance, Wang et al \cite{wang2021molclr} proposed molecular contrastive learning of representations (MolCLR) on molecular graphs for molecular property prediction. 
Three molecular graph augmentation ways are used, i.e., atom masking, bond deletion, and subgraph removal.
Through a contrastive loss, MolCLR learns molecular representations by contrasting positive vs negative molecules, where molecular graph pairs augmented from the same molecule are treated as the positive and the others denoted as negative. 
Experimental results show that MolCLR can effectively transfer the learned representations to downstream tasks and achieve state-of-the-art performance in molecular property prediction. 

In addition to self-supervised learning, other learning paradigms have also been exploited to address the low-data drug discovery challenge.
For example, \textit{meta learning} \cite{vanschoren2018meta} aims to learn a learner to be adapted to new tasks. In a study by Nguyen et al \cite{nguyen2020meta}, the meta-learning initializations outperform multi-task pretraining baselines on 16 out of 20 in-distribution tasks and all out-of-distribution tasks.
A member of the meta-learning family is \textit{few-shot learning} \cite{wang2020generalizing}, the core idea of which is to generalize with a few examples. For instance, Altae Tran et al \cite{altae2017low} proposed a one-shot learning framework for activity classification, which lowers the amount of data required for predictions. Intuitively, through learning a distance metric, molecules can be embedded into the latent space in a more organized way. Thus, when new molecules come, their embeddings can be compared to the exiting labeled molecules for more accurate prediction.
Closely to this idea, another paradigm is \textit{metric learning} \cite{kulis2012metric, yang2021hierarchical}, which mainly deals with data with mixed distribution due to the activity cliffs \cite{bajorath2019duality}. 
Metric learning has been widely applied in computer vision \cite{movshovitz2017no, yang2021hierarchical}, especially in situations where the similarity or distance must be computed for clustering or nearest neighbor classification purpose. 
At its core, a distance metric (e.g., cosine distance) is to be learned, based on which the learned latent representations of the input data can be separated according to their labels. In a recent work by Na et al \cite{na2020machine}, a generalized deep metric learning (GeDML) framework is proposed, which alleviates the structure-property mismatch problem through better separating molecules in the latent space. The representations learned via metric learning are also conducive for goal-directed molecule generation, i.e., search in the chemical space. For example, Koge et al \cite{koge2020embedding} proposed a molecular embedding framework by combining VAEs and metric learning. The idea is to make the molecules' embedding in the latent space consistent with their properties, thereby enabling efficient search during molecule generation.

\subsection{\textbf{Reinforcement Learning}}\label{sec4_2}
With improved performance on molecular property prediction, another challenge still poses for molecule generation, i.e., how to design molecules with the desired properties?
As mentioned in Section~\ref{sec3} ``Model Architectures", VAEs and flow models can be used to generate molecules with preferred properties by sampling from a learned latent space. However, the latent space can be highly dimensional and the objective functions defined in the latent space is usually non-convex, making it difficult to optimize the properties of generated molecules \cite{zhou2019optimization}.
Consequently, reinforcement learning (RL) is often used as the alternative to navigate the chemical space, which mainly deals with how an agent should take actions in a certain state so as to maximize a reward or return \cite{sutton2018reinforcement}. 
RL algorithms can be classified into 1) value-based (e.g., Q-learning), 2) policy-based (e.g., policy gradient) and 3) hybrid (e.g., actor-critic) \cite{arulkumaran2017brief}. 

In drug discovery, the DMTA cycle (see Section~\ref{sec1_1} ``Drug Discovery Overview") underlying drug design \cite{schneider2018automating}, i.e., goal-directed molecule generation, can be potentially automated by RL through connecting the generative model (i.e., the agent for molecule generation) and the predictive model (i.e., assigning rewards based on the predicted property values).
For example, Zhou et al \cite{zhou2019optimization} adopted value-based, double Q-learning (DQN) \cite{van2016deep} to optimize the generated molecules.
Other related works mainly adopt the policy-gradient algorithm, REINFORCE \cite{williams1992simple}, as an estimator of the gradient, such as ORGAN \cite{guimaraes2017objective}, REINVENT \cite{olivecrona2017molecular}, and ORGANIC \cite{sanchez2017optimizing}. 
Another policy-gradient algorithm, proximal policy optimization (PPO) \cite{schulman2017proximal}, is also gaining popularity recently, which is improved from the trust region policy optimisation (TRPO) \cite{schulman2015trust}. 
TRPO employs a trust region so that optimizations are restricted to a region where the approximation of the true cost function holds, thereby preventing policies updated too wildly and lowering the chance of a catastrophically ``bad" update \cite{arulkumaran2017brief}. However, TRPO requires the calculation of second-order gradients, being computationally expensive.
PPO, on the contrary, only requires first-order gradients and can retain the performance of TRPO, exhibiting low sample complexity. 
Studies adopting PPO for \textit{de novo} drug design include the work by Neil et al \cite{neil2018exploring}, GCPN \cite{you2018graph}, DeepGraphMolGen \cite{khemchandani2020deepgraphmolgen} and MNCE-RL \cite{xu2020reinforced}.

Nevertheless, policy-gradient algorithms usually exhibit high variance since the gradient estimation can be noisy \cite{arulkumaran2017brief}.
To reduce the variance, an improved class of algorithms is the hybrid actor-critic method, which combines policy-gradient methods with learned value functions.
For example, off-policy deterministic policy gradient (DPG) extends the standard policy gradients for stochastic policies to deterministic policies, which only integrates over the state space instead of both state and action spaces, thus requiring fewer samples in problems with large action spaces. Later, deep deterministic policy gradient (DDPG) utilizes neural networks on high-dimensional space is introduced, as adopted in MolGAN \cite{de2018molgan}. 
Another technique for variance minimization to accelerate convergence is to subtract the estimated reward from the true reward, which separates the policy training from value estimation, also known as the advantage actor-critic (A2C) algorithm adopted by Neil et al \cite{neil2018exploring}.

With RL training, chemical libraries shifted towards desired properties are expected to be generated. However, drug design is a multi-objective optimization problem \cite{schneider2020rethinking, deng2020towards}. In order to prioritize molecules based on the pre-defined goals, non-dominated sorting or pair-based comparisons are usually exploited to find solutions with the Pareto optimality \cite{yasonik2020multiobjective, domenico2020novo, liu2021drugex}.
Another issue with RL is the trade-off between exploration and exploitation. As illustrated by Zhou et al \cite{zhou2019optimization}, this trade-off is a dilemma underlying by uncertainty. Due to the lack of a complete knowledge of the rewards for all the states, if constantly choosing the best action known to produce the highest reward (exploitation), the model will never learn anything about the rewards of the other states; on the other hand, if always choosing a random action (exploration), the model will not receive sufficient reward. 
A potential solution for the exploration-exploitation trade-off is \textit{active learning}, which is a paradigm where the model can query an expert or any other information sources in an active manner during learning \cite{reker2015active, jimenez2020drug}.  

\section{Discussions}\label{sec5} 
There has been a surge of AI in drug discovery over the past decade, which is still gaining popularity. Nonetheless, there are still challenges to be addressed.
Despite the prosperity of deep learning models, it should be emphasized that data is at the core of developing and evaluating the models  \cite{walters2020assessing, sambasivan2021everyone}. 
To make the models (either predictive or generative) more useful, data must be in sufficient amount and should maintain high quality.
However, a fact to our dismay is that although existing chemical libraries have a large amount of molecules, the number of data points for each specific assay can be very scarce \cite{walters2020assessing}.
Sometimes, even the quality of benchmark datasets is questionable with regard to the representative power for real-world drug discovery imposed by the vast chemical space \cite{walters2021critical}.
Datasets in drug discovery can be highly imbalanced \cite{korkmaz2020deep}. Thus, when evaluating the models, there is need to obtain appropriate datasets and also consider data balancing methods as well as proper evaluation metrics (e.g., AUPRC vs AUROC) \cite{schneider2020rethinking}. 

Besides, for the DMTA cycle (see Section~\ref{sec1_1} ``Drug Discovery Overview") in drug design, it should always be driven by a need or certain hypotheses \cite{schneider2020rethinking}. 
Even equipped with perfect predictive and generative models, a question still remains, i.e., what are the hypotheses for designing a drug candidate? 
In other words, what are the desired properties underlying an ideal drug candidate?
To generate the insights into drug design, real-world data (e.g., electronic health records (EHR) and marketed drug databases) \cite{singh2018real} is receiving substantial attention for understanding the effectiveness and side effects of different therapeutics. 
Recently, we mined a large-scale EHR database for the innate properties underlying opioid analgesics with reduced overdose effects \cite{deng2021large}. 
We also mined the DrugBank database to identify key pharmacological components (i.e., carriers, transporters, enzymes and targets) underlying drug-drug interactions (DDIs) \cite{deng2020informatics}. These patterns emerging from real-world data (RWD) allows hypotheses generation and can calibrate drug design insights \cite{singh2018real}. 

Another challenge is that deep learning, despite its superior performance, still leaves the model elusive for human interpretation.
An ongoing need is, therefore, to develop explainable models with high interpretability.
More specifically, there are four aspects to cover \cite{jimenez2020drug}: 
1) Transparency, which is knowing how the system reaches a particular answer;
2) Justification, which is elucidating why the answer provided by the model is acceptable;
3) Informativeness, which is providing new information to human decision makers; and
4) Uncertainty estimation, which is quantifying how reliable a prediction is. 
An ideal state is that AI can allow scientists to hone their knowledge and beliefs on the investigated process. 
For more details on explainable AI in drug discovery, we refer the readers to the review by Jimenez-Luna et al \cite{jimenez2020drug}.

In addition to the scientific challenges, technical concerns remain. One unignorable reality is that, even for the state-of-the-art representation learning on molecular graphs, fixed fingerprints can still outperform GNN-derived representations for molecular property prediction \cite{jiang2021could}.
In fact, ECFPs are a component of some GNN models \cite{yang2019analyzing, rong2020grover}. 
Besides, there is a lack of a unified protocol for AI-driven drug discovery studies. For example, different benchmark datasets, different split folds and evaluation metrics are used across the studies for molecular property prediction, let alone the varying hyper-parameters tuning, training and evaluation procedure \cite{jiang2021could}.
For molecule generation, Walters et al \cite{walters2020assessing} already proposed a few guidelines to evaluate the novelty of AI-discovered molecules. Likewise, protocols for molecular property prediction are also needed.

Overall, there are many promising opportunities as well as significant challenges when applying AI in drug discovery. In order to launch successful applications, we need to understand the basic concepts and consider the task, the data, the molecule representation, the model architecture and the learning paradigm as a whole. 
In this survey, we have covered multiple aspects centered around AI-driven drug discovery. 
We envision that with these aspects well understood, more meaningful contributions will be made to substantially transform this field.

\section{Author contributions statement}
J.D. and Z.Y. conceived the manuscript and the GitHub repository. J.D. drafted the manuscript and built the GitHub repository. All authors made critical revisions and review on the manuscript. 

\begin{acknowledgement}
This project is partially funded by a Stony Brook University OVPR Seed Grant. The neural networks templates are from Visuals by dair.ai (https://github.com/dair-ai/ml-visuals).  
\end{acknowledgement}

\begin{suppinfo}

https://github.com/dengjianyuan/Survey\_AI\_Drug\_Discovery

\end{suppinfo}

\bibliography{achemso-demo}

\providecommand{\latin}[1]{#1}
\makeatletter
\providecommand{\doi}
  {\begingroup\let\do\@makeother\dospecials
  \catcode`\{=1 \catcode`\}=2 \doi@aux}
\providecommand{\doi@aux}[1]{\endgroup\texttt{#1}}
\makeatother
\providecommand*\mcitethebibliography{\thebibliography}
\csname @ifundefined\endcsname{endmcitethebibliography}
  {\let\endmcitethebibliography\endthebibliography}{}
\begin{mcitethebibliography}{245}
\providecommand*\natexlab[1]{#1}
\providecommand*\mciteSetBstSublistMode[1]{}
\providecommand*\mciteSetBstMaxWidthForm[2]{}
\providecommand*\mciteBstWouldAddEndPuncttrue
  {\def\EndOfBibitem{\unskip.}}
\providecommand*\mciteBstWouldAddEndPunctfalse
  {\let\EndOfBibitem\relax}
\providecommand*\mciteSetBstMidEndSepPunct[3]{}
\providecommand*\mciteSetBstSublistLabelBeginEnd[3]{}
\providecommand*\EndOfBibitem{}
\mciteSetBstSublistMode{f}
\mciteSetBstMaxWidthForm{subitem}{(\alph{mcitesubitemcount})}
\mciteSetBstSublistLabelBeginEnd
  {\mcitemaxwidthsubitemform\space}
  {\relax}
  {\relax}

\bibitem[Mullard(2014)]{mullard2014new}
Mullard,~A. New drugs cost US \$2.6 billion to develop. \emph{Nat. Rev. Drug
  Discov.} \textbf{2014}, \emph{13}, 877\relax
\mciteBstWouldAddEndPuncttrue
\mciteSetBstMidEndSepPunct{\mcitedefaultmidpunct}
{\mcitedefaultendpunct}{\mcitedefaultseppunct}\relax
\EndOfBibitem
\bibitem[Dowden and Munro(2019)Dowden, and Munro]{dowden2019trends}
Dowden,~H.; Munro,~J. Trends in clinical success rates and therapeutic focus.
  \emph{Nat. Rev. Drug Discov.} \textbf{2019}, \emph{18}, 495--497\relax
\mciteBstWouldAddEndPuncttrue
\mciteSetBstMidEndSepPunct{\mcitedefaultmidpunct}
{\mcitedefaultendpunct}{\mcitedefaultseppunct}\relax
\EndOfBibitem
\bibitem[Schneider(2018)]{schneider2018automating}
Schneider,~G. Automating drug discovery. \emph{Nat. Rev. Drug Discov.}
  \textbf{2018}, \emph{17}, 97\relax
\mciteBstWouldAddEndPuncttrue
\mciteSetBstMidEndSepPunct{\mcitedefaultmidpunct}
{\mcitedefaultendpunct}{\mcitedefaultseppunct}\relax
\EndOfBibitem
\bibitem[Chen \latin{et~al.}(2018)Chen, Engkvist, Wang, Olivecrona, and
  Blaschke]{chen2018rise}
Chen,~H.; Engkvist,~O.; Wang,~Y.; Olivecrona,~M.; Blaschke,~T. The rise of deep
  learning in drug discovery. \emph{Drug Discov. Today} \textbf{2018},
  \emph{23}, 1241--1250\relax
\mciteBstWouldAddEndPuncttrue
\mciteSetBstMidEndSepPunct{\mcitedefaultmidpunct}
{\mcitedefaultendpunct}{\mcitedefaultseppunct}\relax
\EndOfBibitem
\bibitem[Mater and Coote(2019)Mater, and Coote]{mater2019deep}
Mater,~A.~C.; Coote,~M.~L. Deep learning in chemistry. \emph{J. Chem. Inf.
  Model} \textbf{2019}, \emph{59}, 2545--2559\relax
\mciteBstWouldAddEndPuncttrue
\mciteSetBstMidEndSepPunct{\mcitedefaultmidpunct}
{\mcitedefaultendpunct}{\mcitedefaultseppunct}\relax
\EndOfBibitem
\bibitem[Vamathevan \latin{et~al.}(2019)Vamathevan, Clark, Czodrowski, Dunham,
  Ferran, Lee, Li, Madabhushi, Shah, Spitzer, \latin{et~al.}
  others]{vamathevan2019applications}
Vamathevan,~J.; Clark,~D.; Czodrowski,~P.; Dunham,~I.; Ferran,~E.; Lee,~G.;
  Li,~B.; Madabhushi,~A.; Shah,~P.; Spitzer,~M., \latin{et~al.}  Applications
  of machine learning in drug discovery and development. \emph{Nat. Rev. Drug
  Discov.} \textbf{2019}, \emph{18}, 463--477\relax
\mciteBstWouldAddEndPuncttrue
\mciteSetBstMidEndSepPunct{\mcitedefaultmidpunct}
{\mcitedefaultendpunct}{\mcitedefaultseppunct}\relax
\EndOfBibitem
\bibitem[Paul \latin{et~al.}(2021)Paul, Sanap, Shenoy, Kalyane, Kalia, and
  Tekade]{paul2020artificial}
Paul,~D.; Sanap,~G.; Shenoy,~S.; Kalyane,~D.; Kalia,~K.; Tekade,~R.~K.
  Artificial intelligence in drug discovery and development. \emph{Drug
  Discovery Today} \textbf{2021}, \emph{26}, 80\relax
\mciteBstWouldAddEndPuncttrue
\mciteSetBstMidEndSepPunct{\mcitedefaultmidpunct}
{\mcitedefaultendpunct}{\mcitedefaultseppunct}\relax
\EndOfBibitem
\bibitem[Stumpfe and Bajorath(2020)Stumpfe, and Bajorath]{stumpfe2020current}
Stumpfe,~D.; Bajorath,~J. Current trends, overlooked issues, and unmet
  challenges in virtual screening. \emph{J. Chem. Inf. Model} \textbf{2020},
  \emph{60}, 4112--4115\relax
\mciteBstWouldAddEndPuncttrue
\mciteSetBstMidEndSepPunct{\mcitedefaultmidpunct}
{\mcitedefaultendpunct}{\mcitedefaultseppunct}\relax
\EndOfBibitem
\bibitem[Schneider \latin{et~al.}(2020)Schneider, Walters, Plowright, Sieroka,
  Listgarten, Goodnow, Fisher, Jansen, Duca, Rush, \latin{et~al.}
  others]{schneider2020rethinking}
Schneider,~P.; Walters,~W.~P.; Plowright,~A.~T.; Sieroka,~N.; Listgarten,~J.;
  Goodnow,~R.~A.; Fisher,~J.; Jansen,~J.~M.; Duca,~J.~S.; Rush,~T.~S.,
  \latin{et~al.}  Rethinking drug design in the artificial intelligence era.
  \emph{Nat. Rev. Drug Discov.} \textbf{2020}, \emph{19}, 353--364\relax
\mciteBstWouldAddEndPuncttrue
\mciteSetBstMidEndSepPunct{\mcitedefaultmidpunct}
{\mcitedefaultendpunct}{\mcitedefaultseppunct}\relax
\EndOfBibitem
\bibitem[Bostr{\"o}m \latin{et~al.}(2018)Bostr{\"o}m, Brown, Young, and
  Keser{\"u}]{bostrom2018expanding}
Bostr{\"o}m,~J.; Brown,~D.~G.; Young,~R.~J.; Keser{\"u},~G.~M. Expanding the
  medicinal chemistry synthetic toolbox. \emph{Nat. Rev. Drug Discov.}
  \textbf{2018}, \emph{17}, 709--727\relax
\mciteBstWouldAddEndPuncttrue
\mciteSetBstMidEndSepPunct{\mcitedefaultmidpunct}
{\mcitedefaultendpunct}{\mcitedefaultseppunct}\relax
\EndOfBibitem
\bibitem[Strokach \latin{et~al.}(2020)Strokach, Becerra, Corbi-Verge,
  Perez-Riba, and Kim]{strokach2020fast}
Strokach,~A.; Becerra,~D.; Corbi-Verge,~C.; Perez-Riba,~A.; Kim,~P.~M. Fast and
  flexible protein design using deep graph neural networks. \emph{Cell Syst.}
  \textbf{2020}, \emph{11}, 402--411\relax
\mciteBstWouldAddEndPuncttrue
\mciteSetBstMidEndSepPunct{\mcitedefaultmidpunct}
{\mcitedefaultendpunct}{\mcitedefaultseppunct}\relax
\EndOfBibitem
\bibitem[Pushpakom \latin{et~al.}(2019)Pushpakom, Iorio, Eyers, Escott, Hopper,
  Wells, Doig, Guilliams, Latimer, McNamee, \latin{et~al.}
  others]{pushpakom2019drug}
Pushpakom,~S.; Iorio,~F.; Eyers,~P.~A.; Escott,~K.~J.; Hopper,~S.; Wells,~A.;
  Doig,~A.; Guilliams,~T.; Latimer,~J.; McNamee,~C., \latin{et~al.}  Drug
  repurposing: progress, challenges and recommendations. \emph{Nat. Rev. Drug
  Discov.} \textbf{2019}, \emph{18}, 41--58\relax
\mciteBstWouldAddEndPuncttrue
\mciteSetBstMidEndSepPunct{\mcitedefaultmidpunct}
{\mcitedefaultendpunct}{\mcitedefaultseppunct}\relax
\EndOfBibitem
\bibitem[Tsigelny(2019)]{tsigelny2019artificial}
Tsigelny,~I.~F. Artificial intelligence in drug combination therapy.
  \emph{Brief. Bioinformatics} \textbf{2019}, \emph{20}, 1434--1448\relax
\mciteBstWouldAddEndPuncttrue
\mciteSetBstMidEndSepPunct{\mcitedefaultmidpunct}
{\mcitedefaultendpunct}{\mcitedefaultseppunct}\relax
\EndOfBibitem
\bibitem[Paananen and Fortino(2020)Paananen, and Fortino]{paananen2020omics}
Paananen,~J.; Fortino,~V. An omics perspective on drug target discovery
  platforms. \emph{Brief. Bioinformatics} \textbf{2020}, \emph{21},
  1937--1953\relax
\mciteBstWouldAddEndPuncttrue
\mciteSetBstMidEndSepPunct{\mcitedefaultmidpunct}
{\mcitedefaultendpunct}{\mcitedefaultseppunct}\relax
\EndOfBibitem
\bibitem[Hughes \latin{et~al.}(2011)Hughes, Rees, Kalindjian, and
  Philpott]{hughes2011principles}
Hughes,~J.~P.; Rees,~S.; Kalindjian,~S.~B.; Philpott,~K.~L. Principles of early
  drug discovery. \emph{Br. J. Pharmacol.} \textbf{2011}, \emph{162},
  1239--1249\relax
\mciteBstWouldAddEndPuncttrue
\mciteSetBstMidEndSepPunct{\mcitedefaultmidpunct}
{\mcitedefaultendpunct}{\mcitedefaultseppunct}\relax
\EndOfBibitem
\bibitem[Pereira and Williams(2007)Pereira, and Williams]{pereira2007origin}
Pereira,~D.; Williams,~J. Origin and evolution of high throughput screening.
  \emph{Br. J. Pharmacol.} \textbf{2007}, \emph{152}, 53--61\relax
\mciteBstWouldAddEndPuncttrue
\mciteSetBstMidEndSepPunct{\mcitedefaultmidpunct}
{\mcitedefaultendpunct}{\mcitedefaultseppunct}\relax
\EndOfBibitem
\bibitem[Bender \latin{et~al.}(2008)Bender, Bojanic, Davies, Crisman,
  Mikhailov, Scheiber, Jenkins, Deng, Hill, Popov, \latin{et~al.}
  others]{bender2008aspects}
Bender,~A.; Bojanic,~D.; Davies,~J.~W.; Crisman,~T.~J.; Mikhailov,~D.;
  Scheiber,~J.; Jenkins,~J.~L.; Deng,~Z.; Hill,~W. A.~G.; Popov,~M.,
  \latin{et~al.}  Which aspects of HTS are empirically correlated with
  downstream success? \emph{Curr Opin Drug Discov Devel} \textbf{2008},
  \emph{11}, 327\relax
\mciteBstWouldAddEndPuncttrue
\mciteSetBstMidEndSepPunct{\mcitedefaultmidpunct}
{\mcitedefaultendpunct}{\mcitedefaultseppunct}\relax
\EndOfBibitem
\bibitem[Wang \latin{et~al.}(2017)Wang, Bryant, Cheng, Wang, Gindulyte,
  Shoemaker, Thiessen, He, and Zhang]{wang2017pubchem}
Wang,~Y.; Bryant,~S.~H.; Cheng,~T.; Wang,~J.; Gindulyte,~A.; Shoemaker,~B.~A.;
  Thiessen,~P.~A.; He,~S.; Zhang,~J. Pubchem bioassay: 2017 update.
  \emph{Nucleic Acids Res.} \textbf{2017}, \emph{45}, D955--D963\relax
\mciteBstWouldAddEndPuncttrue
\mciteSetBstMidEndSepPunct{\mcitedefaultmidpunct}
{\mcitedefaultendpunct}{\mcitedefaultseppunct}\relax
\EndOfBibitem
\bibitem[Sterling and Irwin(2015)Sterling, and Irwin]{sterling2015zinc}
Sterling,~T.; Irwin,~J.~J. ZINC 15--ligand discovery for everyone. \emph{J.
  Chem. Inf. Model} \textbf{2015}, \emph{55}, 2324--2337\relax
\mciteBstWouldAddEndPuncttrue
\mciteSetBstMidEndSepPunct{\mcitedefaultmidpunct}
{\mcitedefaultendpunct}{\mcitedefaultseppunct}\relax
\EndOfBibitem
\bibitem[Kim(2016)]{kim2016getting}
Kim,~S. Getting the most out of PubChem for virtual screening. \emph{Expert
  Opin Drug Discov} \textbf{2016}, \emph{11}, 843--855\relax
\mciteBstWouldAddEndPuncttrue
\mciteSetBstMidEndSepPunct{\mcitedefaultmidpunct}
{\mcitedefaultendpunct}{\mcitedefaultseppunct}\relax
\EndOfBibitem
\bibitem[Scior \latin{et~al.}(2012)Scior, Bender, Tresadern, Medina-Franco,
  Mart{\'\i}nez-Mayorga, Langer, Cuanalo-Contreras, and
  Agrafiotis]{scior2012recognizing}
Scior,~T.; Bender,~A.; Tresadern,~G.; Medina-Franco,~J.~L.;
  Mart{\'\i}nez-Mayorga,~K.; Langer,~T.; Cuanalo-Contreras,~K.;
  Agrafiotis,~D.~K. Recognizing pitfalls in virtual screening: a critical
  review. \emph{J. Chem. Inf. Model} \textbf{2012}, \emph{52}, 867--881\relax
\mciteBstWouldAddEndPuncttrue
\mciteSetBstMidEndSepPunct{\mcitedefaultmidpunct}
{\mcitedefaultendpunct}{\mcitedefaultseppunct}\relax
\EndOfBibitem
\bibitem[Salahudeen and Nishtala(2017)Salahudeen, and
  Nishtala]{salahudeen2017overview}
Salahudeen,~M.~S.; Nishtala,~P.~S. An overview of pharmacodynamic modelling,
  ligand-binding approach and its application in clinical practice. \emph{Saudi
  Pharm J} \textbf{2017}, \emph{25}, 165--175\relax
\mciteBstWouldAddEndPuncttrue
\mciteSetBstMidEndSepPunct{\mcitedefaultmidpunct}
{\mcitedefaultendpunct}{\mcitedefaultseppunct}\relax
\EndOfBibitem
\bibitem[Hu and Bajorath(2013)Hu, and Bajorath]{hu2013compound}
Hu,~Y.; Bajorath,~J. Compound promiscuity: what can we learn from current data?
  \emph{Drug Discov. Today} \textbf{2013}, \emph{18}, 644--650\relax
\mciteBstWouldAddEndPuncttrue
\mciteSetBstMidEndSepPunct{\mcitedefaultmidpunct}
{\mcitedefaultendpunct}{\mcitedefaultseppunct}\relax
\EndOfBibitem
\bibitem[Yusof \latin{et~al.}(2014)Yusof, Shah, Hashimoto, Segall, and
  Greene]{yusof2014finding}
Yusof,~I.; Shah,~F.; Hashimoto,~T.; Segall,~M.~D.; Greene,~N. Finding the rules
  for successful drug optimisation. \emph{Drug Discov. Today} \textbf{2014},
  \emph{19}, 680--687\relax
\mciteBstWouldAddEndPuncttrue
\mciteSetBstMidEndSepPunct{\mcitedefaultmidpunct}
{\mcitedefaultendpunct}{\mcitedefaultseppunct}\relax
\EndOfBibitem
\bibitem[Nicolaou and Brown(2013)Nicolaou, and Brown]{nicolaou2013multi}
Nicolaou,~C.~A.; Brown,~N. Multi-objective optimization methods in drug design.
  \emph{Drug Discov. Today: Technologies} \textbf{2013}, \emph{10},
  e427--e435\relax
\mciteBstWouldAddEndPuncttrue
\mciteSetBstMidEndSepPunct{\mcitedefaultmidpunct}
{\mcitedefaultendpunct}{\mcitedefaultseppunct}\relax
\EndOfBibitem
\bibitem[Muratov \latin{et~al.}(2020)Muratov, Bajorath, Sheridan, Tetko,
  Filimonov, Poroikov, Oprea, Baskin, Varnek, Roitberg, \latin{et~al.}
  others]{muratov2020qsar}
Muratov,~E.~N.; Bajorath,~J.; Sheridan,~R.~P.; Tetko,~I.~V.; Filimonov,~D.;
  Poroikov,~V.; Oprea,~T.~I.; Baskin,~I.~I.; Varnek,~A.; Roitberg,~A.,
  \latin{et~al.}  QSAR without borders. \emph{Chem. Soc. Rev.} \textbf{2020},
  \emph{49}, 3525--3564\relax
\mciteBstWouldAddEndPuncttrue
\mciteSetBstMidEndSepPunct{\mcitedefaultmidpunct}
{\mcitedefaultendpunct}{\mcitedefaultseppunct}\relax
\EndOfBibitem
\bibitem[Schneider and Fechner(2005)Schneider, and
  Fechner]{schneider2005computer}
Schneider,~G.; Fechner,~U. Computer-based de novo design of drug-like
  molecules. \emph{Nat. Rev. Drug Discov.} \textbf{2005}, \emph{4},
  649--663\relax
\mciteBstWouldAddEndPuncttrue
\mciteSetBstMidEndSepPunct{\mcitedefaultmidpunct}
{\mcitedefaultendpunct}{\mcitedefaultseppunct}\relax
\EndOfBibitem
\bibitem[Dobson(2004)]{dobson2004chemical}
Dobson,~C.~M. Chemical space and biology. 2004\relax
\mciteBstWouldAddEndPuncttrue
\mciteSetBstMidEndSepPunct{\mcitedefaultmidpunct}
{\mcitedefaultendpunct}{\mcitedefaultseppunct}\relax
\EndOfBibitem
\bibitem[Sliwoski \latin{et~al.}(2014)Sliwoski, Kothiwale, Meiler, and
  Lowe]{sliwoski2014computational}
Sliwoski,~G.; Kothiwale,~S.; Meiler,~J.; Lowe,~E.~W. Computational methods in
  drug discovery. \emph{Pharmacol. Rev.} \textbf{2014}, \emph{66},
  334--395\relax
\mciteBstWouldAddEndPuncttrue
\mciteSetBstMidEndSepPunct{\mcitedefaultmidpunct}
{\mcitedefaultendpunct}{\mcitedefaultseppunct}\relax
\EndOfBibitem
\bibitem[Van~Drie(2007)]{van2007computer}
Van~Drie,~J.~H. Computer-aided drug design: the next 20 years. \emph{J. Comput.
  Aided Mol. Des.} \textbf{2007}, \emph{21}, 591--601\relax
\mciteBstWouldAddEndPuncttrue
\mciteSetBstMidEndSepPunct{\mcitedefaultmidpunct}
{\mcitedefaultendpunct}{\mcitedefaultseppunct}\relax
\EndOfBibitem
\bibitem[Jim{\'e}nez-Luna \latin{et~al.}(2020)Jim{\'e}nez-Luna, Grisoni, and
  Schneider]{jimenez2020drug}
Jim{\'e}nez-Luna,~J.; Grisoni,~F.; Schneider,~G. Drug discovery with
  explainable artificial intelligence. \emph{Nat. Mach. Intell.} \textbf{2020},
  \emph{2}, 573--584\relax
\mciteBstWouldAddEndPuncttrue
\mciteSetBstMidEndSepPunct{\mcitedefaultmidpunct}
{\mcitedefaultendpunct}{\mcitedefaultseppunct}\relax
\EndOfBibitem
\bibitem[Bajorath(2002)]{bajorath2002integration}
Bajorath,~J. Integration of virtual and high-throughput screening. \emph{Nat.
  Rev. Drug Discov.} \textbf{2002}, \emph{1}, 882--894\relax
\mciteBstWouldAddEndPuncttrue
\mciteSetBstMidEndSepPunct{\mcitedefaultmidpunct}
{\mcitedefaultendpunct}{\mcitedefaultseppunct}\relax
\EndOfBibitem
\bibitem[Schneider(2010)]{schneider2010virtual}
Schneider,~G. Virtual screening: an endless staircase? \emph{Nat. Rev. Drug
  Discov.} \textbf{2010}, \emph{9}, 273--276\relax
\mciteBstWouldAddEndPuncttrue
\mciteSetBstMidEndSepPunct{\mcitedefaultmidpunct}
{\mcitedefaultendpunct}{\mcitedefaultseppunct}\relax
\EndOfBibitem
\bibitem[Polishchuk(2017)]{polishchuk2017interpretation}
Polishchuk,~P. Interpretation of quantitative structure--activity relationship
  models: past, present, and future. \emph{J. Chem. Inf. Model} \textbf{2017},
  \emph{57}, 2618--2639\relax
\mciteBstWouldAddEndPuncttrue
\mciteSetBstMidEndSepPunct{\mcitedefaultmidpunct}
{\mcitedefaultendpunct}{\mcitedefaultseppunct}\relax
\EndOfBibitem
\bibitem[Sydow \latin{et~al.}(2019)Sydow, Burggraaff, Szengel, van Vlijmen,
  IJzerman, van Westen, and Volkamer]{sydow2019advances}
Sydow,~D.; Burggraaff,~L.; Szengel,~A.; van Vlijmen,~H.~W.; IJzerman,~A.~P.;
  van Westen,~G.~J.; Volkamer,~A. Advances and challenges in computational
  target prediction. \emph{J. Chem. Inf. Model} \textbf{2019}, \emph{59},
  1728--1742\relax
\mciteBstWouldAddEndPuncttrue
\mciteSetBstMidEndSepPunct{\mcitedefaultmidpunct}
{\mcitedefaultendpunct}{\mcitedefaultseppunct}\relax
\EndOfBibitem
\bibitem[Maggiora(2006)]{maggiora2006outliers}
Maggiora,~G. On outliers and activity cliffs--why QSAR often disappoints.
  \emph{J Chem Inf Model} \textbf{2006}, \emph{46}, 1535--1535\relax
\mciteBstWouldAddEndPuncttrue
\mciteSetBstMidEndSepPunct{\mcitedefaultmidpunct}
{\mcitedefaultendpunct}{\mcitedefaultseppunct}\relax
\EndOfBibitem
\bibitem[Stumpfe \latin{et~al.}(2014)Stumpfe, Hu, Dimova, and
  Bajorath]{stumpfe2014recent}
Stumpfe,~D.; Hu,~Y.; Dimova,~D.; Bajorath,~J. Recent progress in understanding
  activity cliffs and their utility in medicinal chemistry: miniperspective.
  \emph{J. Med. Chem.} \textbf{2014}, \emph{57}, 18--28\relax
\mciteBstWouldAddEndPuncttrue
\mciteSetBstMidEndSepPunct{\mcitedefaultmidpunct}
{\mcitedefaultendpunct}{\mcitedefaultseppunct}\relax
\EndOfBibitem
\bibitem[Bajorath(2019)]{bajorath2019duality}
Bajorath,~J. Duality of activity cliffs in drug discovery. \emph{Expert Opin
  Drug Discov} \textbf{2019}, \emph{14}, 517--520\relax
\mciteBstWouldAddEndPuncttrue
\mciteSetBstMidEndSepPunct{\mcitedefaultmidpunct}
{\mcitedefaultendpunct}{\mcitedefaultseppunct}\relax
\EndOfBibitem
\bibitem[Ma \latin{et~al.}(2015)Ma, Sheridan, Liaw, Dahl, and
  Svetnik]{ma2015deep}
Ma,~J.; Sheridan,~R.~P.; Liaw,~A.; Dahl,~G.~E.; Svetnik,~V. Deep neural nets as
  a method for quantitative structure--activity relationships. \emph{J. Chem.
  Inf. Model} \textbf{2015}, \emph{55}, 263--274\relax
\mciteBstWouldAddEndPuncttrue
\mciteSetBstMidEndSepPunct{\mcitedefaultmidpunct}
{\mcitedefaultendpunct}{\mcitedefaultseppunct}\relax
\EndOfBibitem
\bibitem[Lavecchia(2015)]{lavecchia2015machine}
Lavecchia,~A. Machine-learning approaches in drug discovery: methods and
  applications. \emph{Drug Discov. Today} \textbf{2015}, \emph{20},
  318--331\relax
\mciteBstWouldAddEndPuncttrue
\mciteSetBstMidEndSepPunct{\mcitedefaultmidpunct}
{\mcitedefaultendpunct}{\mcitedefaultseppunct}\relax
\EndOfBibitem
\bibitem[Krizhevsky \latin{et~al.}(2012)Krizhevsky, Sutskever, and
  Hinton]{krizhevsky2012imagenet}
Krizhevsky,~A.; Sutskever,~I.; Hinton,~G.~E. Imagenet classification with deep
  convolutional neural networks. \emph{Advances in Neural Information
  Processing Systems} \textbf{2012}, \emph{25}, 1097--1105\relax
\mciteBstWouldAddEndPuncttrue
\mciteSetBstMidEndSepPunct{\mcitedefaultmidpunct}
{\mcitedefaultendpunct}{\mcitedefaultseppunct}\relax
\EndOfBibitem
\bibitem[Alom \latin{et~al.}(2018)Alom, Taha, Yakopcic, Westberg, Sidike,
  Nasrin, Van~Esesn, Awwal, and Asari]{alom2018history}
Alom,~M.~Z.; Taha,~T.~M.; Yakopcic,~C.; Westberg,~S.; Sidike,~P.;
  Nasrin,~M.~S.; Van~Esesn,~B.~C.; Awwal,~A. A.~S.; Asari,~V.~K. The history
  began from alexnet: A comprehensive survey on deep learning approaches.
  \emph{arXiv preprint arXiv:1803.01164} \textbf{2018}, \relax
\mciteBstWouldAddEndPunctfalse
\mciteSetBstMidEndSepPunct{\mcitedefaultmidpunct}
{}{\mcitedefaultseppunct}\relax
\EndOfBibitem
\bibitem[{\"O}zt{\"u}rk \latin{et~al.}(2020){\"O}zt{\"u}rk, {\"O}zg{\"u}r,
  Schwaller, Laino, and Ozkirimli]{ozturk2020exploring}
{\"O}zt{\"u}rk,~H.; {\"O}zg{\"u}r,~A.; Schwaller,~P.; Laino,~T.; Ozkirimli,~E.
  Exploring chemical space using natural language processing methodologies for
  drug discovery. \emph{Drug Discov. Today} \textbf{2020}, \emph{25},
  689--705\relax
\mciteBstWouldAddEndPuncttrue
\mciteSetBstMidEndSepPunct{\mcitedefaultmidpunct}
{\mcitedefaultendpunct}{\mcitedefaultseppunct}\relax
\EndOfBibitem
\bibitem[Jim{\'e}nez-Luna \latin{et~al.}(2021)Jim{\'e}nez-Luna, Grisoni,
  Weskamp, and Schneider]{jimenez2021artificial}
Jim{\'e}nez-Luna,~J.; Grisoni,~F.; Weskamp,~N.; Schneider,~G. Artificial
  intelligence in drug discovery: Recent advances and future perspectives.
  \emph{Expert Opin Drug Discov} \textbf{2021}, 1--11\relax
\mciteBstWouldAddEndPuncttrue
\mciteSetBstMidEndSepPunct{\mcitedefaultmidpunct}
{\mcitedefaultendpunct}{\mcitedefaultseppunct}\relax
\EndOfBibitem
\bibitem[Zhavoronkov \latin{et~al.}(2019)Zhavoronkov, Ivanenkov, Aliper,
  Veselov, Aladinskiy, Aladinskaya, Terentiev, Polykovskiy, Kuznetsov,
  Asadulaev, \latin{et~al.} others]{zhavoronkov2019deep}
Zhavoronkov,~A.; Ivanenkov,~Y.~A.; Aliper,~A.; Veselov,~M.~S.;
  Aladinskiy,~V.~A.; Aladinskaya,~A.~V.; Terentiev,~V.~A.; Polykovskiy,~D.~A.;
  Kuznetsov,~M.~D.; Asadulaev,~A., \latin{et~al.}  Deep learning enables rapid
  identification of potent DDR1 kinase inhibitors. \emph{Nat. Biotechnol.}
  \textbf{2019}, \emph{37}, 1038--1040\relax
\mciteBstWouldAddEndPuncttrue
\mciteSetBstMidEndSepPunct{\mcitedefaultmidpunct}
{\mcitedefaultendpunct}{\mcitedefaultseppunct}\relax
\EndOfBibitem
\bibitem[Stokes \latin{et~al.}(2020)Stokes, Yang, Swanson, Jin, Cubillos-Ruiz,
  Donghia, MacNair, French, Carfrae, Bloom-Ackermann, \latin{et~al.}
  others]{stokes2020deep}
Stokes,~J.~M.; Yang,~K.; Swanson,~K.; Jin,~W.; Cubillos-Ruiz,~A.;
  Donghia,~N.~M.; MacNair,~C.~R.; French,~S.; Carfrae,~L.~A.;
  Bloom-Ackermann,~Z., \latin{et~al.}  A deep learning approach to antibiotic
  discovery. \emph{Cell} \textbf{2020}, \emph{180}, 688--702\relax
\mciteBstWouldAddEndPuncttrue
\mciteSetBstMidEndSepPunct{\mcitedefaultmidpunct}
{\mcitedefaultendpunct}{\mcitedefaultseppunct}\relax
\EndOfBibitem
\bibitem[Chuang \latin{et~al.}(2020)Chuang, Gunsalus, and
  Keiser]{chuang2020learning}
Chuang,~K.~V.; Gunsalus,~L.~M.; Keiser,~M.~J. Learning Molecular
  Representations for Medicinal Chemistry: Miniperspective. \emph{J. Med.
  Chem.} \textbf{2020}, \emph{63}, 8705--8722\relax
\mciteBstWouldAddEndPuncttrue
\mciteSetBstMidEndSepPunct{\mcitedefaultmidpunct}
{\mcitedefaultendpunct}{\mcitedefaultseppunct}\relax
\EndOfBibitem
\bibitem[Mayr \latin{et~al.}(2016)Mayr, Klambauer, Unterthiner, and
  Hochreiter]{mayr2016deeptox}
Mayr,~A.; Klambauer,~G.; Unterthiner,~T.; Hochreiter,~S. DeepTox: toxicity
  prediction using deep learning. \emph{Frontiers in Environmental Science}
  \textbf{2016}, \emph{3}, 80\relax
\mciteBstWouldAddEndPuncttrue
\mciteSetBstMidEndSepPunct{\mcitedefaultmidpunct}
{\mcitedefaultendpunct}{\mcitedefaultseppunct}\relax
\EndOfBibitem
\bibitem[Andrade \latin{et~al.}(2019)Andrade, Chalasani, Bj{\"o}rnsson, Suzuki,
  Kullak-Ublick, Watkins, Devarbhavi, Merz, Lucena, Kaplowitz, \latin{et~al.}
  others]{andrade2019drug}
Andrade,~R.~J.; Chalasani,~N.; Bj{\"o}rnsson,~E.~S.; Suzuki,~A.;
  Kullak-Ublick,~G.~A.; Watkins,~P.~B.; Devarbhavi,~H.; Merz,~M.;
  Lucena,~M.~I.; Kaplowitz,~N., \latin{et~al.}  Drug-induced liver injury.
  \emph{Nat. Rev. Dis. Primers} \textbf{2019}, \emph{5}, 1--22\relax
\mciteBstWouldAddEndPuncttrue
\mciteSetBstMidEndSepPunct{\mcitedefaultmidpunct}
{\mcitedefaultendpunct}{\mcitedefaultseppunct}\relax
\EndOfBibitem
\bibitem[Elton \latin{et~al.}(2019)Elton, Boukouvalas, Fuge, and
  Chung]{elton2019deep}
Elton,~D.~C.; Boukouvalas,~Z.; Fuge,~M.~D.; Chung,~P.~W. Deep learning for
  molecular design—a review of the state of the art. \emph{Mol. Syst. Des.
  Eng.} \textbf{2019}, \emph{4}, 828--849\relax
\mciteBstWouldAddEndPuncttrue
\mciteSetBstMidEndSepPunct{\mcitedefaultmidpunct}
{\mcitedefaultendpunct}{\mcitedefaultseppunct}\relax
\EndOfBibitem
\bibitem[Mercado \latin{et~al.}(2020)Mercado, Rastemo, Lindelöf, Klambauer,
  Engkvist, Chen, and Bjerrum]{mercado2020practical}
Mercado,~R.; Rastemo,~T.; Lindelöf,~E.; Klambauer,~G.; Engkvist,~O.; Chen,~H.;
  Bjerrum,~E.~J. {Practical Notes on Building Molecular Graph Generative
  Models}. \emph{Applied AI Letters} \textbf{2020}, \relax
\mciteBstWouldAddEndPunctfalse
\mciteSetBstMidEndSepPunct{\mcitedefaultmidpunct}
{}{\mcitedefaultseppunct}\relax
\EndOfBibitem
\bibitem[Schaduangrat \latin{et~al.}(2020)Schaduangrat, Lampa, Simeon, Gleeson,
  Spjuth, and Nantasenamat]{schaduangrat2020towards}
Schaduangrat,~N.; Lampa,~S.; Simeon,~S.; Gleeson,~M.~P.; Spjuth,~O.;
  Nantasenamat,~C. Towards reproducible computational drug discovery. \emph{J.
  Cheminformatics} \textbf{2020}, \emph{12}, 9\relax
\mciteBstWouldAddEndPuncttrue
\mciteSetBstMidEndSepPunct{\mcitedefaultmidpunct}
{\mcitedefaultendpunct}{\mcitedefaultseppunct}\relax
\EndOfBibitem
\bibitem[Bender and Cortes-Ciriano(2020)Bender, and
  Cortes-Ciriano]{bender2020artificial}
Bender,~A.; Cortes-Ciriano,~I. Artificial intelligence in drug discovery: what
  is realistic, what are illusions? Part 1: Ways to make an impact, and why we
  are not there yet. \emph{Drug Discov. Today} \textbf{2020}, \relax
\mciteBstWouldAddEndPunctfalse
\mciteSetBstMidEndSepPunct{\mcitedefaultmidpunct}
{}{\mcitedefaultseppunct}\relax
\EndOfBibitem
\bibitem[Bender and Cortes-Ciriano(2021)Bender, and
  Cortes-Ciriano]{bender2021artificial}
Bender,~A.; Cortes-Ciriano,~I. Artificial intelligence in drug discovery: what
  is realistic, what are illusions? Part 2: a discussion of chemical and
  biological data used for AI in drug discovery. \emph{Drug Discov. Today}
  \textbf{2021}, \relax
\mciteBstWouldAddEndPunctfalse
\mciteSetBstMidEndSepPunct{\mcitedefaultmidpunct}
{}{\mcitedefaultseppunct}\relax
\EndOfBibitem
\bibitem[Walters and Barzilay(2021)Walters, and Barzilay]{walters2021critical}
Walters,~W.~P.; Barzilay,~R. Critical assessment of AI in drug discovery.
  \emph{Expert Opin Drug Discov} \textbf{2021}, 1--11\relax
\mciteBstWouldAddEndPuncttrue
\mciteSetBstMidEndSepPunct{\mcitedefaultmidpunct}
{\mcitedefaultendpunct}{\mcitedefaultseppunct}\relax
\EndOfBibitem
\bibitem[Rifaioglu \latin{et~al.}(2019)Rifaioglu, Atas, Martin, Cetin-Atalay,
  Atalay, and Do{\u{g}}an]{rifaioglu2019recent}
Rifaioglu,~A.~S.; Atas,~H.; Martin,~M.~J.; Cetin-Atalay,~R.; Atalay,~V.;
  Do{\u{g}}an,~T. Recent applications of deep learning and machine intelligence
  on in silico drug discovery: methods, tools and databases. \emph{Brief.
  Bioinformatics} \textbf{2019}, \emph{20}, 1878--1912\relax
\mciteBstWouldAddEndPuncttrue
\mciteSetBstMidEndSepPunct{\mcitedefaultmidpunct}
{\mcitedefaultendpunct}{\mcitedefaultseppunct}\relax
\EndOfBibitem
\bibitem[Kim \latin{et~al.}(2021)Kim, Chen, Cheng, Gindulyte, He, He, Li,
  Shoemaker, Thiessen, Yu, \latin{et~al.} others]{kim2021pubchem}
Kim,~S.; Chen,~J.; Cheng,~T.; Gindulyte,~A.; He,~J.; He,~S.; Li,~Q.;
  Shoemaker,~B.~A.; Thiessen,~P.~A.; Yu,~B., \latin{et~al.}  PubChem in 2021:
  new data content and improved web interfaces. \emph{Nucleic Acids Res.}
  \textbf{2021}, \emph{49}, D1388--D1395\relax
\mciteBstWouldAddEndPuncttrue
\mciteSetBstMidEndSepPunct{\mcitedefaultmidpunct}
{\mcitedefaultendpunct}{\mcitedefaultseppunct}\relax
\EndOfBibitem
\bibitem[Korkmaz(2020)]{korkmaz2020deep}
Korkmaz,~S. Deep learning-based imbalanced data classification for drug
  discovery. \emph{J. Chem. Inf. Model} \textbf{2020}, \emph{60},
  4180--4190\relax
\mciteBstWouldAddEndPuncttrue
\mciteSetBstMidEndSepPunct{\mcitedefaultmidpunct}
{\mcitedefaultendpunct}{\mcitedefaultseppunct}\relax
\EndOfBibitem
\bibitem[Chithrananda \latin{et~al.}(2020)Chithrananda, Grand, and
  Ramsundar]{chithrananda2020chemberta}
Chithrananda,~S.; Grand,~G.; Ramsundar,~B. ChemBERTa: Large-Scale
  Self-Supervised Pretraining for Molecular Property Prediction. \emph{arXiv
  preprint arXiv:2010.09885} \textbf{2020}, \relax
\mciteBstWouldAddEndPunctfalse
\mciteSetBstMidEndSepPunct{\mcitedefaultmidpunct}
{}{\mcitedefaultseppunct}\relax
\EndOfBibitem
\bibitem[Gaulton \latin{et~al.}(2017)Gaulton, Hersey, Nowotka, Bento, Chambers,
  Mendez, Mutowo, Atkinson, Bellis, Cibri{\'a}n-Uhalte, \latin{et~al.}
  others]{gaulton2017chembl}
Gaulton,~A.; Hersey,~A.; Nowotka,~M.; Bento,~A.~P.; Chambers,~J.; Mendez,~D.;
  Mutowo,~P.; Atkinson,~F.; Bellis,~L.~J.; Cibri{\'a}n-Uhalte,~E.,
  \latin{et~al.}  The ChEMBL database in 2017. \emph{Nucleic Acids Res.}
  \textbf{2017}, \emph{45}, D945--D954\relax
\mciteBstWouldAddEndPuncttrue
\mciteSetBstMidEndSepPunct{\mcitedefaultmidpunct}
{\mcitedefaultendpunct}{\mcitedefaultseppunct}\relax
\EndOfBibitem
\bibitem[Davies \latin{et~al.}(2015)Davies, Nowotka, Papadatos, Dedman,
  Gaulton, Atkinson, Bellis, and Overington]{davies2015chembl}
Davies,~M.; Nowotka,~M.; Papadatos,~G.; Dedman,~N.; Gaulton,~A.; Atkinson,~F.;
  Bellis,~L.; Overington,~J.~P. ChEMBL web services: streamlining access to
  drug discovery data and utilities. \emph{Nucleic Acids Res.} \textbf{2015},
  \emph{43}, W612--W620\relax
\mciteBstWouldAddEndPuncttrue
\mciteSetBstMidEndSepPunct{\mcitedefaultmidpunct}
{\mcitedefaultendpunct}{\mcitedefaultseppunct}\relax
\EndOfBibitem
\bibitem[Mayr \latin{et~al.}(2018)Mayr, Klambauer, Unterthiner, Steijaert,
  Wegner, Ceulemans, Clevert, and Hochreiter]{mayr2018large}
Mayr,~A.; Klambauer,~G.; Unterthiner,~T.; Steijaert,~M.; Wegner,~J.~K.;
  Ceulemans,~H.; Clevert,~D.-A.; Hochreiter,~S. Large-scale comparison of
  machine learning methods for drug target prediction on ChEMBL. \emph{Chem.
  Sci} \textbf{2018}, \emph{9}, 5441--5451\relax
\mciteBstWouldAddEndPuncttrue
\mciteSetBstMidEndSepPunct{\mcitedefaultmidpunct}
{\mcitedefaultendpunct}{\mcitedefaultseppunct}\relax
\EndOfBibitem
\bibitem[Rong \latin{et~al.}(2020)Rong, Bian, Xu, Xie, Wei, Huang, and
  Huang]{rong2020grover}
Rong,~Y.; Bian,~Y.; Xu,~T.; Xie,~W.; Wei,~Y.; Huang,~W.; Huang,~J. Grover:
  Self-supervised message passing transformer on large-scale molecular data.
  \emph{arXiv preprint arXiv:2007.02835} \textbf{2020}, \relax
\mciteBstWouldAddEndPunctfalse
\mciteSetBstMidEndSepPunct{\mcitedefaultmidpunct}
{}{\mcitedefaultseppunct}\relax
\EndOfBibitem
\bibitem[Polykovskiy \latin{et~al.}(2020)Polykovskiy, Zhebrak,
  Sanchez-Lengeling, Golovanov, Tatanov, Belyaev, Kurbanov, Artamonov,
  Aladinskiy, Veselov, \latin{et~al.} others]{polykovskiy2020molecular}
Polykovskiy,~D.; Zhebrak,~A.; Sanchez-Lengeling,~B.; Golovanov,~S.;
  Tatanov,~O.; Belyaev,~S.; Kurbanov,~R.; Artamonov,~A.; Aladinskiy,~V.;
  Veselov,~M., \latin{et~al.}  Molecular sets (MOSES): a benchmarking platform
  for molecular generation models. \emph{Front. Pharmacol.} \textbf{2020},
  \emph{11}\relax
\mciteBstWouldAddEndPuncttrue
\mciteSetBstMidEndSepPunct{\mcitedefaultmidpunct}
{\mcitedefaultendpunct}{\mcitedefaultseppunct}\relax
\EndOfBibitem
\bibitem[Lagarde \latin{et~al.}(2015)Lagarde, Zagury, and
  Montes]{lagarde2015benchmarking}
Lagarde,~N.; Zagury,~J.-F.; Montes,~M. Benchmarking data sets for the
  evaluation of virtual ligand screening methods: review and perspectives.
  \emph{J. Chem. Inf. Model} \textbf{2015}, \emph{55}, 1297--1307\relax
\mciteBstWouldAddEndPuncttrue
\mciteSetBstMidEndSepPunct{\mcitedefaultmidpunct}
{\mcitedefaultendpunct}{\mcitedefaultseppunct}\relax
\EndOfBibitem
\bibitem[Chen \latin{et~al.}(2016)Chen, Suzuki, Thakkar, Yu, Hu, and
  Tong]{chen2016dilirank}
Chen,~M.; Suzuki,~A.; Thakkar,~S.; Yu,~K.; Hu,~C.; Tong,~W. DILIrank: the
  largest reference drug list ranked by the risk for developing drug-induced
  liver injury in humans. \emph{Drug Discov Today} \textbf{2016}, \emph{21},
  648--653\relax
\mciteBstWouldAddEndPuncttrue
\mciteSetBstMidEndSepPunct{\mcitedefaultmidpunct}
{\mcitedefaultendpunct}{\mcitedefaultseppunct}\relax
\EndOfBibitem
\bibitem[David \latin{et~al.}(2020)David, Thakkar, Mercado, and
  Engkvist]{david2020molecular}
David,~L.; Thakkar,~A.; Mercado,~R.; Engkvist,~O. Molecular representations in
  AI-driven drug discovery: a review and practical guide. \emph{J.
  Cheminformatics} \textbf{2020}, \emph{12}, 1--22\relax
\mciteBstWouldAddEndPuncttrue
\mciteSetBstMidEndSepPunct{\mcitedefaultmidpunct}
{\mcitedefaultendpunct}{\mcitedefaultseppunct}\relax
\EndOfBibitem
\bibitem[Morgan(1965)]{morgan1965generation}
Morgan,~H.~L. The generation of a unique machine description for chemical
  structures-a technique developed at chemical abstracts service. \emph{J.
  Chem. Doc} \textbf{1965}, \emph{5}, 107--113\relax
\mciteBstWouldAddEndPuncttrue
\mciteSetBstMidEndSepPunct{\mcitedefaultmidpunct}
{\mcitedefaultendpunct}{\mcitedefaultseppunct}\relax
\EndOfBibitem
\bibitem[Subramanian \latin{et~al.}(2016)Subramanian, Ramsundar, Pande, and
  Denny]{subramanian2016computational}
Subramanian,~G.; Ramsundar,~B.; Pande,~V.; Denny,~R.~A. Computational modeling
  of $\beta$-secretase 1 (BACE-1) inhibitors using ligand based approaches.
  \emph{J. Chem. Inf. Model} \textbf{2016}, \emph{56}, 1936--1949\relax
\mciteBstWouldAddEndPuncttrue
\mciteSetBstMidEndSepPunct{\mcitedefaultmidpunct}
{\mcitedefaultendpunct}{\mcitedefaultseppunct}\relax
\EndOfBibitem
\bibitem[Zang \latin{et~al.}(2017)Zang, Mansouri, Williams, Judson, Allen,
  Casey, and Kleinstreuer]{zang2017silico}
Zang,~Q.; Mansouri,~K.; Williams,~A.~J.; Judson,~R.~S.; Allen,~D.~G.;
  Casey,~W.~M.; Kleinstreuer,~N.~C. In silico prediction of physicochemical
  properties of environmental chemicals using molecular fingerprints and
  machine learning. \emph{J. Chem. Inf. Model} \textbf{2017}, \emph{57},
  36--49\relax
\mciteBstWouldAddEndPuncttrue
\mciteSetBstMidEndSepPunct{\mcitedefaultmidpunct}
{\mcitedefaultendpunct}{\mcitedefaultseppunct}\relax
\EndOfBibitem
\bibitem[Yang \latin{et~al.}(2019)Yang, Swanson, Jin, Coley, Eiden, Gao,
  Guzman-Perez, Hopper, Kelley, Mathea, \latin{et~al.}
  others]{yang2019analyzing}
Yang,~K.; Swanson,~K.; Jin,~W.; Coley,~C.; Eiden,~P.; Gao,~H.;
  Guzman-Perez,~A.; Hopper,~T.; Kelley,~B.; Mathea,~M., \latin{et~al.}
  Analyzing learned molecular representations for property prediction. \emph{J.
  Chem. Inf. Model} \textbf{2019}, \emph{59}, 3370--3388\relax
\mciteBstWouldAddEndPuncttrue
\mciteSetBstMidEndSepPunct{\mcitedefaultmidpunct}
{\mcitedefaultendpunct}{\mcitedefaultseppunct}\relax
\EndOfBibitem
\bibitem[Mercado \latin{et~al.}(2020)Mercado, Rastemo, Lindelöf, Klambauer,
  Engkvist, Chen, and Bjerrum]{mercado2020graph}
Mercado,~R.; Rastemo,~T.; Lindelöf,~E.; Klambauer,~G.; Engkvist,~O.; Chen,~H.;
  Bjerrum,~E.~J. {Graph Networks for Molecular Design}. \emph{Mach. Learn.:
  Sci. Technol.} \textbf{2020}, \relax
\mciteBstWouldAddEndPunctfalse
\mciteSetBstMidEndSepPunct{\mcitedefaultmidpunct}
{}{\mcitedefaultseppunct}\relax
\EndOfBibitem
\bibitem[Jin \latin{et~al.}(2020)Jin, Barzilay, and Jaakkola]{jin2020multi}
Jin,~W.; Barzilay,~R.; Jaakkola,~T. Multi-objective molecule generation using
  interpretable substructures. International Conference on Machine Learning.
  2020; pp 4849--4859\relax
\mciteBstWouldAddEndPuncttrue
\mciteSetBstMidEndSepPunct{\mcitedefaultmidpunct}
{\mcitedefaultendpunct}{\mcitedefaultseppunct}\relax
\EndOfBibitem
\bibitem[Weininger(1988)]{weininger1988smiles}
Weininger,~D. SMILES, a chemical language and information system. 1.
  Introduction to methodology and encoding rules. \emph{J Chem Inform Comput
  Sci} \textbf{1988}, \emph{28}, 31--36\relax
\mciteBstWouldAddEndPuncttrue
\mciteSetBstMidEndSepPunct{\mcitedefaultmidpunct}
{\mcitedefaultendpunct}{\mcitedefaultseppunct}\relax
\EndOfBibitem
\bibitem[Weininger \latin{et~al.}(1989)Weininger, Weininger, and
  Weininger]{weininger1989smiles}
Weininger,~D.; Weininger,~A.; Weininger,~J.~L. SMILES. 2. Algorithm for
  generation of unique SMILES notation. \emph{J Chem Inform Comput Sci}
  \textbf{1989}, \emph{29}, 97--101\relax
\mciteBstWouldAddEndPuncttrue
\mciteSetBstMidEndSepPunct{\mcitedefaultmidpunct}
{\mcitedefaultendpunct}{\mcitedefaultseppunct}\relax
\EndOfBibitem
\bibitem[Bian and Xie(2021)Bian, and Xie]{bian2021generative}
Bian,~Y.; Xie,~X.-Q. Generative chemistry: drug discovery with deep learning
  generative models. \emph{J. Mol. Model.} \textbf{2021}, \emph{27},
  1--18\relax
\mciteBstWouldAddEndPuncttrue
\mciteSetBstMidEndSepPunct{\mcitedefaultmidpunct}
{\mcitedefaultendpunct}{\mcitedefaultseppunct}\relax
\EndOfBibitem
\bibitem[Xiong \latin{et~al.}(2019)Xiong, Wang, Liu, Zhong, Wan, Li, Li, Luo,
  Chen, Jiang, \latin{et~al.} others]{xiong2019pushing}
Xiong,~Z.; Wang,~D.; Liu,~X.; Zhong,~F.; Wan,~X.; Li,~X.; Li,~Z.; Luo,~X.;
  Chen,~K.; Jiang,~H., \latin{et~al.}  Pushing the boundaries of molecular
  representation for drug discovery with the graph attention mechanism.
  \emph{J. Med. Chem.} \textbf{2019}, \emph{63}, 8749--8760\relax
\mciteBstWouldAddEndPuncttrue
\mciteSetBstMidEndSepPunct{\mcitedefaultmidpunct}
{\mcitedefaultendpunct}{\mcitedefaultseppunct}\relax
\EndOfBibitem
\bibitem[G{\'o}mez-Bombarelli \latin{et~al.}(2018)G{\'o}mez-Bombarelli, Wei,
  Duvenaud, Hern{\'a}ndez-Lobato, S{\'a}nchez-Lengeling, Sheberla,
  Aguilera-Iparraguirre, Hirzel, Adams, and Aspuru-Guzik]{gomez2018automatic}
G{\'o}mez-Bombarelli,~R.; Wei,~J.~N.; Duvenaud,~D.;
  Hern{\'a}ndez-Lobato,~J.~M.; S{\'a}nchez-Lengeling,~B.; Sheberla,~D.;
  Aguilera-Iparraguirre,~J.; Hirzel,~T.~D.; Adams,~R.~P.; Aspuru-Guzik,~A.
  Automatic chemical design using a data-driven continuous representation of
  molecules. \emph{ACS Cent. Sci.} \textbf{2018}, \emph{4}, 268--276\relax
\mciteBstWouldAddEndPuncttrue
\mciteSetBstMidEndSepPunct{\mcitedefaultmidpunct}
{\mcitedefaultendpunct}{\mcitedefaultseppunct}\relax
\EndOfBibitem
\bibitem[Popova \latin{et~al.}(2018)Popova, Isayev, and
  Tropsha]{popova2018deep}
Popova,~M.; Isayev,~O.; Tropsha,~A. Deep reinforcement learning for de novo
  drug design. \emph{Sci. Adv.} \textbf{2018}, \emph{4}, eaap7885\relax
\mciteBstWouldAddEndPuncttrue
\mciteSetBstMidEndSepPunct{\mcitedefaultmidpunct}
{\mcitedefaultendpunct}{\mcitedefaultseppunct}\relax
\EndOfBibitem
\bibitem[Ragoza \latin{et~al.}(2017)Ragoza, Hochuli, Idrobo, Sunseri, and
  Koes]{ragoza2017protein}
Ragoza,~M.; Hochuli,~J.; Idrobo,~E.; Sunseri,~J.; Koes,~D.~R. Protein--ligand
  scoring with convolutional neural networks. \emph{J. Chem. Inf. Model}
  \textbf{2017}, \emph{57}, 942--957\relax
\mciteBstWouldAddEndPuncttrue
\mciteSetBstMidEndSepPunct{\mcitedefaultmidpunct}
{\mcitedefaultendpunct}{\mcitedefaultseppunct}\relax
\EndOfBibitem
\bibitem[Jim{\'e}nez \latin{et~al.}(2018)Jim{\'e}nez, Skalic, Martinez-Rosell,
  and De~Fabritiis]{jimenez2018k}
Jim{\'e}nez,~J.; Skalic,~M.; Martinez-Rosell,~G.; De~Fabritiis,~G. K deep:
  protein--ligand absolute binding affinity prediction via 3d-convolutional
  neural networks. \emph{J. Chem. Inf. Model} \textbf{2018}, \emph{58},
  287--296\relax
\mciteBstWouldAddEndPuncttrue
\mciteSetBstMidEndSepPunct{\mcitedefaultmidpunct}
{\mcitedefaultendpunct}{\mcitedefaultseppunct}\relax
\EndOfBibitem
\bibitem[Lim \latin{et~al.}(2019)Lim, Ryu, Park, Choe, Ham, and
  Kim]{lim2019predicting}
Lim,~J.; Ryu,~S.; Park,~K.; Choe,~Y.~J.; Ham,~J.; Kim,~W.~Y. Predicting
  drug--target interaction using a novel graph neural network with 3D
  structure-embedded graph representation. \emph{J. Chem. Inf. Model}
  \textbf{2019}, \emph{59}, 3981--3988\relax
\mciteBstWouldAddEndPuncttrue
\mciteSetBstMidEndSepPunct{\mcitedefaultmidpunct}
{\mcitedefaultendpunct}{\mcitedefaultseppunct}\relax
\EndOfBibitem
\bibitem[Hernandez \latin{et~al.}(2019)Hernandez, Liang~Gan, Linvill, Dukatz,
  Feng, and Bhisetti]{hernandez2019quantum}
Hernandez,~M.; Liang~Gan,~G.; Linvill,~K.; Dukatz,~C.; Feng,~J.; Bhisetti,~G. A
  quantum-inspired method for three-dimensional ligand-based virtual screening.
  \emph{J. Chem. Inf. Model} \textbf{2019}, \emph{59}, 4475--4485\relax
\mciteBstWouldAddEndPuncttrue
\mciteSetBstMidEndSepPunct{\mcitedefaultmidpunct}
{\mcitedefaultendpunct}{\mcitedefaultseppunct}\relax
\EndOfBibitem
\bibitem[Wu and Wei(2018)Wu, and Wei]{wu2018quantitative}
Wu,~K.; Wei,~G.-W. Quantitative toxicity prediction using topology based
  multitask deep neural networks. \emph{J. Chem. Inf. Model} \textbf{2018},
  \emph{58}, 520--531\relax
\mciteBstWouldAddEndPuncttrue
\mciteSetBstMidEndSepPunct{\mcitedefaultmidpunct}
{\mcitedefaultendpunct}{\mcitedefaultseppunct}\relax
\EndOfBibitem
\bibitem[Skalic \latin{et~al.}(2019)Skalic, Jim{\'e}nez, Sabbadin, and
  De~Fabritiis]{skalic2019shape}
Skalic,~M.; Jim{\'e}nez,~J.; Sabbadin,~D.; De~Fabritiis,~G. Shape-based
  generative modeling for de novo drug design. \emph{J. Chem. Inf. Model}
  \textbf{2019}, \emph{59}, 1205--1214\relax
\mciteBstWouldAddEndPuncttrue
\mciteSetBstMidEndSepPunct{\mcitedefaultmidpunct}
{\mcitedefaultendpunct}{\mcitedefaultseppunct}\relax
\EndOfBibitem
\bibitem[Simm \latin{et~al.}(2020)Simm, Pinsler, and
  Hern{\'a}ndez-Lobato]{simm2020reinforcement}
Simm,~G.; Pinsler,~R.; Hern{\'a}ndez-Lobato,~J.~M. Reinforcement learning for
  molecular design guided by quantum mechanics. International Conference on
  Machine Learning. 2020; pp 8959--8969\relax
\mciteBstWouldAddEndPuncttrue
\mciteSetBstMidEndSepPunct{\mcitedefaultmidpunct}
{\mcitedefaultendpunct}{\mcitedefaultseppunct}\relax
\EndOfBibitem
\bibitem[Hemmerich \latin{et~al.}(2020)Hemmerich, Asilar, and
  Ecker]{hemmerich2020cover}
Hemmerich,~J.; Asilar,~E.; Ecker,~G.~F. COVER: conformational oversampling as
  data augmentation for molecules. \emph{J. Cheminformatics} \textbf{2020},
  \emph{12}, 1--12\relax
\mciteBstWouldAddEndPuncttrue
\mciteSetBstMidEndSepPunct{\mcitedefaultmidpunct}
{\mcitedefaultendpunct}{\mcitedefaultseppunct}\relax
\EndOfBibitem
\bibitem[Fernandez \latin{et~al.}(2018)Fernandez, Ban, Woo, Hsing, Yamazaki,
  LeBlanc, Rennie, Welch, and Cherkasov]{fernandez2018toxic}
Fernandez,~M.; Ban,~F.; Woo,~G.; Hsing,~M.; Yamazaki,~T.; LeBlanc,~E.;
  Rennie,~P.~S.; Welch,~W.~J.; Cherkasov,~A. Toxic colors: the use of deep
  learning for predicting toxicity of compounds merely from their graphic
  images. \emph{J. Chem. Inf. Model} \textbf{2018}, \emph{58}, 1533--1543\relax
\mciteBstWouldAddEndPuncttrue
\mciteSetBstMidEndSepPunct{\mcitedefaultmidpunct}
{\mcitedefaultendpunct}{\mcitedefaultseppunct}\relax
\EndOfBibitem
\bibitem[Meyer \latin{et~al.}(2019)Meyer, Liu, Miller, Coon, and
  Gitter]{meyer2019learning}
Meyer,~J.~G.; Liu,~S.; Miller,~I.~J.; Coon,~J.~J.; Gitter,~A. Learning drug
  functions from chemical structures with convolutional neural networks and
  random forests. \emph{J. Chem. Inf. Model} \textbf{2019}, \emph{59},
  4438--4449\relax
\mciteBstWouldAddEndPuncttrue
\mciteSetBstMidEndSepPunct{\mcitedefaultmidpunct}
{\mcitedefaultendpunct}{\mcitedefaultseppunct}\relax
\EndOfBibitem
\bibitem[Cort{\'e}s-Ciriano and Bender(2019)Cort{\'e}s-Ciriano, and
  Bender]{cortes2019kekulescope}
Cort{\'e}s-Ciriano,~I.; Bender,~A. KekuleScope: prediction of cancer cell line
  sensitivity and compound potency using convolutional neural networks trained
  on compound images. \emph{J. Cheminformatics} \textbf{2019}, \emph{11},
  1--16\relax
\mciteBstWouldAddEndPuncttrue
\mciteSetBstMidEndSepPunct{\mcitedefaultmidpunct}
{\mcitedefaultendpunct}{\mcitedefaultseppunct}\relax
\EndOfBibitem
\bibitem[Rifaioglu \latin{et~al.}(2020)Rifaioglu, Nalbat, Atalay, Martin,
  Cetin-Atalay, and Do{\u{g}}an]{rifaioglu2020deepscreen}
Rifaioglu,~A.~S.; Nalbat,~E.; Atalay,~V.; Martin,~M.~J.; Cetin-Atalay,~R.;
  Do{\u{g}}an,~T. DEEPScreen: high performance drug--target interaction
  prediction with convolutional neural networks using 2-D structural compound
  representations. \emph{Chem. Sci} \textbf{2020}, \emph{11}, 2531--2557\relax
\mciteBstWouldAddEndPuncttrue
\mciteSetBstMidEndSepPunct{\mcitedefaultmidpunct}
{\mcitedefaultendpunct}{\mcitedefaultseppunct}\relax
\EndOfBibitem
\bibitem[Rajan \latin{et~al.}(2021)Rajan, Brinkhaus, Sorokina, Zielesny, and
  Steinbeck]{rajan2021decimer}
Rajan,~K.; Brinkhaus,~H.~O.; Sorokina,~M.; Zielesny,~A.; Steinbeck,~C.
  DECIMER-Segmentation: Automated extraction of chemical structure depictions
  from scientific literature. \emph{J. Cheminformatics} \textbf{2021},
  \emph{13}, 1--9\relax
\mciteBstWouldAddEndPuncttrue
\mciteSetBstMidEndSepPunct{\mcitedefaultmidpunct}
{\mcitedefaultendpunct}{\mcitedefaultseppunct}\relax
\EndOfBibitem
\bibitem[Wu \latin{et~al.}(2018)Wu, Ramsundar, Feinberg, Gomes, Geniesse,
  Pappu, Leswing, and Pande]{wu2018moleculenet}
Wu,~Z.; Ramsundar,~B.; Feinberg,~E.~N.; Gomes,~J.; Geniesse,~C.; Pappu,~A.~S.;
  Leswing,~K.; Pande,~V. MoleculeNet: a benchmark for molecular machine
  learning. \emph{Chem. Sci} \textbf{2018}, \emph{9}, 513--530\relax
\mciteBstWouldAddEndPuncttrue
\mciteSetBstMidEndSepPunct{\mcitedefaultmidpunct}
{\mcitedefaultendpunct}{\mcitedefaultseppunct}\relax
\EndOfBibitem
\bibitem[Ramsundar \latin{et~al.}(2019)Ramsundar, Eastman, Walters, Pande,
  Leswing, and Wu]{Ramsundar-et-al-2019}
Ramsundar,~B.; Eastman,~P.; Walters,~P.; Pande,~V.; Leswing,~K.; Wu,~Z.
  \emph{Deep Learning for the Life Sciences}; O'Reilly Media, 2019;
  \url{https://www.amazon.com/Deep-Learning-Life-Sciences-Microscopy/dp/1492039837}\relax
\mciteBstWouldAddEndPuncttrue
\mciteSetBstMidEndSepPunct{\mcitedefaultmidpunct}
{\mcitedefaultendpunct}{\mcitedefaultseppunct}\relax
\EndOfBibitem
\bibitem[Fabian \latin{et~al.}(2020)Fabian, Edlich, Gaspar, Segler, Meyers,
  Fiscato, and Ahmed]{fabian2020molecular}
Fabian,~B.; Edlich,~T.; Gaspar,~H.; Segler,~M.; Meyers,~J.; Fiscato,~M.;
  Ahmed,~M. Molecular representation learning with language models and
  domain-relevant auxiliary tasks. \emph{arXiv preprint arXiv:2011.13230}
  \textbf{2020}, \relax
\mciteBstWouldAddEndPunctfalse
\mciteSetBstMidEndSepPunct{\mcitedefaultmidpunct}
{}{\mcitedefaultseppunct}\relax
\EndOfBibitem
\bibitem[Shen \latin{et~al.}(2021)Shen, Zeng, Zhu, li~Wang, Qin, Tan, Jiang,
  and Chen]{shen2021out}
Shen,~W.~X.; Zeng,~X.; Zhu,~F.; li~Wang,~Y.; Qin,~C.; Tan,~Y.; Jiang,~Y.~Y.;
  Chen,~Y.~Z. Out-of-the-box deep learning prediction of pharmaceutical
  properties by broadly learned knowledge-based molecular representations.
  \emph{Nat. Mach. Intell.} \textbf{2021}, \emph{3}, 334--343\relax
\mciteBstWouldAddEndPuncttrue
\mciteSetBstMidEndSepPunct{\mcitedefaultmidpunct}
{\mcitedefaultendpunct}{\mcitedefaultseppunct}\relax
\EndOfBibitem
\bibitem[Olivecrona \latin{et~al.}(2017)Olivecrona, Blaschke, Engkvist, and
  Chen]{olivecrona2017molecular}
Olivecrona,~M.; Blaschke,~T.; Engkvist,~O.; Chen,~H. Molecular de-novo design
  through deep reinforcement learning. \emph{J. Cheminformatics} \textbf{2017},
  \emph{9}, 1--14\relax
\mciteBstWouldAddEndPuncttrue
\mciteSetBstMidEndSepPunct{\mcitedefaultmidpunct}
{\mcitedefaultendpunct}{\mcitedefaultseppunct}\relax
\EndOfBibitem
\bibitem[Blaschke \latin{et~al.}(2020)Blaschke, Ar{\'u}s-Pous, Chen,
  Margreitter, Tyrchan, Engkvist, Papadopoulos, and
  Patronov]{blaschke2020reinvent}
Blaschke,~T.; Ar{\'u}s-Pous,~J.; Chen,~H.; Margreitter,~C.; Tyrchan,~C.;
  Engkvist,~O.; Papadopoulos,~K.; Patronov,~A. REINVENT 2.0: An AI Tool for De
  Novo Drug Design. \emph{J. Chem. Inf. Model} \textbf{2020}, \relax
\mciteBstWouldAddEndPunctfalse
\mciteSetBstMidEndSepPunct{\mcitedefaultmidpunct}
{}{\mcitedefaultseppunct}\relax
\EndOfBibitem
\bibitem[Brown \latin{et~al.}(2019)Brown, Fiscato, Segler, and
  Vaucher]{brown2019guacamol}
Brown,~N.; Fiscato,~M.; Segler,~M.~H.; Vaucher,~A.~C. GuacaMol: benchmarking
  models for de novo molecular design. \emph{J. Chem. Inf. Model}
  \textbf{2019}, \emph{59}, 1096--1108\relax
\mciteBstWouldAddEndPuncttrue
\mciteSetBstMidEndSepPunct{\mcitedefaultmidpunct}
{\mcitedefaultendpunct}{\mcitedefaultseppunct}\relax
\EndOfBibitem
\bibitem[Segler \latin{et~al.}(2018)Segler, Kogej, Tyrchan, and
  Waller]{segler2018generating}
Segler,~M.~H.; Kogej,~T.; Tyrchan,~C.; Waller,~M.~P. Generating focused
  molecule libraries for drug discovery with recurrent neural networks.
  \emph{ACS Cent. Sci.} \textbf{2018}, \emph{4}, 120--131\relax
\mciteBstWouldAddEndPuncttrue
\mciteSetBstMidEndSepPunct{\mcitedefaultmidpunct}
{\mcitedefaultendpunct}{\mcitedefaultseppunct}\relax
\EndOfBibitem
\bibitem[Walters and Murcko(2002)Walters, and Murcko]{walters2002prediction}
Walters,~W.~P.; Murcko,~M.~A. Prediction of ‘drug-likeness’. \emph{Adv.
  Drug Deliv. Rev.} \textbf{2002}, \emph{54}, 255--271\relax
\mciteBstWouldAddEndPuncttrue
\mciteSetBstMidEndSepPunct{\mcitedefaultmidpunct}
{\mcitedefaultendpunct}{\mcitedefaultseppunct}\relax
\EndOfBibitem
\bibitem[Schneider \latin{et~al.}(2008)Schneider, J{\"a}ckels, Andres, and
  Hutter]{schneider2008gradual}
Schneider,~N.; J{\"a}ckels,~C.; Andres,~C.; Hutter,~M.~C. Gradual in silico
  filtering for druglike substances. \emph{J. Chem. Inf. Model} \textbf{2008},
  \emph{48}, 613--628\relax
\mciteBstWouldAddEndPuncttrue
\mciteSetBstMidEndSepPunct{\mcitedefaultmidpunct}
{\mcitedefaultendpunct}{\mcitedefaultseppunct}\relax
\EndOfBibitem
\bibitem[Palmer \latin{et~al.}(2007)Palmer, O'Boyle, Glen, and
  Mitchell]{palmer2007random}
Palmer,~D.~S.; O'Boyle,~N.~M.; Glen,~R.~C.; Mitchell,~J.~B. Random forest
  models to predict aqueous solubility. \emph{J. Chem. Inf. Model}
  \textbf{2007}, \emph{47}, 150--158\relax
\mciteBstWouldAddEndPuncttrue
\mciteSetBstMidEndSepPunct{\mcitedefaultmidpunct}
{\mcitedefaultendpunct}{\mcitedefaultseppunct}\relax
\EndOfBibitem
\bibitem[Schroeter \latin{et~al.}(2007)Schroeter, Schwaighofer, Mika, Ter~Laak,
  Suelzle, Ganzer, Heinrich, and M{\"u}ller]{schroeter2007machine}
Schroeter,~T.; Schwaighofer,~A.; Mika,~S.; Ter~Laak,~A.; Suelzle,~D.;
  Ganzer,~U.; Heinrich,~N.; M{\"u}ller,~K.-R. Machine learning models for
  lipophilicity and their domain of applicability. \emph{Mol. Pharm.}
  \textbf{2007}, \emph{4}, 524--538\relax
\mciteBstWouldAddEndPuncttrue
\mciteSetBstMidEndSepPunct{\mcitedefaultmidpunct}
{\mcitedefaultendpunct}{\mcitedefaultseppunct}\relax
\EndOfBibitem
\bibitem[Hou \latin{et~al.}(2007)Hou, Wang, and Li]{hou2007adme}
Hou,~T.; Wang,~J.; Li,~Y. ADME evaluation in drug discovery. 8. The prediction
  of human intestinal absorption by a support vector machine. \emph{J. Chem.
  Inf. Model} \textbf{2007}, \emph{47}, 2408--2415\relax
\mciteBstWouldAddEndPuncttrue
\mciteSetBstMidEndSepPunct{\mcitedefaultmidpunct}
{\mcitedefaultendpunct}{\mcitedefaultseppunct}\relax
\EndOfBibitem
\bibitem[Tian \latin{et~al.}(2011)Tian, Li, Wang, Zhang, and Hou]{tian2011adme}
Tian,~S.; Li,~Y.; Wang,~J.; Zhang,~J.; Hou,~T. ADME evaluation in drug
  discovery. 9. Prediction of oral bioavailability in humans based on molecular
  properties and structural fingerprints. \emph{Mol. Pharm.} \textbf{2011},
  \emph{8}, 841--851\relax
\mciteBstWouldAddEndPuncttrue
\mciteSetBstMidEndSepPunct{\mcitedefaultmidpunct}
{\mcitedefaultendpunct}{\mcitedefaultseppunct}\relax
\EndOfBibitem
\bibitem[Sakiyama \latin{et~al.}(2008)Sakiyama, Yuki, Moriya, Hattori, Suzuki,
  Shimada, and Honma]{sakiyama2008predicting}
Sakiyama,~Y.; Yuki,~H.; Moriya,~T.; Hattori,~K.; Suzuki,~M.; Shimada,~K.;
  Honma,~T. Predicting human liver microsomal stability with machine learning
  techniques. \emph{J. Mol. Graph. Model.} \textbf{2008}, \emph{26},
  907--915\relax
\mciteBstWouldAddEndPuncttrue
\mciteSetBstMidEndSepPunct{\mcitedefaultmidpunct}
{\mcitedefaultendpunct}{\mcitedefaultseppunct}\relax
\EndOfBibitem
\bibitem[Vasanthanathan \latin{et~al.}(2009)Vasanthanathan, Taboureau,
  Oostenbrink, Vermeulen, Olsen, and
  J{\o}rgensen]{vasanthanathan2009classification}
Vasanthanathan,~P.; Taboureau,~O.; Oostenbrink,~C.; Vermeulen,~N.~P.;
  Olsen,~L.; J{\o}rgensen,~F.~S. Classification of cytochrome P450 1A2
  inhibitors and noninhibitors by machine learning techniques. \emph{Drug
  Metab. Dispos} \textbf{2009}, \emph{37}, 658--664\relax
\mciteBstWouldAddEndPuncttrue
\mciteSetBstMidEndSepPunct{\mcitedefaultmidpunct}
{\mcitedefaultendpunct}{\mcitedefaultseppunct}\relax
\EndOfBibitem
\bibitem[Riddick \latin{et~al.}(2011)Riddick, Song, Ahn, Walling,
  Borges-Rivera, Zhang, and Fine]{riddick2011predicting}
Riddick,~G.; Song,~H.; Ahn,~S.; Walling,~J.; Borges-Rivera,~D.; Zhang,~W.;
  Fine,~H.~A. Predicting in vitro drug sensitivity using Random Forests.
  \emph{Bioinformatics} \textbf{2011}, \emph{27}, 220--224\relax
\mciteBstWouldAddEndPuncttrue
\mciteSetBstMidEndSepPunct{\mcitedefaultmidpunct}
{\mcitedefaultendpunct}{\mcitedefaultseppunct}\relax
\EndOfBibitem
\bibitem[Zhao \latin{et~al.}(2006)Zhao, Zhang, Zhang, Liu, Hu, and
  Fan]{zhao2006application}
Zhao,~C.; Zhang,~H.; Zhang,~X.; Liu,~M.; Hu,~Z.; Fan,~B. Application of support
  vector machine (SVM) for prediction toxic activity of different data sets.
  \emph{Toxicology} \textbf{2006}, \emph{217}, 105--119\relax
\mciteBstWouldAddEndPuncttrue
\mciteSetBstMidEndSepPunct{\mcitedefaultmidpunct}
{\mcitedefaultendpunct}{\mcitedefaultseppunct}\relax
\EndOfBibitem
\bibitem[Heikamp and Bajorath(2014)Heikamp, and Bajorath]{heikamp2014support}
Heikamp,~K.; Bajorath,~J. Support vector machines for drug discovery.
  \emph{Expert Opin Drug Discov} \textbf{2014}, \emph{9}, 93--104\relax
\mciteBstWouldAddEndPuncttrue
\mciteSetBstMidEndSepPunct{\mcitedefaultmidpunct}
{\mcitedefaultendpunct}{\mcitedefaultseppunct}\relax
\EndOfBibitem
\bibitem[Svetnik \latin{et~al.}(2003)Svetnik, Liaw, Tong, Culberson, Sheridan,
  and Feuston]{svetnik2003random}
Svetnik,~V.; Liaw,~A.; Tong,~C.; Culberson,~J.~C.; Sheridan,~R.~P.;
  Feuston,~B.~P. Random forest: a classification and regression tool for
  compound classification and QSAR modeling. \emph{J Chem Inform Comput Sci}
  \textbf{2003}, \emph{43}, 1947--1958\relax
\mciteBstWouldAddEndPuncttrue
\mciteSetBstMidEndSepPunct{\mcitedefaultmidpunct}
{\mcitedefaultendpunct}{\mcitedefaultseppunct}\relax
\EndOfBibitem
\bibitem[Dahl(2012)]{dahl2012deep}
Dahl,~G. Deep learning how I did it: Merck 1st place interview. \emph{Online
  article available from http://blog. kaggle.
  com/2012/11/01/deep-learning-how-i-did-it-merck-1st-place-interview}
  \textbf{2012}, \relax
\mciteBstWouldAddEndPunctfalse
\mciteSetBstMidEndSepPunct{\mcitedefaultmidpunct}
{}{\mcitedefaultseppunct}\relax
\EndOfBibitem
\bibitem[LeCun \latin{et~al.}(2015)LeCun, Bengio, and Hinton]{lecun2015deep}
LeCun,~Y.; Bengio,~Y.; Hinton,~G. Deep learning. \emph{Nature} \textbf{2015},
  \emph{521}, 436--444\relax
\mciteBstWouldAddEndPuncttrue
\mciteSetBstMidEndSepPunct{\mcitedefaultmidpunct}
{\mcitedefaultendpunct}{\mcitedefaultseppunct}\relax
\EndOfBibitem
\bibitem[Simm \latin{et~al.}(2018)Simm, Klambauer, Arany, Steijaert, Wegner,
  Gustin, Chupakhin, Chong, Vialard, Buijnsters, \latin{et~al.}
  others]{simm2018repurposing}
Simm,~J.; Klambauer,~G.; Arany,~A.; Steijaert,~M.; Wegner,~J.~K.; Gustin,~E.;
  Chupakhin,~V.; Chong,~Y.~T.; Vialard,~J.; Buijnsters,~P., \latin{et~al.}
  Repurposing high-throughput image assays enables biological activity
  prediction for drug discovery. \emph{Cell Chem. Biol.} \textbf{2018},
  \emph{25}, 611--618\relax
\mciteBstWouldAddEndPuncttrue
\mciteSetBstMidEndSepPunct{\mcitedefaultmidpunct}
{\mcitedefaultendpunct}{\mcitedefaultseppunct}\relax
\EndOfBibitem
\bibitem[Hofmarcher \latin{et~al.}(2019)Hofmarcher, Rumetshofer, Clevert,
  Hochreiter, and Klambauer]{hofmarcher2019accurate}
Hofmarcher,~M.; Rumetshofer,~E.; Clevert,~D.-A.; Hochreiter,~S.; Klambauer,~G.
  Accurate prediction of biological assays with high-throughput microscopy
  images and convolutional networks. \emph{J. Chem. Inf. Model} \textbf{2019},
  \emph{59}, 1163--1171\relax
\mciteBstWouldAddEndPuncttrue
\mciteSetBstMidEndSepPunct{\mcitedefaultmidpunct}
{\mcitedefaultendpunct}{\mcitedefaultseppunct}\relax
\EndOfBibitem
\bibitem[Ramsundar \latin{et~al.}(2015)Ramsundar, Kearnes, Riley, Webster,
  Konerding, and Pande]{ramsundar2015massively}
Ramsundar,~B.; Kearnes,~S.; Riley,~P.; Webster,~D.; Konerding,~D.; Pande,~V.
  Massively multitask networks for drug discovery. \emph{arXiv preprint
  arXiv:1502.02072} \textbf{2015}, \relax
\mciteBstWouldAddEndPunctfalse
\mciteSetBstMidEndSepPunct{\mcitedefaultmidpunct}
{}{\mcitedefaultseppunct}\relax
\EndOfBibitem
\bibitem[Duvenaud \latin{et~al.}(2015)Duvenaud, Maclaurin,
  Aguilera-Iparraguirre, G{\'o}mez-Bombarelli, Hirzel, Aspuru-Guzik, and
  Adams]{duvenaud2015convolutional}
Duvenaud,~D.; Maclaurin,~D.; Aguilera-Iparraguirre,~J.;
  G{\'o}mez-Bombarelli,~R.; Hirzel,~T.; Aspuru-Guzik,~A.; Adams,~R.~P.
  Convolutional networks on graphs for learning molecular fingerprints.
  \emph{arXiv preprint arXiv:1509.09292} \textbf{2015}, \relax
\mciteBstWouldAddEndPunctfalse
\mciteSetBstMidEndSepPunct{\mcitedefaultmidpunct}
{}{\mcitedefaultseppunct}\relax
\EndOfBibitem
\bibitem[Glen \latin{et~al.}(2006)Glen, Bender, Arnby, Carlsson, Boyer, and
  Smith]{glen2006circular}
Glen,~R.~C.; Bender,~A.; Arnby,~C.~H.; Carlsson,~L.; Boyer,~S.; Smith,~J.
  Circular fingerprints: flexible molecular descriptors with applications from
  physical chemistry to ADME. \emph{IDrugs} \textbf{2006}, \emph{9}, 199\relax
\mciteBstWouldAddEndPuncttrue
\mciteSetBstMidEndSepPunct{\mcitedefaultmidpunct}
{\mcitedefaultendpunct}{\mcitedefaultseppunct}\relax
\EndOfBibitem
\bibitem[Goh \latin{et~al.}(2017)Goh, Siegel, Vishnu, Hodas, and
  Baker]{goh2017chemception}
Goh,~G.~B.; Siegel,~C.; Vishnu,~A.; Hodas,~N.~O.; Baker,~N. Chemception: a deep
  neural network with minimal chemistry knowledge matches the performance of
  expert-developed QSAR/QSPR models. \emph{arXiv preprint arXiv:1706.06689}
  \textbf{2017}, \relax
\mciteBstWouldAddEndPunctfalse
\mciteSetBstMidEndSepPunct{\mcitedefaultmidpunct}
{}{\mcitedefaultseppunct}\relax
\EndOfBibitem
\bibitem[Huang \latin{et~al.}(2017)Huang, Liu, Van Der~Maaten, and
  Weinberger]{huang2017densely}
Huang,~G.; Liu,~Z.; Van Der~Maaten,~L.; Weinberger,~K.~Q. Densely connected
  convolutional networks. Proceedings of the IEEE Conference on Computer Vision
  and Pattern Recognition. 2017; pp 4700--4708\relax
\mciteBstWouldAddEndPuncttrue
\mciteSetBstMidEndSepPunct{\mcitedefaultmidpunct}
{\mcitedefaultendpunct}{\mcitedefaultseppunct}\relax
\EndOfBibitem
\bibitem[He \latin{et~al.}(2016)He, Zhang, Ren, and Sun]{he2016deep}
He,~K.; Zhang,~X.; Ren,~S.; Sun,~J. Deep residual learning for image
  recognition. Proceedings of the IEEE Conference on Computer Vision and
  Pattern Recognition. 2016; pp 770--778\relax
\mciteBstWouldAddEndPuncttrue
\mciteSetBstMidEndSepPunct{\mcitedefaultmidpunct}
{\mcitedefaultendpunct}{\mcitedefaultseppunct}\relax
\EndOfBibitem
\bibitem[Simonyan and Zisserman(2014)Simonyan, and Zisserman]{simonyan2014very}
Simonyan,~K.; Zisserman,~A. Very deep convolutional networks for large-scale
  image recognition. \emph{arXiv preprint arXiv:1409.1556} \textbf{2014},
  \relax
\mciteBstWouldAddEndPunctfalse
\mciteSetBstMidEndSepPunct{\mcitedefaultmidpunct}
{}{\mcitedefaultseppunct}\relax
\EndOfBibitem
\bibitem[Staker \latin{et~al.}(2019)Staker, Marshall, Abel, and
  McQuaw]{staker2019molecular}
Staker,~J.; Marshall,~K.; Abel,~R.; McQuaw,~C.~M. Molecular structure
  extraction from documents using deep learning. \emph{J. Chem. Inf. Model}
  \textbf{2019}, \emph{59}, 1017--1029\relax
\mciteBstWouldAddEndPuncttrue
\mciteSetBstMidEndSepPunct{\mcitedefaultmidpunct}
{\mcitedefaultendpunct}{\mcitedefaultseppunct}\relax
\EndOfBibitem
\bibitem[Rajan \latin{et~al.}(2020)Rajan, Zielesny, and
  Steinbeck]{rajan2020decimer}
Rajan,~K.; Zielesny,~A.; Steinbeck,~C. DECIMER: towards deep learning for
  chemical image recognition. \emph{J. Cheminformatics} \textbf{2020},
  \emph{12}, 1--9\relax
\mciteBstWouldAddEndPuncttrue
\mciteSetBstMidEndSepPunct{\mcitedefaultmidpunct}
{\mcitedefaultendpunct}{\mcitedefaultseppunct}\relax
\EndOfBibitem
\bibitem[Hossain \latin{et~al.}(2019)Hossain, Sohel, Shiratuddin, and
  Laga]{hossain2019comprehensive}
Hossain,~M.~Z.; Sohel,~F.; Shiratuddin,~M.~F.; Laga,~H. A comprehensive survey
  of deep learning for image captioning. \emph{ACM Computing Surveys (CsUR)}
  \textbf{2019}, \emph{51}, 1--36\relax
\mciteBstWouldAddEndPuncttrue
\mciteSetBstMidEndSepPunct{\mcitedefaultmidpunct}
{\mcitedefaultendpunct}{\mcitedefaultseppunct}\relax
\EndOfBibitem
\bibitem[Mikolov \latin{et~al.}(2011)Mikolov, Kombrink, Burget,
  {\v{C}}ernock{\`y}, and Khudanpur]{mikolov2011extensions}
Mikolov,~T.; Kombrink,~S.; Burget,~L.; {\v{C}}ernock{\`y},~J.; Khudanpur,~S.
  Extensions of recurrent neural network language model. 2011 IEEE
  International Conference on Acoustics, Speech and Signal Processing (ICASSP).
  2011; pp 5528--5531\relax
\mciteBstWouldAddEndPuncttrue
\mciteSetBstMidEndSepPunct{\mcitedefaultmidpunct}
{\mcitedefaultendpunct}{\mcitedefaultseppunct}\relax
\EndOfBibitem
\bibitem[Boulanger-Lewandowski \latin{et~al.}(2012)Boulanger-Lewandowski,
  Bengio, and Vincent]{boulanger2012modeling}
Boulanger-Lewandowski,~N.; Bengio,~Y.; Vincent,~P. Modeling temporal
  dependencies in high-dimensional sequences: Application to polyphonic music
  generation and transcription. \emph{arXiv preprint arXiv:1206.6392}
  \textbf{2012}, \relax
\mciteBstWouldAddEndPunctfalse
\mciteSetBstMidEndSepPunct{\mcitedefaultmidpunct}
{}{\mcitedefaultseppunct}\relax
\EndOfBibitem
\bibitem[Hochreiter and Schmidhuber(1997)Hochreiter, and
  Schmidhuber]{hochreiter1997long}
Hochreiter,~S.; Schmidhuber,~J. Long short-term memory. \emph{Neural Comput.}
  \textbf{1997}, \emph{9}, 1735--1780\relax
\mciteBstWouldAddEndPuncttrue
\mciteSetBstMidEndSepPunct{\mcitedefaultmidpunct}
{\mcitedefaultendpunct}{\mcitedefaultseppunct}\relax
\EndOfBibitem
\bibitem[Chung \latin{et~al.}(2014)Chung, Gulcehre, Cho, and
  Bengio]{chung2014empirical}
Chung,~J.; Gulcehre,~C.; Cho,~K.; Bengio,~Y. Empirical evaluation of gated
  recurrent neural networks on sequence modeling. \emph{arXiv preprint
  arXiv:1412.3555} \textbf{2014}, \relax
\mciteBstWouldAddEndPunctfalse
\mciteSetBstMidEndSepPunct{\mcitedefaultmidpunct}
{}{\mcitedefaultseppunct}\relax
\EndOfBibitem
\bibitem[Goh \latin{et~al.}(2017)Goh, Hodas, Siegel, and
  Vishnu]{goh2017smiles2vec}
Goh,~G.~B.; Hodas,~N.~O.; Siegel,~C.; Vishnu,~A. Smiles2vec: An interpretable
  general-purpose deep neural network for predicting chemical properties.
  \emph{arXiv preprint arXiv:1712.02034} \textbf{2017}, \relax
\mciteBstWouldAddEndPunctfalse
\mciteSetBstMidEndSepPunct{\mcitedefaultmidpunct}
{}{\mcitedefaultseppunct}\relax
\EndOfBibitem
\bibitem[Neil \latin{et~al.}(2018)Neil, Segler, Guasch, Ahmed, Plumbley,
  Sellwood, and Brown]{neil2018exploring}
Neil,~D.; Segler,~M.; Guasch,~L.; Ahmed,~M.; Plumbley,~D.; Sellwood,~M.;
  Brown,~N. Exploring deep recurrent models with reinforcement learning for
  molecule design. Proceedings of The International Conference on Learning
  Representations. 2018\relax
\mciteBstWouldAddEndPuncttrue
\mciteSetBstMidEndSepPunct{\mcitedefaultmidpunct}
{\mcitedefaultendpunct}{\mcitedefaultseppunct}\relax
\EndOfBibitem
\bibitem[Joulin and Mikolov(2015)Joulin, and Mikolov]{joulin2015inferring}
Joulin,~A.; Mikolov,~T. Inferring algorithmic patterns with stack-augmented
  recurrent nets. \emph{arXiv preprint arXiv:1503.01007} \textbf{2015}, \relax
\mciteBstWouldAddEndPunctfalse
\mciteSetBstMidEndSepPunct{\mcitedefaultmidpunct}
{}{\mcitedefaultseppunct}\relax
\EndOfBibitem
\bibitem[St{\aa}hl \latin{et~al.}(2019)St{\aa}hl, Falkman, Karlsson, Mathiason,
  and Bostrom]{staahl2019deep}
St{\aa}hl,~N.; Falkman,~G.; Karlsson,~A.; Mathiason,~G.; Bostrom,~J. Deep
  reinforcement learning for multiparameter optimization in de novo drug
  design. \emph{J. Chem. Inf. Model} \textbf{2019}, \emph{59}, 3166--3176\relax
\mciteBstWouldAddEndPuncttrue
\mciteSetBstMidEndSepPunct{\mcitedefaultmidpunct}
{\mcitedefaultendpunct}{\mcitedefaultseppunct}\relax
\EndOfBibitem
\bibitem[Zheng \latin{et~al.}(2019)Zheng, Yan, Yang, and
  Xu]{zheng2019identifying}
Zheng,~S.; Yan,~X.; Yang,~Y.; Xu,~J. Identifying structure--property
  relationships through SMILES syntax analysis with self-attention mechanism.
  \emph{J. Chem. Inf. Model} \textbf{2019}, \emph{59}, 914--923\relax
\mciteBstWouldAddEndPuncttrue
\mciteSetBstMidEndSepPunct{\mcitedefaultmidpunct}
{\mcitedefaultendpunct}{\mcitedefaultseppunct}\relax
\EndOfBibitem
\bibitem[You \latin{et~al.}(2018)You, Ying, Ren, Hamilton, and
  Leskovec]{you2018graphrnn}
You,~J.; Ying,~R.; Ren,~X.; Hamilton,~W.; Leskovec,~J. Graphrnn: Generating
  realistic graphs with deep auto-regressive models. International Conference
  on Machine Learning. 2018; pp 5708--5717\relax
\mciteBstWouldAddEndPuncttrue
\mciteSetBstMidEndSepPunct{\mcitedefaultmidpunct}
{\mcitedefaultendpunct}{\mcitedefaultseppunct}\relax
\EndOfBibitem
\bibitem[Li \latin{et~al.}(2018)Li, Vinyals, Dyer, Pascanu, and
  Battaglia]{li2018learning}
Li,~Y.; Vinyals,~O.; Dyer,~C.; Pascanu,~R.; Battaglia,~P. Learning deep
  generative models of graphs. \emph{arXiv preprint arXiv:1803.03324}
  \textbf{2018}, \relax
\mciteBstWouldAddEndPunctfalse
\mciteSetBstMidEndSepPunct{\mcitedefaultmidpunct}
{}{\mcitedefaultseppunct}\relax
\EndOfBibitem
\bibitem[Li \latin{et~al.}(2018)Li, Zhang, and Liu]{li2018multi}
Li,~Y.; Zhang,~L.; Liu,~Z. Multi-objective de novo drug design with conditional
  graph generative model. \emph{J. Cheminformatics} \textbf{2018}, \emph{10},
  1--24\relax
\mciteBstWouldAddEndPuncttrue
\mciteSetBstMidEndSepPunct{\mcitedefaultmidpunct}
{\mcitedefaultendpunct}{\mcitedefaultseppunct}\relax
\EndOfBibitem
\bibitem[Popova \latin{et~al.}(2019)Popova, Shvets, Oliva, and
  Isayev]{popova2019molecularrnn}
Popova,~M.; Shvets,~M.; Oliva,~J.; Isayev,~O. MolecularRNN: Generating
  realistic molecular graphs with optimized properties. \emph{arXiv preprint
  arXiv:1905.13372} \textbf{2019}, \relax
\mciteBstWouldAddEndPunctfalse
\mciteSetBstMidEndSepPunct{\mcitedefaultmidpunct}
{}{\mcitedefaultseppunct}\relax
\EndOfBibitem
\bibitem[You \latin{et~al.}(2018)You, Liu, Ying, Pande, and
  Leskovec]{you2018graph}
You,~J.; Liu,~B.; Ying,~R.; Pande,~V.; Leskovec,~J. Graph convolutional policy
  network for goal-directed molecular graph generation. \emph{arXiv preprint
  arXiv:1806.02473} \textbf{2018}, \relax
\mciteBstWouldAddEndPunctfalse
\mciteSetBstMidEndSepPunct{\mcitedefaultmidpunct}
{}{\mcitedefaultseppunct}\relax
\EndOfBibitem
\bibitem[Sattarov \latin{et~al.}(2019)Sattarov, Baskin, Horvath, Marcou,
  Bjerrum, and Varnek]{sattarov2019novo}
Sattarov,~B.; Baskin,~I.~I.; Horvath,~D.; Marcou,~G.; Bjerrum,~E.~J.;
  Varnek,~A. De novo molecular design by combining deep autoencoder recurrent
  neural networks with generative topographic mapping. \emph{J. Chem. Inf.
  Model} \textbf{2019}, \emph{59}, 1182--1196\relax
\mciteBstWouldAddEndPuncttrue
\mciteSetBstMidEndSepPunct{\mcitedefaultmidpunct}
{\mcitedefaultendpunct}{\mcitedefaultseppunct}\relax
\EndOfBibitem
\bibitem[Guimaraes \latin{et~al.}(2017)Guimaraes, Sanchez-Lengeling, Outeiral,
  Farias, and Aspuru-Guzik]{guimaraes2017objective}
Guimaraes,~G.~L.; Sanchez-Lengeling,~B.; Outeiral,~C.; Farias,~P. L.~C.;
  Aspuru-Guzik,~A. Objective-reinforced generative adversarial networks (ORGAN)
  for sequence generation models. \emph{arXiv preprint arXiv:1705.10843}
  \textbf{2017}, \relax
\mciteBstWouldAddEndPunctfalse
\mciteSetBstMidEndSepPunct{\mcitedefaultmidpunct}
{}{\mcitedefaultseppunct}\relax
\EndOfBibitem
\bibitem[Sanchez-Lengeling \latin{et~al.}(2017)Sanchez-Lengeling, Outeiral,
  Guimaraes, and Aspuru-Guzik]{sanchez2017optimizing}
Sanchez-Lengeling,~B.; Outeiral,~C.; Guimaraes,~G.~L.; Aspuru-Guzik,~A.
  Optimizing distributions over molecular space. An objective-reinforced
  generative adversarial network for inverse-design chemistry (ORGANIC).
  \emph{ChemRxiv} \textbf{2017}, \emph{2017}\relax
\mciteBstWouldAddEndPuncttrue
\mciteSetBstMidEndSepPunct{\mcitedefaultmidpunct}
{\mcitedefaultendpunct}{\mcitedefaultseppunct}\relax
\EndOfBibitem
\bibitem[Wu \latin{et~al.}(2020)Wu, Pan, Chen, Long, Zhang, and
  Philip]{wu2020comprehensive}
Wu,~Z.; Pan,~S.; Chen,~F.; Long,~G.; Zhang,~C.; Philip,~S.~Y. A comprehensive
  survey on graph neural networks. \emph{IEEE Trans Neural Netw Learn}
  \textbf{2020}, \relax
\mciteBstWouldAddEndPunctfalse
\mciteSetBstMidEndSepPunct{\mcitedefaultmidpunct}
{}{\mcitedefaultseppunct}\relax
\EndOfBibitem
\bibitem[Li \latin{et~al.}(2015)Li, Tarlow, Brockschmidt, and
  Zemel]{li2015gated}
Li,~Y.; Tarlow,~D.; Brockschmidt,~M.; Zemel,~R. Gated graph sequence neural
  networks. \emph{arXiv preprint arXiv:1511.05493} \textbf{2015}, \relax
\mciteBstWouldAddEndPunctfalse
\mciteSetBstMidEndSepPunct{\mcitedefaultmidpunct}
{}{\mcitedefaultseppunct}\relax
\EndOfBibitem
\bibitem[Defferrard \latin{et~al.}(2016)Defferrard, Bresson, and
  Vandergheynst]{defferrard2016convolutional}
Defferrard,~M.; Bresson,~X.; Vandergheynst,~P. Convolutional neural networks on
  graphs with fast localized spectral filtering. \emph{arXiv preprint
  arXiv:1606.09375} \textbf{2016}, \relax
\mciteBstWouldAddEndPunctfalse
\mciteSetBstMidEndSepPunct{\mcitedefaultmidpunct}
{}{\mcitedefaultseppunct}\relax
\EndOfBibitem
\bibitem[Kipf and Welling(2016)Kipf, and Welling]{kipf2016semi}
Kipf,~T.~N.; Welling,~M. Semi-supervised classification with graph
  convolutional networks. \emph{arXiv preprint arXiv:1609.02907} \textbf{2016},
  \relax
\mciteBstWouldAddEndPunctfalse
\mciteSetBstMidEndSepPunct{\mcitedefaultmidpunct}
{}{\mcitedefaultseppunct}\relax
\EndOfBibitem
\bibitem[Gilmer \latin{et~al.}(2017)Gilmer, Schoenholz, Riley, Vinyals, and
  Dahl]{gilmer2017neural}
Gilmer,~J.; Schoenholz,~S.~S.; Riley,~P.~F.; Vinyals,~O.; Dahl,~G.~E. Neural
  message passing for quantum chemistry. International Conference on Machine
  Learning. 2017; pp 1263--1272\relax
\mciteBstWouldAddEndPuncttrue
\mciteSetBstMidEndSepPunct{\mcitedefaultmidpunct}
{\mcitedefaultendpunct}{\mcitedefaultseppunct}\relax
\EndOfBibitem
\bibitem[Hamilton \latin{et~al.}(2017)Hamilton, Ying, and
  Leskovec]{hamilton2017inductive}
Hamilton,~W.~L.; Ying,~R.; Leskovec,~J. Inductive representation learning on
  large graphs. \emph{arXiv preprint arXiv:1706.02216} \textbf{2017}, \relax
\mciteBstWouldAddEndPunctfalse
\mciteSetBstMidEndSepPunct{\mcitedefaultmidpunct}
{}{\mcitedefaultseppunct}\relax
\EndOfBibitem
\bibitem[Veli{\v{c}}kovi{\'c} \latin{et~al.}(2017)Veli{\v{c}}kovi{\'c},
  Cucurull, Casanova, Romero, Lio, and Bengio]{velivckovic2017graph}
Veli{\v{c}}kovi{\'c},~P.; Cucurull,~G.; Casanova,~A.; Romero,~A.; Lio,~P.;
  Bengio,~Y. Graph attention networks. \emph{arXiv preprint arXiv:1710.10903}
  \textbf{2017}, \relax
\mciteBstWouldAddEndPunctfalse
\mciteSetBstMidEndSepPunct{\mcitedefaultmidpunct}
{}{\mcitedefaultseppunct}\relax
\EndOfBibitem
\bibitem[Xu \latin{et~al.}(2018)Xu, Hu, Leskovec, and Jegelka]{xu2018powerful}
Xu,~K.; Hu,~W.; Leskovec,~J.; Jegelka,~S. How powerful are graph neural
  networks? \emph{arXiv preprint arXiv:1810.00826} \textbf{2018}, \relax
\mciteBstWouldAddEndPunctfalse
\mciteSetBstMidEndSepPunct{\mcitedefaultmidpunct}
{}{\mcitedefaultseppunct}\relax
\EndOfBibitem
\bibitem[Kearnes \latin{et~al.}(2016)Kearnes, McCloskey, Berndl, Pande, and
  Riley]{kearnes2016molecular}
Kearnes,~S.; McCloskey,~K.; Berndl,~M.; Pande,~V.; Riley,~P. Molecular graph
  convolutions: moving beyond fingerprints. \emph{J. Comput. Aided Mol. Des.}
  \textbf{2016}, \emph{30}, 595--608\relax
\mciteBstWouldAddEndPuncttrue
\mciteSetBstMidEndSepPunct{\mcitedefaultmidpunct}
{\mcitedefaultendpunct}{\mcitedefaultseppunct}\relax
\EndOfBibitem
\bibitem[Landrum(2016)]{Landrum2016RDKit2016_09_4}
Landrum,~G. RDKit: Open-Source Cheminformatics Software. \emph{RDKit}
  \textbf{2016}, \relax
\mciteBstWouldAddEndPunctfalse
\mciteSetBstMidEndSepPunct{\mcitedefaultmidpunct}
{}{\mcitedefaultseppunct}\relax
\EndOfBibitem
\bibitem[Withnall \latin{et~al.}(2020)Withnall, Lindel{\"o}f, Engkvist, and
  Chen]{withnall2020building}
Withnall,~M.; Lindel{\"o}f,~E.; Engkvist,~O.; Chen,~H. Building attention and
  edge message passing neural networks for bioactivity and physical--chemical
  property prediction. \emph{J. Cheminformatics} \textbf{2020}, \emph{12},
  1--18\relax
\mciteBstWouldAddEndPuncttrue
\mciteSetBstMidEndSepPunct{\mcitedefaultmidpunct}
{\mcitedefaultendpunct}{\mcitedefaultseppunct}\relax
\EndOfBibitem
\bibitem[Sch{\"u}tt \latin{et~al.}(2017)Sch{\"u}tt, Kindermans, Sauceda,
  Chmiela, Tkatchenko, and M{\"u}ller]{schutt2017schnet}
Sch{\"u}tt,~K.~T.; Kindermans,~P.-J.; Sauceda,~H.~E.; Chmiela,~S.;
  Tkatchenko,~A.; M{\"u}ller,~K.-R. Schnet: A continuous-filter convolutional
  neural network for modeling quantum interactions. \emph{arXiv preprint
  arXiv:1706.08566} \textbf{2017}, \relax
\mciteBstWouldAddEndPunctfalse
\mciteSetBstMidEndSepPunct{\mcitedefaultmidpunct}
{}{\mcitedefaultseppunct}\relax
\EndOfBibitem
\bibitem[Feinberg \latin{et~al.}(2018)Feinberg, Sur, Wu, Husic, Mai, Li, Sun,
  Yang, Ramsundar, and Pande]{feinberg2018potentialnet}
Feinberg,~E.~N.; Sur,~D.; Wu,~Z.; Husic,~B.~E.; Mai,~H.; Li,~Y.; Sun,~S.;
  Yang,~J.; Ramsundar,~B.; Pande,~V.~S. PotentialNet for molecular property
  prediction. \emph{ACS Cent. Sci.} \textbf{2018}, \emph{4}, 1520--1530\relax
\mciteBstWouldAddEndPuncttrue
\mciteSetBstMidEndSepPunct{\mcitedefaultmidpunct}
{\mcitedefaultendpunct}{\mcitedefaultseppunct}\relax
\EndOfBibitem
\bibitem[Klicpera \latin{et~al.}(2020)Klicpera, Gro{\ss}, and
  G{\"u}nnemann]{klicpera2020directional}
Klicpera,~J.; Gro{\ss},~J.; G{\"u}nnemann,~S. Directional message passing for
  molecular graphs. \emph{arXiv preprint arXiv:2003.03123} \textbf{2020},
  \relax
\mciteBstWouldAddEndPunctfalse
\mciteSetBstMidEndSepPunct{\mcitedefaultmidpunct}
{}{\mcitedefaultseppunct}\relax
\EndOfBibitem
\bibitem[Altae-Tran \latin{et~al.}(2017)Altae-Tran, Ramsundar, Pappu, and
  Pande]{altae2017low}
Altae-Tran,~H.; Ramsundar,~B.; Pappu,~A.~S.; Pande,~V. Low data drug discovery
  with one-shot learning. \emph{ACS Cent. Sci.} \textbf{2017}, \emph{3},
  283--293\relax
\mciteBstWouldAddEndPuncttrue
\mciteSetBstMidEndSepPunct{\mcitedefaultmidpunct}
{\mcitedefaultendpunct}{\mcitedefaultseppunct}\relax
\EndOfBibitem
\bibitem[Liu \latin{et~al.}(2018)Liu, Demirel, and Liang]{liu2018n}
Liu,~S.; Demirel,~M.~F.; Liang,~Y. N-gram graph: Simple unsupervised
  representation for graphs, with applications to molecules. \emph{arXiv
  preprint arXiv:1806.09206} \textbf{2018}, \relax
\mciteBstWouldAddEndPunctfalse
\mciteSetBstMidEndSepPunct{\mcitedefaultmidpunct}
{}{\mcitedefaultseppunct}\relax
\EndOfBibitem
\bibitem[Lu \latin{et~al.}(2019)Lu, Liu, Wang, Huang, Lin, and
  He]{lu2019molecular}
Lu,~C.; Liu,~Q.; Wang,~C.; Huang,~Z.; Lin,~P.; He,~L. Molecular property
  prediction: A multilevel quantum interactions modeling perspective.
  Proceedings of the AAAI Conference on Artificial Intelligence. 2019; pp
  1052--1060\relax
\mciteBstWouldAddEndPuncttrue
\mciteSetBstMidEndSepPunct{\mcitedefaultmidpunct}
{\mcitedefaultendpunct}{\mcitedefaultseppunct}\relax
\EndOfBibitem
\bibitem[Cai \latin{et~al.}(2019)Cai, Guo, Zhou, Zhou, Wang, Zhang, Fang, and
  Cheng]{cai2019deep}
Cai,~C.; Guo,~P.; Zhou,~Y.; Zhou,~J.; Wang,~Q.; Zhang,~F.; Fang,~J.; Cheng,~F.
  Deep learning-based prediction of drug-induced cardiotoxicity. \emph{J. Chem.
  Inf. Model} \textbf{2019}, \emph{59}, 1073--1084\relax
\mciteBstWouldAddEndPuncttrue
\mciteSetBstMidEndSepPunct{\mcitedefaultmidpunct}
{\mcitedefaultendpunct}{\mcitedefaultseppunct}\relax
\EndOfBibitem
\bibitem[Wang \latin{et~al.}(2019)Wang, Li, Jiang, Wang, Zhang, and
  Wei]{wang2019molecule}
Wang,~X.; Li,~Z.; Jiang,~M.; Wang,~S.; Zhang,~S.; Wei,~Z. Molecule property
  prediction based on spatial graph embedding. \emph{J. Chem. Inf. Model}
  \textbf{2019}, \emph{59}, 3817--3828\relax
\mciteBstWouldAddEndPuncttrue
\mciteSetBstMidEndSepPunct{\mcitedefaultmidpunct}
{\mcitedefaultendpunct}{\mcitedefaultseppunct}\relax
\EndOfBibitem
\bibitem[Hu \latin{et~al.}(2019)Hu, Liu, Gomes, Zitnik, Liang, Pande, and
  Leskovec]{hu2019strategies}
Hu,~W.; Liu,~B.; Gomes,~J.; Zitnik,~M.; Liang,~P.; Pande,~V.; Leskovec,~J.
  Strategies for pre-training graph neural networks. \emph{arXiv preprint
  arXiv:1905.12265} \textbf{2019}, \relax
\mciteBstWouldAddEndPunctfalse
\mciteSetBstMidEndSepPunct{\mcitedefaultmidpunct}
{}{\mcitedefaultseppunct}\relax
\EndOfBibitem
\bibitem[Hao \latin{et~al.}(2020)Hao, Lu, Huang, Wang, Hu, Liu, Chen, and
  Lee]{hao2020asgn}
Hao,~Z.; Lu,~C.; Huang,~Z.; Wang,~H.; Hu,~Z.; Liu,~Q.; Chen,~E.; Lee,~C. ASGN:
  An active semi-supervised graph neural network for molecular property
  prediction. Proceedings of the 26th ACM SIGKDD International Conference on
  Knowledge Discovery \& Data Mining. 2020; pp 731--752\relax
\mciteBstWouldAddEndPuncttrue
\mciteSetBstMidEndSepPunct{\mcitedefaultmidpunct}
{\mcitedefaultendpunct}{\mcitedefaultseppunct}\relax
\EndOfBibitem
\bibitem[Nguyen \latin{et~al.}(2020)Nguyen, Kreatsoulas, and
  Branson]{nguyen2020meta}
Nguyen,~C.~Q.; Kreatsoulas,~C.; Branson,~K.~M. Meta-Learning Initializations
  for Low-Resource Drug Discovery. \emph{arXiv preprint arXiv:2003.05996}
  \textbf{2020}, \relax
\mciteBstWouldAddEndPunctfalse
\mciteSetBstMidEndSepPunct{\mcitedefaultmidpunct}
{}{\mcitedefaultseppunct}\relax
\EndOfBibitem
\bibitem[Li \latin{et~al.}(2019)Li, Yan, Gu, Zhou, Wu, and
  Xu]{li2019deepchemstable}
Li,~X.; Yan,~X.; Gu,~Q.; Zhou,~H.; Wu,~D.; Xu,~J. DeepChemStable: Chemical
  stability prediction with an attention-based graph convolution network.
  \emph{J. Chem. Inf. Model} \textbf{2019}, \emph{59}, 1044--1049\relax
\mciteBstWouldAddEndPuncttrue
\mciteSetBstMidEndSepPunct{\mcitedefaultmidpunct}
{\mcitedefaultendpunct}{\mcitedefaultseppunct}\relax
\EndOfBibitem
\bibitem[Tang \latin{et~al.}(2020)Tang, Kramer, Fang, Qiu, Wu, and
  Xu]{tang2020self}
Tang,~B.; Kramer,~S.~T.; Fang,~M.; Qiu,~Y.; Wu,~Z.; Xu,~D. A self-attention
  based message passing neural network for predicting molecular lipophilicity
  and aqueous solubility. \emph{J. Cheminformatics} \textbf{2020}, \emph{12},
  1--9\relax
\mciteBstWouldAddEndPuncttrue
\mciteSetBstMidEndSepPunct{\mcitedefaultmidpunct}
{\mcitedefaultendpunct}{\mcitedefaultseppunct}\relax
\EndOfBibitem
\bibitem[Pathak \latin{et~al.}(2020)Pathak, Laghuvarapu, Mehta, and
  Priyakumar]{pathak2020chemically}
Pathak,~Y.; Laghuvarapu,~S.; Mehta,~S.; Priyakumar,~U.~D. Chemically
  interpretable graph interaction network for prediction of pharmacokinetic
  properties of drug-like molecules. Proceedings of the AAAI Conference on
  Artificial Intelligence. 2020; pp 873--880\relax
\mciteBstWouldAddEndPuncttrue
\mciteSetBstMidEndSepPunct{\mcitedefaultmidpunct}
{\mcitedefaultendpunct}{\mcitedefaultseppunct}\relax
\EndOfBibitem
\bibitem[Zhou \latin{et~al.}(2019)Zhou, Kearnes, Li, Zare, and
  Riley]{zhou2019optimization}
Zhou,~Z.; Kearnes,~S.; Li,~L.; Zare,~R.~N.; Riley,~P. Optimization of molecules
  via deep reinforcement learning. \emph{Scientific reports} \textbf{2019},
  \emph{9}, 1--10\relax
\mciteBstWouldAddEndPuncttrue
\mciteSetBstMidEndSepPunct{\mcitedefaultmidpunct}
{\mcitedefaultendpunct}{\mcitedefaultseppunct}\relax
\EndOfBibitem
\bibitem[Khemchandani \latin{et~al.}(2020)Khemchandani, O’Hagan, Samanta,
  Swainston, Roberts, Bollegala, and Kell]{khemchandani2020deepgraphmolgen}
Khemchandani,~Y.; O’Hagan,~S.; Samanta,~S.; Swainston,~N.; Roberts,~T.~J.;
  Bollegala,~D.; Kell,~D.~B. DeepGraphMolGen, a multi-objective, computational
  strategy for generating molecules with desirable properties: a graph
  convolution and reinforcement learning approach. \emph{J. Cheminformatics}
  \textbf{2020}, \emph{12}, 1--17\relax
\mciteBstWouldAddEndPuncttrue
\mciteSetBstMidEndSepPunct{\mcitedefaultmidpunct}
{\mcitedefaultendpunct}{\mcitedefaultseppunct}\relax
\EndOfBibitem
\bibitem[Xu \latin{et~al.}(2020)Xu, Liu, Huang, and Jiang]{xu2020reinforced}
Xu,~C.; Liu,~Q.; Huang,~M.; Jiang,~T. Reinforced Molecular Optimization with
  Neighborhood-Controlled Grammars. \emph{arXiv preprint arXiv:2011.07225}
  \textbf{2020}, \relax
\mciteBstWouldAddEndPunctfalse
\mciteSetBstMidEndSepPunct{\mcitedefaultmidpunct}
{}{\mcitedefaultseppunct}\relax
\EndOfBibitem
\bibitem[Kingma and Welling(2013)Kingma, and Welling]{kingma2013auto}
Kingma,~D.~P.; Welling,~M. Auto-encoding variational bayes. \emph{arXiv
  preprint arXiv:1312.6114} \textbf{2013}, \relax
\mciteBstWouldAddEndPunctfalse
\mciteSetBstMidEndSepPunct{\mcitedefaultmidpunct}
{}{\mcitedefaultseppunct}\relax
\EndOfBibitem
\bibitem[Kingma and Welling(2019)Kingma, and Welling]{kingma2019introduction}
Kingma,~D.~P.; Welling,~M. An introduction to variational autoencoders.
  \emph{arXiv preprint arXiv:1906.02691} \textbf{2019}, \relax
\mciteBstWouldAddEndPunctfalse
\mciteSetBstMidEndSepPunct{\mcitedefaultmidpunct}
{}{\mcitedefaultseppunct}\relax
\EndOfBibitem
\bibitem[Kusner \latin{et~al.}(2017)Kusner, Paige, and
  Hern{\'a}ndez-Lobato]{kusner2017grammar}
Kusner,~M.~J.; Paige,~B.; Hern{\'a}ndez-Lobato,~J.~M. Grammar variational
  autoencoder. International Conference on Machine Learning. 2017; pp
  1945--1954\relax
\mciteBstWouldAddEndPuncttrue
\mciteSetBstMidEndSepPunct{\mcitedefaultmidpunct}
{\mcitedefaultendpunct}{\mcitedefaultseppunct}\relax
\EndOfBibitem
\bibitem[Dai \latin{et~al.}(2018)Dai, Tian, Dai, Skiena, and
  Song]{dai2018syntax}
Dai,~H.; Tian,~Y.; Dai,~B.; Skiena,~S.; Song,~L. Syntax-directed variational
  autoencoder for structured data. \emph{arXiv preprint arXiv:1802.08786}
  \textbf{2018}, \relax
\mciteBstWouldAddEndPunctfalse
\mciteSetBstMidEndSepPunct{\mcitedefaultmidpunct}
{}{\mcitedefaultseppunct}\relax
\EndOfBibitem
\bibitem[Kang and Cho(2018)Kang, and Cho]{kang2018conditional}
Kang,~S.; Cho,~K. Conditional molecular design with deep generative models.
  \emph{J. Chem. Inf. Model} \textbf{2018}, \emph{59}, 43--52\relax
\mciteBstWouldAddEndPuncttrue
\mciteSetBstMidEndSepPunct{\mcitedefaultmidpunct}
{\mcitedefaultendpunct}{\mcitedefaultseppunct}\relax
\EndOfBibitem
\bibitem[Lim \latin{et~al.}(2018)Lim, Ryu, Kim, and Kim]{lim2018molecular}
Lim,~J.; Ryu,~S.; Kim,~J.~W.; Kim,~W.~Y. Molecular generative model based on
  conditional variational autoencoder for de novo molecular design. \emph{J.
  Cheminformatics} \textbf{2018}, \emph{10}, 1--9\relax
\mciteBstWouldAddEndPuncttrue
\mciteSetBstMidEndSepPunct{\mcitedefaultmidpunct}
{\mcitedefaultendpunct}{\mcitedefaultseppunct}\relax
\EndOfBibitem
\bibitem[Liu \latin{et~al.}(2018)Liu, Allamanis, Brockschmidt, and
  Gaunt]{liu2018constrained}
Liu,~Q.; Allamanis,~M.; Brockschmidt,~M.; Gaunt,~A.~L. Constrained graph
  variational autoencoders for molecule design. \emph{arXiv preprint
  arXiv:1805.09076} \textbf{2018}, \relax
\mciteBstWouldAddEndPunctfalse
\mciteSetBstMidEndSepPunct{\mcitedefaultmidpunct}
{}{\mcitedefaultseppunct}\relax
\EndOfBibitem
\bibitem[Samanta \latin{et~al.}(2020)Samanta, De, Jana, G{\'o}mez, Chattaraj,
  Ganguly, and Gomez-Rodriguez]{samanta2020nevae}
Samanta,~B.; De,~A.; Jana,~G.; G{\'o}mez,~V.; Chattaraj,~P.~K.; Ganguly,~N.;
  Gomez-Rodriguez,~M. Nevae: A deep generative model for molecular graphs.
  \emph{J Mach Learn Res} \textbf{2020}, \relax
\mciteBstWouldAddEndPunctfalse
\mciteSetBstMidEndSepPunct{\mcitedefaultmidpunct}
{}{\mcitedefaultseppunct}\relax
\EndOfBibitem
\bibitem[Chenthamarakshan \latin{et~al.}(2020)Chenthamarakshan, Das, Padhi,
  Strobelt, Lim, Hoover, Hoffman, and Mojsilovic]{chenthamarakshan2020target}
Chenthamarakshan,~V.; Das,~P.; Padhi,~I.; Strobelt,~H.; Lim,~K.~W.; Hoover,~B.;
  Hoffman,~S.~C.; Mojsilovic,~A. Target-specific and selective drug design for
  covid-19 using deep generative models. \emph{arXiv preprint arXiv:2004.01215}
  \textbf{2020}, \relax
\mciteBstWouldAddEndPunctfalse
\mciteSetBstMidEndSepPunct{\mcitedefaultmidpunct}
{}{\mcitedefaultseppunct}\relax
\EndOfBibitem
\bibitem[Makhzani \latin{et~al.}(2015)Makhzani, Shlens, Jaitly, Goodfellow, and
  Frey]{makhzani2015adversarial}
Makhzani,~A.; Shlens,~J.; Jaitly,~N.; Goodfellow,~I.; Frey,~B. Adversarial
  autoencoders. \emph{arXiv preprint arXiv:1511.05644} \textbf{2015}, \relax
\mciteBstWouldAddEndPunctfalse
\mciteSetBstMidEndSepPunct{\mcitedefaultmidpunct}
{}{\mcitedefaultseppunct}\relax
\EndOfBibitem
\bibitem[Kadurin \latin{et~al.}(2017)Kadurin, Nikolenko, Khrabrov, Aliper, and
  Zhavoronkov]{kadurin2017drugan}
Kadurin,~A.; Nikolenko,~S.; Khrabrov,~K.; Aliper,~A.; Zhavoronkov,~A. druGAN:
  an advanced generative adversarial autoencoder model for de novo generation
  of new molecules with desired molecular properties in silico. \emph{Mol.
  Pharm.} \textbf{2017}, \emph{14}, 3098--3104\relax
\mciteBstWouldAddEndPuncttrue
\mciteSetBstMidEndSepPunct{\mcitedefaultmidpunct}
{\mcitedefaultendpunct}{\mcitedefaultseppunct}\relax
\EndOfBibitem
\bibitem[Blaschke \latin{et~al.}(2018)Blaschke, Olivecrona, Engkvist, Bajorath,
  and Chen]{blaschke2018application}
Blaschke,~T.; Olivecrona,~M.; Engkvist,~O.; Bajorath,~J.; Chen,~H. Application
  of generative autoencoder in de novo molecular design. \emph{Mol. Inform.}
  \textbf{2018}, \emph{37}, 1700123\relax
\mciteBstWouldAddEndPuncttrue
\mciteSetBstMidEndSepPunct{\mcitedefaultmidpunct}
{\mcitedefaultendpunct}{\mcitedefaultseppunct}\relax
\EndOfBibitem
\bibitem[Polykovskiy \latin{et~al.}(2018)Polykovskiy, Zhebrak, Vetrov,
  Ivanenkov, Aladinskiy, Mamoshina, Bozdaganyan, Aliper, Zhavoronkov, and
  Kadurin]{polykovskiy2018entangled}
Polykovskiy,~D.; Zhebrak,~A.; Vetrov,~D.; Ivanenkov,~Y.; Aladinskiy,~V.;
  Mamoshina,~P.; Bozdaganyan,~M.; Aliper,~A.; Zhavoronkov,~A.; Kadurin,~A.
  Entangled conditional adversarial autoencoder for de novo drug discovery.
  \emph{Mol. Pharm.} \textbf{2018}, \emph{15}, 4398--4405\relax
\mciteBstWouldAddEndPuncttrue
\mciteSetBstMidEndSepPunct{\mcitedefaultmidpunct}
{\mcitedefaultendpunct}{\mcitedefaultseppunct}\relax
\EndOfBibitem
\bibitem[Simonovsky and Komodakis(2018)Simonovsky, and
  Komodakis]{simonovsky2018graphvae}
Simonovsky,~M.; Komodakis,~N. Graphvae: Towards generation of small graphs
  using variational autoencoders. International Conference on Artificial Neural
  Networks. 2018; pp 412--422\relax
\mciteBstWouldAddEndPuncttrue
\mciteSetBstMidEndSepPunct{\mcitedefaultmidpunct}
{\mcitedefaultendpunct}{\mcitedefaultseppunct}\relax
\EndOfBibitem
\bibitem[Jin \latin{et~al.}(2018)Jin, Barzilay, and Jaakkola]{jin2018junction}
Jin,~W.; Barzilay,~R.; Jaakkola,~T. Junction tree variational autoencoder for
  molecular graph generation. International Conference on Machine Learning.
  2018; pp 2323--2332\relax
\mciteBstWouldAddEndPuncttrue
\mciteSetBstMidEndSepPunct{\mcitedefaultmidpunct}
{\mcitedefaultendpunct}{\mcitedefaultseppunct}\relax
\EndOfBibitem
\bibitem[Ma \latin{et~al.}(2018)Ma, Chen, and Xiao]{ma2018constrained}
Ma,~T.; Chen,~J.; Xiao,~C. Constrained generation of semantically valid graphs
  via regularizing variational autoencoders. \emph{arXiv preprint
  arXiv:1809.02630} \textbf{2018}, \relax
\mciteBstWouldAddEndPunctfalse
\mciteSetBstMidEndSepPunct{\mcitedefaultmidpunct}
{}{\mcitedefaultseppunct}\relax
\EndOfBibitem
\bibitem[Kajino(2019)]{kajino2019molecular}
Kajino,~H. Molecular hypergraph grammar with its application to molecular
  optimization. International Conference on Machine Learning. 2019; pp
  3183--3191\relax
\mciteBstWouldAddEndPuncttrue
\mciteSetBstMidEndSepPunct{\mcitedefaultmidpunct}
{\mcitedefaultendpunct}{\mcitedefaultseppunct}\relax
\EndOfBibitem
\bibitem[Kwon \latin{et~al.}(2019)Kwon, Yoo, Choi, Son, Lee, and
  Kang]{kwon2019efficient}
Kwon,~Y.; Yoo,~J.; Choi,~Y.-S.; Son,~W.-J.; Lee,~D.; Kang,~S. Efficient
  learning of non-autoregressive graph variational autoencoders for molecular
  graph generation. \emph{J. Cheminformatics} \textbf{2019}, \emph{11},
  1--10\relax
\mciteBstWouldAddEndPuncttrue
\mciteSetBstMidEndSepPunct{\mcitedefaultmidpunct}
{\mcitedefaultendpunct}{\mcitedefaultseppunct}\relax
\EndOfBibitem
\bibitem[Lim \latin{et~al.}(2019)Lim, Hwang, Kim, Moon, and
  Kim]{lim2019scaffold}
Lim,~J.; Hwang,~S.-Y.; Kim,~S.; Moon,~S.; Kim,~W.~Y. Scaffold-based molecular
  design using graph generative model. \emph{arXiv preprint arXiv:1905.13639}
  \textbf{2019}, \relax
\mciteBstWouldAddEndPunctfalse
\mciteSetBstMidEndSepPunct{\mcitedefaultmidpunct}
{}{\mcitedefaultseppunct}\relax
\EndOfBibitem
\bibitem[Fu \latin{et~al.}(2020)Fu, Xiao, and Sun]{fu2020core}
Fu,~T.; Xiao,~C.; Sun,~J. Core: Automatic molecule optimization using copy \&
  refine strategy. Proceedings of the AAAI Conference on Artificial
  Intelligence. 2020; pp 638--645\relax
\mciteBstWouldAddEndPuncttrue
\mciteSetBstMidEndSepPunct{\mcitedefaultmidpunct}
{\mcitedefaultendpunct}{\mcitedefaultseppunct}\relax
\EndOfBibitem
\bibitem[Kwon \latin{et~al.}(2020)Kwon, Lee, Choi, Shin, and
  Kang]{kwon2020compressed}
Kwon,~Y.; Lee,~D.; Choi,~Y.-S.; Shin,~K.; Kang,~S. Compressed graph
  representation for scalable molecular graph generation. \emph{J.
  Cheminformatics} \textbf{2020}, \emph{12}, 1--8\relax
\mciteBstWouldAddEndPuncttrue
\mciteSetBstMidEndSepPunct{\mcitedefaultmidpunct}
{\mcitedefaultendpunct}{\mcitedefaultseppunct}\relax
\EndOfBibitem
\bibitem[Jin \latin{et~al.}(2020)Jin, Barzilay, and
  Jaakkola]{jin2020hierarchical}
Jin,~W.; Barzilay,~R.; Jaakkola,~T. Hierarchical generation of molecular graphs
  using structural motifs. International Conference on Machine Learning. 2020;
  pp 4839--4848\relax
\mciteBstWouldAddEndPuncttrue
\mciteSetBstMidEndSepPunct{\mcitedefaultmidpunct}
{\mcitedefaultendpunct}{\mcitedefaultseppunct}\relax
\EndOfBibitem
\bibitem[Goodfellow \latin{et~al.}(2014)Goodfellow, Pouget-Abadie, Mirza, Xu,
  Warde-Farley, Ozair, Courville, and Bengio]{goodfellow2014generative}
Goodfellow,~I.~J.; Pouget-Abadie,~J.; Mirza,~M.; Xu,~B.; Warde-Farley,~D.;
  Ozair,~S.; Courville,~A.; Bengio,~Y. Generative adversarial networks.
  \emph{arXiv preprint arXiv:1406.2661} \textbf{2014}, \relax
\mciteBstWouldAddEndPunctfalse
\mciteSetBstMidEndSepPunct{\mcitedefaultmidpunct}
{}{\mcitedefaultseppunct}\relax
\EndOfBibitem
\bibitem[Yu \latin{et~al.}(2017)Yu, Zhang, Wang, and Yu]{yu2017seqgan}
Yu,~L.; Zhang,~W.; Wang,~J.; Yu,~Y. Seqgan: Sequence generative adversarial
  nets with policy gradient. Proceedings of the AAAI Conference on Artificial
  Intelligence. 2017\relax
\mciteBstWouldAddEndPuncttrue
\mciteSetBstMidEndSepPunct{\mcitedefaultmidpunct}
{\mcitedefaultendpunct}{\mcitedefaultseppunct}\relax
\EndOfBibitem
\bibitem[Putin \latin{et~al.}(2018)Putin, Asadulaev, Ivanenkov, Aladinskiy,
  Sanchez-Lengeling, Aspuru-Guzik, and Zhavoronkov]{putin2018reinforced}
Putin,~E.; Asadulaev,~A.; Ivanenkov,~Y.; Aladinskiy,~V.; Sanchez-Lengeling,~B.;
  Aspuru-Guzik,~A.; Zhavoronkov,~A. Reinforced adversarial neural computer for
  de novo molecular design. \emph{J. Chem. Inf. Model} \textbf{2018},
  \emph{58}, 1194--1204\relax
\mciteBstWouldAddEndPuncttrue
\mciteSetBstMidEndSepPunct{\mcitedefaultmidpunct}
{\mcitedefaultendpunct}{\mcitedefaultseppunct}\relax
\EndOfBibitem
\bibitem[De~Cao and Kipf(2018)De~Cao, and Kipf]{de2018molgan}
De~Cao,~N.; Kipf,~T. MolGAN: An implicit generative model for small molecular
  graphs. \emph{arXiv preprint arXiv:1805.11973} \textbf{2018}, \relax
\mciteBstWouldAddEndPunctfalse
\mciteSetBstMidEndSepPunct{\mcitedefaultmidpunct}
{}{\mcitedefaultseppunct}\relax
\EndOfBibitem
\bibitem[Salimans \latin{et~al.}(2016)Salimans, Goodfellow, Zaremba, Cheung,
  Radford, and Chen]{salimans2016improved}
Salimans,~T.; Goodfellow,~I.; Zaremba,~W.; Cheung,~V.; Radford,~A.; Chen,~X.
  Improved techniques for training gans. \emph{arXiv preprint arXiv:1606.03498}
  \textbf{2016}, \relax
\mciteBstWouldAddEndPunctfalse
\mciteSetBstMidEndSepPunct{\mcitedefaultmidpunct}
{}{\mcitedefaultseppunct}\relax
\EndOfBibitem
\bibitem[Kobyzev \latin{et~al.}(2020)Kobyzev, Prince, and
  Brubaker]{kobyzev2020normalizing}
Kobyzev,~I.; Prince,~S.; Brubaker,~M. Normalizing flows: An introduction and
  review of current methods. \emph{IEEE PAMI} \textbf{2020}, \relax
\mciteBstWouldAddEndPunctfalse
\mciteSetBstMidEndSepPunct{\mcitedefaultmidpunct}
{}{\mcitedefaultseppunct}\relax
\EndOfBibitem
\bibitem[Dinh \latin{et~al.}(2014)Dinh, Krueger, and Bengio]{dinh2014nice}
Dinh,~L.; Krueger,~D.; Bengio,~Y. Nice: Non-linear independent components
  estimation. \emph{arXiv preprint arXiv:1410.8516} \textbf{2014}, \relax
\mciteBstWouldAddEndPunctfalse
\mciteSetBstMidEndSepPunct{\mcitedefaultmidpunct}
{}{\mcitedefaultseppunct}\relax
\EndOfBibitem
\bibitem[Dinh \latin{et~al.}(2016)Dinh, Sohl-Dickstein, and
  Bengio]{dinh2016density}
Dinh,~L.; Sohl-Dickstein,~J.; Bengio,~S. Density estimation using real nvp.
  \emph{arXiv preprint arXiv:1605.08803} \textbf{2016}, \relax
\mciteBstWouldAddEndPunctfalse
\mciteSetBstMidEndSepPunct{\mcitedefaultmidpunct}
{}{\mcitedefaultseppunct}\relax
\EndOfBibitem
\bibitem[Madhawa \latin{et~al.}(2019)Madhawa, Ishiguro, Nakago, and
  Abe]{madhawa2019graphnvp}
Madhawa,~K.; Ishiguro,~K.; Nakago,~K.; Abe,~M. Graphnvp: An invertible flow
  model for generating molecular graphs. \emph{arXiv preprint arXiv:1905.11600}
  \textbf{2019}, \relax
\mciteBstWouldAddEndPunctfalse
\mciteSetBstMidEndSepPunct{\mcitedefaultmidpunct}
{}{\mcitedefaultseppunct}\relax
\EndOfBibitem
\bibitem[Honda \latin{et~al.}(2019)Honda, Akita, Ishiguro, Nakanishi, and
  Oono]{honda2019graph}
Honda,~S.; Akita,~H.; Ishiguro,~K.; Nakanishi,~T.; Oono,~K. Graph residual flow
  for molecular graph generation. \emph{arXiv preprint arXiv:1909.13521}
  \textbf{2019}, \relax
\mciteBstWouldAddEndPunctfalse
\mciteSetBstMidEndSepPunct{\mcitedefaultmidpunct}
{}{\mcitedefaultseppunct}\relax
\EndOfBibitem
\bibitem[Shi \latin{et~al.}(2020)Shi, Xu, Zhu, Zhang, Zhang, and
  Tang]{shi2020graphaf}
Shi,~C.; Xu,~M.; Zhu,~Z.; Zhang,~W.; Zhang,~M.; Tang,~J. Graphaf: a flow-based
  autoregressive model for molecular graph generation. \emph{arXiv preprint
  arXiv:2001.09382} \textbf{2020}, \relax
\mciteBstWouldAddEndPunctfalse
\mciteSetBstMidEndSepPunct{\mcitedefaultmidpunct}
{}{\mcitedefaultseppunct}\relax
\EndOfBibitem
\bibitem[Zang and Wang(2020)Zang, and Wang]{zang2020moflow}
Zang,~C.; Wang,~F. MoFlow: an invertible flow model for generating molecular
  graphs. Proceedings of the 26th ACM SIGKDD International Conference on
  Knowledge Discovery \& Data Mining. 2020; pp 617--626\relax
\mciteBstWouldAddEndPuncttrue
\mciteSetBstMidEndSepPunct{\mcitedefaultmidpunct}
{\mcitedefaultendpunct}{\mcitedefaultseppunct}\relax
\EndOfBibitem
\bibitem[Luo \latin{et~al.}(2021)Luo, Yan, and Ji]{luo2021graphdf}
Luo,~Y.; Yan,~K.; Ji,~S. GraphDF: A Discrete Flow Model for Molecular Graph
  Generation. \emph{arXiv preprint arXiv:2102.01189} \textbf{2021}, \relax
\mciteBstWouldAddEndPunctfalse
\mciteSetBstMidEndSepPunct{\mcitedefaultmidpunct}
{}{\mcitedefaultseppunct}\relax
\EndOfBibitem
\bibitem[Vaswani \latin{et~al.}(2017)Vaswani, Shazeer, Parmar, Uszkoreit,
  Jones, Gomez, Kaiser, and Polosukhin]{vaswani2017attention}
Vaswani,~A.; Shazeer,~N.; Parmar,~N.; Uszkoreit,~J.; Jones,~L.; Gomez,~A.~N.;
  Kaiser,~L.; Polosukhin,~I. Attention is all you need. \emph{arXiv preprint
  arXiv:1706.03762} \textbf{2017}, \relax
\mciteBstWouldAddEndPunctfalse
\mciteSetBstMidEndSepPunct{\mcitedefaultmidpunct}
{}{\mcitedefaultseppunct}\relax
\EndOfBibitem
\bibitem[Radford \latin{et~al.}(2018)Radford, Narasimhan, Salimans, and
  Sutskever]{radford2018improving}
Radford,~A.; Narasimhan,~K.; Salimans,~T.; Sutskever,~I. Improving language
  understanding by generative pre-training. \emph{OpenAI} \textbf{2018}, \relax
\mciteBstWouldAddEndPunctfalse
\mciteSetBstMidEndSepPunct{\mcitedefaultmidpunct}
{}{\mcitedefaultseppunct}\relax
\EndOfBibitem
\bibitem[Devlin \latin{et~al.}(2018)Devlin, Chang, Lee, and
  Toutanova]{devlin2018bert}
Devlin,~J.; Chang,~M.-W.; Lee,~K.; Toutanova,~K. Bert: Pre-training of deep
  bidirectional transformers for language understanding. \emph{arXiv preprint
  arXiv:1810.04805} \textbf{2018}, \relax
\mciteBstWouldAddEndPunctfalse
\mciteSetBstMidEndSepPunct{\mcitedefaultmidpunct}
{}{\mcitedefaultseppunct}\relax
\EndOfBibitem
\bibitem[Radford \latin{et~al.}(2019)Radford, Wu, Child, Luan, Amodei, and
  Sutskever]{radford2019language}
Radford,~A.; Wu,~J.; Child,~R.; Luan,~D.; Amodei,~D.; Sutskever,~I. Language
  models are unsupervised multitask learners. \emph{OpenAI blog} \textbf{2019},
  \emph{1}, 9\relax
\mciteBstWouldAddEndPuncttrue
\mciteSetBstMidEndSepPunct{\mcitedefaultmidpunct}
{\mcitedefaultendpunct}{\mcitedefaultseppunct}\relax
\EndOfBibitem
\bibitem[Liu \latin{et~al.}(2019)Liu, Ott, Goyal, Du, Joshi, Chen, Levy, Lewis,
  Zettlemoyer, and Stoyanov]{liu2019roberta}
Liu,~Y.; Ott,~M.; Goyal,~N.; Du,~J.; Joshi,~M.; Chen,~D.; Levy,~O.; Lewis,~M.;
  Zettlemoyer,~L.; Stoyanov,~V. Roberta: A robustly optimized bert pretraining
  approach. \emph{arXiv preprint arXiv:1907.11692} \textbf{2019}, \relax
\mciteBstWouldAddEndPunctfalse
\mciteSetBstMidEndSepPunct{\mcitedefaultmidpunct}
{}{\mcitedefaultseppunct}\relax
\EndOfBibitem
\bibitem[Brown \latin{et~al.}(2020)Brown, Mann, Ryder, Subbiah, Kaplan,
  Dhariwal, Neelakantan, Shyam, Sastry, Askell, \latin{et~al.}
  others]{brown2020language}
Brown,~T.~B.; Mann,~B.; Ryder,~N.; Subbiah,~M.; Kaplan,~J.; Dhariwal,~P.;
  Neelakantan,~A.; Shyam,~P.; Sastry,~G.; Askell,~A., \latin{et~al.}  Language
  models are few-shot learners. \emph{arXiv preprint arXiv:2005.14165}
  \textbf{2020}, \relax
\mciteBstWouldAddEndPunctfalse
\mciteSetBstMidEndSepPunct{\mcitedefaultmidpunct}
{}{\mcitedefaultseppunct}\relax
\EndOfBibitem
\bibitem[Carion \latin{et~al.}(2020)Carion, Massa, Synnaeve, Usunier, Kirillov,
  and Zagoruyko]{carion2020end}
Carion,~N.; Massa,~F.; Synnaeve,~G.; Usunier,~N.; Kirillov,~A.; Zagoruyko,~S.
  End-to-end object detection with transformers. European Conference on
  Computer Vision. 2020; pp 213--229\relax
\mciteBstWouldAddEndPuncttrue
\mciteSetBstMidEndSepPunct{\mcitedefaultmidpunct}
{\mcitedefaultendpunct}{\mcitedefaultseppunct}\relax
\EndOfBibitem
\bibitem[Dosovitskiy \latin{et~al.}(2020)Dosovitskiy, Beyer, Kolesnikov,
  Weissenborn, Zhai, Unterthiner, Dehghani, Minderer, Heigold, Gelly,
  \latin{et~al.} others]{dosovitskiy2020image}
Dosovitskiy,~A.; Beyer,~L.; Kolesnikov,~A.; Weissenborn,~D.; Zhai,~X.;
  Unterthiner,~T.; Dehghani,~M.; Minderer,~M.; Heigold,~G.; Gelly,~S.,
  \latin{et~al.}  An image is worth 16x16 words: Transformers for image
  recognition at scale. \emph{arXiv preprint arXiv:2010.11929} \textbf{2020},
  \relax
\mciteBstWouldAddEndPunctfalse
\mciteSetBstMidEndSepPunct{\mcitedefaultmidpunct}
{}{\mcitedefaultseppunct}\relax
\EndOfBibitem
\bibitem[Wang \latin{et~al.}(2019)Wang, Guo, Wang, Sun, and
  Huang]{wang2019smiles}
Wang,~S.; Guo,~Y.; Wang,~Y.; Sun,~H.; Huang,~J. SMILES-BERT: large scale
  unsupervised pre-training for molecular property prediction. Proceedings of
  the 10th ACM International Conference on Bioinformatics, Computational
  Biology and Health Informatics. 2019; pp 429--436\relax
\mciteBstWouldAddEndPuncttrue
\mciteSetBstMidEndSepPunct{\mcitedefaultmidpunct}
{\mcitedefaultendpunct}{\mcitedefaultseppunct}\relax
\EndOfBibitem
\bibitem[Honda \latin{et~al.}(2019)Honda, Shi, and Ueda]{honda2019smiles}
Honda,~S.; Shi,~S.; Ueda,~H.~R. SMILES transformer: pre-trained molecular
  fingerprint for low data drug discovery. \emph{arXiv preprint
  arXiv:1911.04738} \textbf{2019}, \relax
\mciteBstWouldAddEndPunctfalse
\mciteSetBstMidEndSepPunct{\mcitedefaultmidpunct}
{}{\mcitedefaultseppunct}\relax
\EndOfBibitem
\bibitem[Bradshaw \latin{et~al.}(2019)Bradshaw, Paige, Kusner, Segler, and
  Hern{\'a}ndez-Lobato]{bradshaw2019model}
Bradshaw,~J.; Paige,~B.; Kusner,~M.~J.; Segler,~M.~H.;
  Hern{\'a}ndez-Lobato,~J.~M. A model to search for synthesizable molecules.
  \emph{arXiv preprint arXiv:1906.05221} \textbf{2019}, \relax
\mciteBstWouldAddEndPunctfalse
\mciteSetBstMidEndSepPunct{\mcitedefaultmidpunct}
{}{\mcitedefaultseppunct}\relax
\EndOfBibitem
\bibitem[Grechishnikova(2021)]{grechishnikova2021transformer}
Grechishnikova,~D. Transformer neural network for protein-specific de novo drug
  generation as a machine translation problem. \emph{Scientific reports}
  \textbf{2021}, \emph{11}, 1--13\relax
\mciteBstWouldAddEndPuncttrue
\mciteSetBstMidEndSepPunct{\mcitedefaultmidpunct}
{\mcitedefaultendpunct}{\mcitedefaultseppunct}\relax
\EndOfBibitem
\bibitem[Liu \latin{et~al.}(2020)Liu, Zhang, Hou, Wang, Mian, Zhang, and
  Tang]{liu2020self}
Liu,~X.; Zhang,~F.; Hou,~Z.; Wang,~Z.; Mian,~L.; Zhang,~J.; Tang,~J.
  Self-supervised learning: Generative or contrastive. \emph{arXiv preprint
  arXiv:2006.08218} \textbf{2020}, \emph{1}\relax
\mciteBstWouldAddEndPuncttrue
\mciteSetBstMidEndSepPunct{\mcitedefaultmidpunct}
{\mcitedefaultendpunct}{\mcitedefaultseppunct}\relax
\EndOfBibitem
\bibitem[Wang \latin{et~al.}(2021)Wang, Wang, Cao, and
  Farimani]{wang2021molclr}
Wang,~Y.; Wang,~J.; Cao,~Z.; Farimani,~A.~B. MolCLR: Molecular contrastive
  learning of representations via graph neural networks. \emph{arXiv preprint
  arXiv:2102.10056} \textbf{2021}, \relax
\mciteBstWouldAddEndPunctfalse
\mciteSetBstMidEndSepPunct{\mcitedefaultmidpunct}
{}{\mcitedefaultseppunct}\relax
\EndOfBibitem
\bibitem[Vanschoren(2018)]{vanschoren2018meta}
Vanschoren,~J. Meta-learning: A survey. \emph{arXiv preprint arXiv:1810.03548}
  \textbf{2018}, \relax
\mciteBstWouldAddEndPunctfalse
\mciteSetBstMidEndSepPunct{\mcitedefaultmidpunct}
{}{\mcitedefaultseppunct}\relax
\EndOfBibitem
\bibitem[Wang \latin{et~al.}(2020)Wang, Yao, Kwok, and
  Ni]{wang2020generalizing}
Wang,~Y.; Yao,~Q.; Kwok,~J.~T.; Ni,~L.~M. Generalizing from a few examples: A
  survey on few-shot learning. \emph{ACM Computing Surveys (CSUR)}
  \textbf{2020}, \emph{53}, 1--34\relax
\mciteBstWouldAddEndPuncttrue
\mciteSetBstMidEndSepPunct{\mcitedefaultmidpunct}
{\mcitedefaultendpunct}{\mcitedefaultseppunct}\relax
\EndOfBibitem
\bibitem[Kulis \latin{et~al.}(2012)Kulis, \latin{et~al.}
  others]{kulis2012metric}
Kulis,~B., \latin{et~al.}  Metric learning: A survey. \emph{Found. Trends Mach.
  Learn.} \textbf{2012}, \emph{5}, 287--364\relax
\mciteBstWouldAddEndPuncttrue
\mciteSetBstMidEndSepPunct{\mcitedefaultmidpunct}
{\mcitedefaultendpunct}{\mcitedefaultseppunct}\relax
\EndOfBibitem
\bibitem[Yang \latin{et~al.}(2021)Yang, Bastan, Zhu, Gray, and
  Samaras]{yang2021hierarchical}
Yang,~Z.; Bastan,~M.; Zhu,~X.; Gray,~D.; Samaras,~D. Hierarchical Proxy-based
  Loss for Deep Metric Learning. \emph{arXiv preprint arXiv:2103.13538}
  \textbf{2021}, \relax
\mciteBstWouldAddEndPunctfalse
\mciteSetBstMidEndSepPunct{\mcitedefaultmidpunct}
{}{\mcitedefaultseppunct}\relax
\EndOfBibitem
\bibitem[Movshovitz-Attias \latin{et~al.}(2017)Movshovitz-Attias, Toshev,
  Leung, Ioffe, and Singh]{movshovitz2017no}
Movshovitz-Attias,~Y.; Toshev,~A.; Leung,~T.~K.; Ioffe,~S.; Singh,~S. No fuss
  distance metric learning using proxies. Proceedings of the IEEE International
  Conference on Computer Vision. 2017; pp 360--368\relax
\mciteBstWouldAddEndPuncttrue
\mciteSetBstMidEndSepPunct{\mcitedefaultmidpunct}
{\mcitedefaultendpunct}{\mcitedefaultseppunct}\relax
\EndOfBibitem
\bibitem[Na \latin{et~al.}(2020)Na, Chang, and Kim]{na2020machine}
Na,~G.~S.; Chang,~H.; Kim,~H.~W. Machine-guided representation for accurate
  graph-based molecular machine learning. \emph{Phys. Chem. Chem. Phys.}
  \textbf{2020}, \emph{22}, 18526--18535\relax
\mciteBstWouldAddEndPuncttrue
\mciteSetBstMidEndSepPunct{\mcitedefaultmidpunct}
{\mcitedefaultendpunct}{\mcitedefaultseppunct}\relax
\EndOfBibitem
\bibitem[Koge \latin{et~al.}(2020)Koge, Ono, Huang, Altaf-Ul-Amin, and
  Kanaya]{koge2020embedding}
Koge,~D.; Ono,~N.; Huang,~M.; Altaf-Ul-Amin,~M.; Kanaya,~S. Embedding of
  Molecular Structure Using Molecular Hypergraph Variational Autoencoder with
  Metric Learning. \emph{Mol. Inform.} \textbf{2020}, \relax
\mciteBstWouldAddEndPunctfalse
\mciteSetBstMidEndSepPunct{\mcitedefaultmidpunct}
{}{\mcitedefaultseppunct}\relax
\EndOfBibitem
\bibitem[Sutton and Barto(2018)Sutton, and Barto]{sutton2018reinforcement}
Sutton,~R.~S.; Barto,~A.~G. \emph{Reinforcement learning: An introduction}; MIT
  press, 2018\relax
\mciteBstWouldAddEndPuncttrue
\mciteSetBstMidEndSepPunct{\mcitedefaultmidpunct}
{\mcitedefaultendpunct}{\mcitedefaultseppunct}\relax
\EndOfBibitem
\bibitem[Arulkumaran \latin{et~al.}(2017)Arulkumaran, Deisenroth, Brundage, and
  Bharath]{arulkumaran2017brief}
Arulkumaran,~K.; Deisenroth,~M.~P.; Brundage,~M.; Bharath,~A.~A. A brief survey
  of deep reinforcement learning. \emph{arXiv preprint arXiv:1708.05866}
  \textbf{2017}, \relax
\mciteBstWouldAddEndPunctfalse
\mciteSetBstMidEndSepPunct{\mcitedefaultmidpunct}
{}{\mcitedefaultseppunct}\relax
\EndOfBibitem
\bibitem[Van~Hasselt \latin{et~al.}(2016)Van~Hasselt, Guez, and
  Silver]{van2016deep}
Van~Hasselt,~H.; Guez,~A.; Silver,~D. Deep reinforcement learning with double
  q-learning. Proceedings of the AAAI Conference on Artificial Intelligence.
  2016\relax
\mciteBstWouldAddEndPuncttrue
\mciteSetBstMidEndSepPunct{\mcitedefaultmidpunct}
{\mcitedefaultendpunct}{\mcitedefaultseppunct}\relax
\EndOfBibitem
\bibitem[Williams(1992)]{williams1992simple}
Williams,~R.~J. Simple statistical gradient-following algorithms for
  connectionist reinforcement learning. \emph{Machine learning} \textbf{1992},
  \emph{8}, 229--256\relax
\mciteBstWouldAddEndPuncttrue
\mciteSetBstMidEndSepPunct{\mcitedefaultmidpunct}
{\mcitedefaultendpunct}{\mcitedefaultseppunct}\relax
\EndOfBibitem
\bibitem[Schulman \latin{et~al.}(2017)Schulman, Wolski, Dhariwal, Radford, and
  Klimov]{schulman2017proximal}
Schulman,~J.; Wolski,~F.; Dhariwal,~P.; Radford,~A.; Klimov,~O. Proximal policy
  optimization algorithms. \emph{arXiv preprint arXiv:1707.06347}
  \textbf{2017}, \relax
\mciteBstWouldAddEndPunctfalse
\mciteSetBstMidEndSepPunct{\mcitedefaultmidpunct}
{}{\mcitedefaultseppunct}\relax
\EndOfBibitem
\bibitem[Schulman \latin{et~al.}(2015)Schulman, Levine, Abbeel, Jordan, and
  Moritz]{schulman2015trust}
Schulman,~J.; Levine,~S.; Abbeel,~P.; Jordan,~M.; Moritz,~P. Trust region
  policy optimization. International Conference on Machine Learning. 2015; pp
  1889--1897\relax
\mciteBstWouldAddEndPuncttrue
\mciteSetBstMidEndSepPunct{\mcitedefaultmidpunct}
{\mcitedefaultendpunct}{\mcitedefaultseppunct}\relax
\EndOfBibitem
\bibitem[Deng \latin{et~al.}(2020)Deng, Yang, Li, Samaras, and
  Wang]{deng2020towards}
Deng,~J.; Yang,~Z.; Li,~Y.; Samaras,~D.; Wang,~F. Towards Better Opioid
  Antagonists Using Deep Reinforcement Learning. \emph{arXiv preprint
  arXiv:2004.04768} \textbf{2020}, \relax
\mciteBstWouldAddEndPunctfalse
\mciteSetBstMidEndSepPunct{\mcitedefaultmidpunct}
{}{\mcitedefaultseppunct}\relax
\EndOfBibitem
\bibitem[Yasonik(2020)]{yasonik2020multiobjective}
Yasonik,~J. Multiobjective de novo drug design with recurrent neural networks
  and nondominated sorting. \emph{J. Cheminformatics} \textbf{2020}, \emph{12},
  1--9\relax
\mciteBstWouldAddEndPuncttrue
\mciteSetBstMidEndSepPunct{\mcitedefaultmidpunct}
{\mcitedefaultendpunct}{\mcitedefaultseppunct}\relax
\EndOfBibitem
\bibitem[Domenico \latin{et~al.}(2020)Domenico, Nicola, Daniela, Fulvio,
  Nicola, and Orazio]{domenico2020novo}
Domenico,~A.; Nicola,~G.; Daniela,~T.; Fulvio,~C.; Nicola,~A.; Orazio,~N. De
  novo drug design of targeted chemical libraries based on artificial
  intelligence and pair-based multiobjective optimization. \emph{J. Chem. Inf.
  Model} \textbf{2020}, \emph{60}, 4582--4593\relax
\mciteBstWouldAddEndPuncttrue
\mciteSetBstMidEndSepPunct{\mcitedefaultmidpunct}
{\mcitedefaultendpunct}{\mcitedefaultseppunct}\relax
\EndOfBibitem
\bibitem[Liu \latin{et~al.}(2021)Liu, Ye, Van~Vlijmen, Emmerich, IJzerman, and
  van Westen]{liu2021drugex}
Liu,~X.; Ye,~K.; Van~Vlijmen,~H.; Emmerich,~M.; IJzerman,~A.~P.; van Westen,~G.
  DrugEx v2: De Novo Design of Drug Molecule by Pareto-based Multi-Objective
  Reinforcement Learning in Polypharmacology. \emph{ChemRxiv.14474127.v2}
  \textbf{2021}, \relax
\mciteBstWouldAddEndPunctfalse
\mciteSetBstMidEndSepPunct{\mcitedefaultmidpunct}
{}{\mcitedefaultseppunct}\relax
\EndOfBibitem
\bibitem[Reker and Schneider(2015)Reker, and Schneider]{reker2015active}
Reker,~D.; Schneider,~G. Active-learning strategies in computer-assisted drug
  discovery. \emph{Drug Discov. Today} \textbf{2015}, \emph{20}, 458--465\relax
\mciteBstWouldAddEndPuncttrue
\mciteSetBstMidEndSepPunct{\mcitedefaultmidpunct}
{\mcitedefaultendpunct}{\mcitedefaultseppunct}\relax
\EndOfBibitem
\bibitem[Walters and Murcko(2020)Walters, and Murcko]{walters2020assessing}
Walters,~W.~P.; Murcko,~M. Assessing the impact of generative AI on medicinal
  chemistry. \emph{Nat. Biotechnol.} \textbf{2020}, \emph{38}, 143--145\relax
\mciteBstWouldAddEndPuncttrue
\mciteSetBstMidEndSepPunct{\mcitedefaultmidpunct}
{\mcitedefaultendpunct}{\mcitedefaultseppunct}\relax
\EndOfBibitem
\bibitem[Sambasivan \latin{et~al.}(2021)Sambasivan, Kapania, Highfill, Akrong,
  Paritosh, and Aroyo]{sambasivan2021everyone}
Sambasivan,~N.; Kapania,~S.; Highfill,~H.; Akrong,~D.; Paritosh,~P.;
  Aroyo,~L.~M. “Everyone wants to do the model work, not the data work”:
  Data Cascades in High-Stakes AI. proceedings of the 2021 CHI Conference on
  Human Factors in Computing Systems. 2021; pp 1--15\relax
\mciteBstWouldAddEndPuncttrue
\mciteSetBstMidEndSepPunct{\mcitedefaultmidpunct}
{\mcitedefaultendpunct}{\mcitedefaultseppunct}\relax
\EndOfBibitem
\bibitem[Singh \latin{et~al.}(2018)Singh, Schulthess, Hughes, Vannieuwenhuyse,
  and Kalra]{singh2018real}
Singh,~G.; Schulthess,~D.; Hughes,~N.; Vannieuwenhuyse,~B.; Kalra,~D. Real
  world big data for clinical research and drug development. \emph{Drug Discov.
  Today} \textbf{2018}, \emph{23}, 652--660\relax
\mciteBstWouldAddEndPuncttrue
\mciteSetBstMidEndSepPunct{\mcitedefaultmidpunct}
{\mcitedefaultendpunct}{\mcitedefaultseppunct}\relax
\EndOfBibitem
\bibitem[Deng \latin{et~al.}(2021)Deng, Hou, Dong, Hajagos, Saltz, Saltz, and
  Wang]{deng2021large}
Deng,~J.; Hou,~W.; Dong,~X.; Hajagos,~J.; Saltz,~M.; Saltz,~J.; Wang,~F. A
  Large-Scale Observational Study on the Temporal Trends and Risk Factors of
  Opioid Overdose: Real-World Evidence for Better Opioids. \emph{Drugs-Real
  World Outcomes} \textbf{2021}, 1--14\relax
\mciteBstWouldAddEndPuncttrue
\mciteSetBstMidEndSepPunct{\mcitedefaultmidpunct}
{\mcitedefaultendpunct}{\mcitedefaultseppunct}\relax
\EndOfBibitem
\bibitem[Deng and Wang(2020)Deng, and Wang]{deng2020informatics}
Deng,~J.; Wang,~F. An Informatics-based Approach to Identify Key
  Pharmacological Components in Drug-Drug Interactions. \emph{AMIA Jt Summits
  Transl Sci Proc} \textbf{2020}, \emph{2020}, 142\relax
\mciteBstWouldAddEndPuncttrue
\mciteSetBstMidEndSepPunct{\mcitedefaultmidpunct}
{\mcitedefaultendpunct}{\mcitedefaultseppunct}\relax
\EndOfBibitem
\bibitem[Jiang \latin{et~al.}(2021)Jiang, Wu, Hsieh, Chen, Liao, Wang, Shen,
  Cao, Wu, and Hou]{jiang2021could}
Jiang,~D.; Wu,~Z.; Hsieh,~C.-Y.; Chen,~G.; Liao,~B.; Wang,~Z.; Shen,~C.;
  Cao,~D.; Wu,~J.; Hou,~T. Could graph neural networks learn better molecular
  representation for drug discovery? A comparison study of descriptor-based and
  graph-based models. \emph{J. Cheminformatics} \textbf{2021}, \emph{13},
  1--23\relax
\mciteBstWouldAddEndPuncttrue
\mciteSetBstMidEndSepPunct{\mcitedefaultmidpunct}
{\mcitedefaultendpunct}{\mcitedefaultseppunct}\relax
\EndOfBibitem
\end{mcitethebibliography}


\begin{tocentry}

Some journals require a graphical entry for the Table of Contents.
This should be laid out ``print ready'' so that the sizing of the text is correct.

Inside the \texttt{tocentry} environment, the font used is Helvetica 8\,pt, as required by \emph{Journal of the American Chemical
Society}.

The surrounding frame is 9\,cm by 3.5\,cm, which is the maximum permitted for  \emph{Journal of the American Chemical Society} graphical table of content entries. The box will not resize if the content is too big: instead it will overflow the edge of the box.

This box and the associated title will always be printed on a separate page at the end of the document.

\end{tocentry}

\end{document}